\documentclass[letterpaper]{article} 
\usepackage[submission]{aaai24}  
\usepackage{times}  
\usepackage{helvet}  
\usepackage{courier}  
\usepackage[hyphens]{url}  
\usepackage{graphicx} 
\usepackage{natbib}  
\usepackage{caption} 
\usepackage{algorithm2e}
\usepackage{algorithmic}
\usepackage{pythonhighlight}
\usepackage{multirow}
\usepackage{booktabs}
\title{PT-Tuning: Bridging the Gap between Time Series\\Masked Reconstruction and Forecasting via Prompt Token Tuning}
\author{
    Hao Liu\textsuperscript{\rm 1},
    Jinrui Gan\textsuperscript{\rm 1},
    Xiaoxuan Fan\textsuperscript{\rm 1},
    Yi Zhang\textsuperscript{\rm 1},
    Chuanxian Luo\textsuperscript{\rm 2},
    Jing Zhang\textsuperscript{\rm 2},\\
    Guangxin Jiang\textsuperscript{\rm 3},
    Yucheng Qian\textsuperscript{\rm 4},
    Changwei Zhao\textsuperscript{\rm 4},
    Huan Ma\textsuperscript{\rm 5},
    Zhenyu Guo\textsuperscript{\rm 5}
}
\affiliations{
    \textsuperscript{\rm 1}State Grid Smart Grid Research Institute Co.Ltd.\\
    \textsuperscript{\rm 2}Wuhan Nari Limited Liability Company of State Grid Electric Power Research Institute\\
    \textsuperscript{\rm 3}State Grid Inner Mongolia East Power Co., Ltd\\
    \textsuperscript{\rm 4}Electric Power Research Institute of State Grid Anhui Electric Power Co., Ltd\\
    \textsuperscript{\rm 5}State Grid Anhui Ultra High Voltage Company \\
}

\usepackage{bibentry}

\begin{document}

\maketitle

\begin{abstract}
Self-supervised learning has been actively studied in time series domain recently, especially for masked reconstruction. Most of these methods follow the ``Pre-training + Fine-tuning'' paradigm in which a new decoder replaces the pre-trained decoder to fit for a specific downstream task, leading to inconsistency of upstream and downstream tasks. In this paper, we first point out that the unification of task objectives and adaptation for task difficulty are critical for bridging the gap between time series masked reconstruction and forecasting. By reserving the pre-trained mask token during fine-tuning stage, the forecasting task can be taken as a special case of masked reconstruction, where the future values are masked and reconstructed based on history values. It guarantees the consistency of task objectives but there is still a gap in task difficulty. Because masked reconstruction can utilize contextual information while forecasting can only use historical information to reconstruct. To further mitigate the existed gap, we propose a simple yet effective prompt token tuning (PT-Tuning) paradigm, in which all pre-trained parameters are frozen and only a few trainable prompt tokens are added to extended mask tokens in element-wise manner. Extensive experiments on real-world datasets demonstrate the superiority of our proposed paradigm with state-of-the-art performance compared to representation learning and end-to-end supervised forecasting methods.
\end{abstract}
\section{Introduction}

Time series forecasting has became an important research problem, due to its significance in various domains, such as energy management, financial analysis, weather forecast, traffic flow estimation and so on. Recently, end-to-end methods have been actively explored, including transformer and MLP-based architectures \cite{haixu-2021,haoyietal-informer-2021,Shizhan-2021,tian-2021,Non-Stationary,N-HiTS,DLinear,TimesNet,Crossformer,PatchTST}. Owing to their powerful ability to model time series data, the forecasting performance has made great progress. However, existing models resort to similar patterns in historical time series data for predicting future values, which generally induces severe distribution problems. 

\begin{figure}[!htbp]
	\begin{center}
		\includegraphics[width=0.45\textwidth]{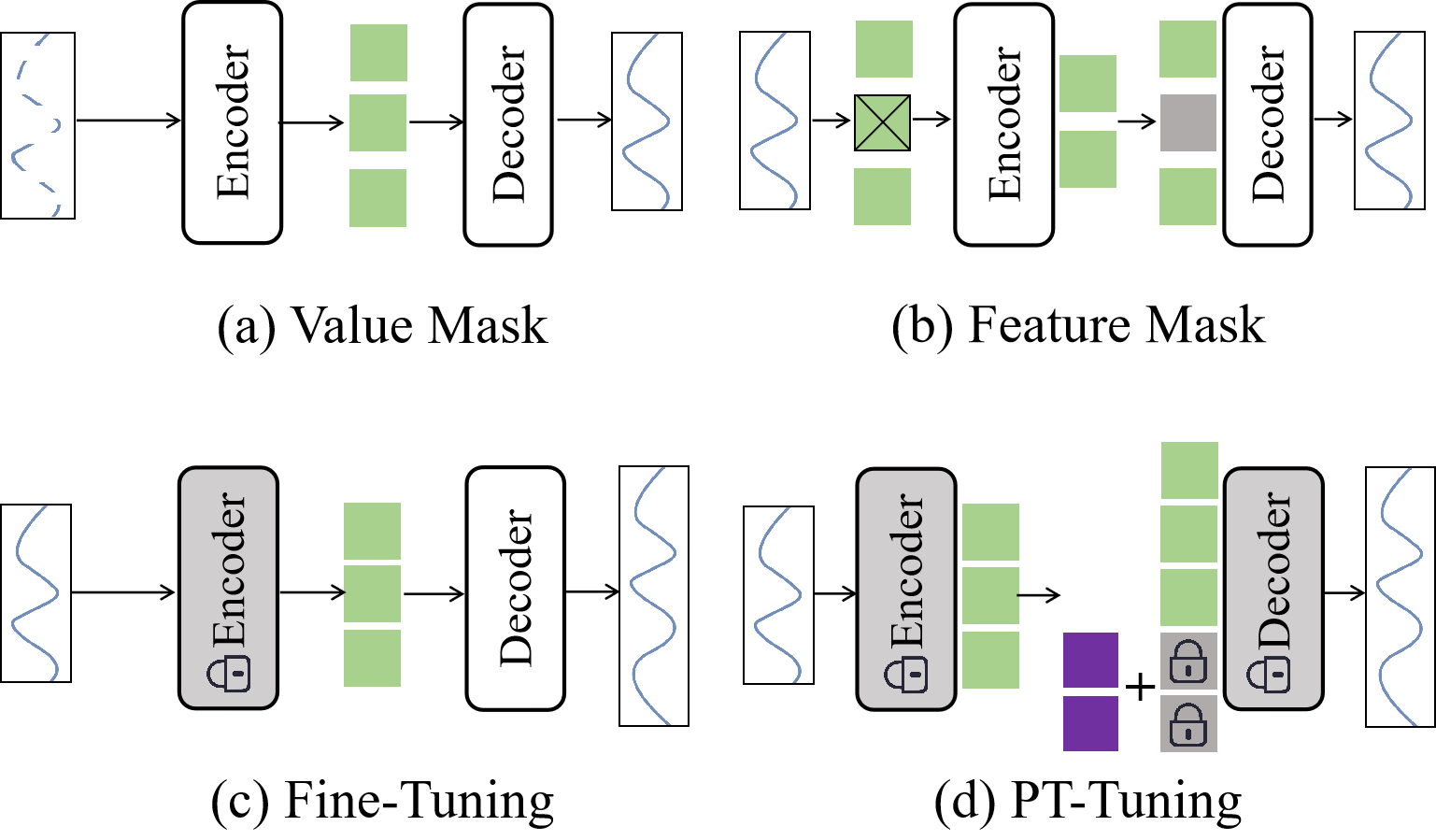}
	\end{center}
	\caption{The illustration of ``Pre-training + Fine-tuning'' paradigm. The top part includes two kinds of pre-training paradigm, value mask and feature mask. The bottom part includes two type of fine-tuning paradigm, conventional fine-tuning and our proposed PT-Tuning.}
	\label{fig1}
\end{figure}

To tackle the issue, self-supervised representation learning has aroused significant interests of researchers to extract salient temporal patterns, including contrastive learning methods \cite{Eldele-2021,Zhihan-2021,Woo-2022,BTSF,LaST,TF-C,MHCCL} and masked reconstruction methods \cite{George-2021,Ti-Gen,zezhi2022,Ti-MAE,SimMTM}. Because contrastive learning methods mainly focus on the high-level informamtion that inherently mismatches the low-level tasks \cite{Zhenda-2022}, masked reconstruction methods achieve more accurate results on time series forecasting. And masked reconstruction methods perform masking operation in two different ways. As shown in Figure 1(a), one way is to replace the values of random time series positions by zeros. We call this type of masking strategy as value mask. Another way is to apply masking on input embedding and reconstruct original time series with mask tokens, as shown in Figure 1(b). We call this kind of strategy as feature mask. These masked reconstruction methods follow the ``Pre-training + Fine-tuning'' paradigm that a task-specific decoder replaces the pre-trained decoder, as shown in Figure 1(c). However, both of these two kinds of masking strategies bring in inconsistency of task objectives between pre-traing and fine-tuning, i.e., masking strategies are introduced for reconstruction during the pre-training stage and discarded during the fine-tuning stage, making time series forecasting no longer a masked reconstruction task. 

In this paper, we firstly investigate the existed gap between time series masked reconstruction and forecasting by unifying task objectives. Specifically, by reserving the pre-trained mask token, the forecasting task can be taken as a special case of masked reconstruction, where the future values are masked and reconstructed based on history values. By doing so, it achieves better forcasting performance for short-term prediction lengths compared with traditional fine-tuning paradigm, which proves the necessity of unification for task objectives. However, forecasting task can only make use of historical information to reconstruct while masked reconstruction can utilize contextual information, leading to inconsistency of task difficulty. Although there exist some forecasting cases during pre-training stage, i.e., masked positions on the far right of time series can only use unmasked information on the left, it is only beneficial for short-term prediction. Because the range of continuous masking is not long enough to be suitable for long-term forecasting, resulting in worse performance. To further mitigate the gap, we propose a simple yet effective prompt token tuning paradigm, in which all pre-trained parameters are frozen and only a few trainable prompt tokens are added to extended mask tokens in element-wise manner, as shown in Figure 1(d). It makes the pre-trained mask token more compatible for long-term prediction. Extensive experiments on several real-world datasets support our analysis and conclusions. The main contributions of this work can be summarized as:

\begin{itemize}
    \item We point out the unification of task objectives and adaptation for task difficulty are critical for bridging the gap between time series masked reconstruction and forecasting tasks.
    \item A simple and effective prompt token tuning paradigm is proposed for the fine-tuning stage to mitigate the inconsistency problem of task difficulty via a few trainable prompt tokens.
    \item Compared with representation learning and end-to-end methods, our method achieves state-of-the-art performance on several public real-world datasets.
\end{itemize}

\section{Related Work}

\subsection{Time Series Forecasting Model}

Time series forecasting has typically been tackled as an end-to-end supervised learning task. RNN-based models \cite{RNN,Guokun-2018,Albert-2022} are often used to modeling time series but suffer from gradient vanishing problem. Owing to the ability to capture long-range dependencies based on the self-attention, recent transformer-based methods \cite{Shiyang-2019,Shizhan-2021,haoyietal-informer-2021,haixu-2021,tian-2021,Non-Stationary,Crossformer,PatchTST} adapt vanilla transformer more compatible for time series forecasting to reduce model complexity and modeling temporal correlation. Besides, MLP-based methods \cite{N-HiTS,LightTS,DLinear} adopt the MLP along the temporal dimension and encode the temporal dependencies into the fixed parameter of MLP layers.

\subsection{Time Series Representation Learning}

Recently, self-supervised learning has been actively studied in time series domain, including contrastive learning and masked reconstruction. For contrastive learning, TS2Vec \cite{Zhihan-2021} constructs temporally hierarchical representations. CoST \cite{Woo-2022} and TF-C \cite{TF-C} explore both time domain and frequency domain to learn effective representations. Moreover, LaST \cite{LaST} aims to disentangle the seasonal-trend representations in the latent space based on variational inference. However, contrastive learning mainly focuses on the high-level information and inherently mismatches the low-level tasks, such as time series forecasting. Hence, motivated by the masked modeling in language \cite{Jacob-2019} and vision \cite{BEiT,Kaiming-2022}, TST \cite{George-2021} and SimMTM \cite{SimMTM} adopt the value mask paradigm that predicts the masked time points based on the remaining points. And Ti-MAE \cite{Ti-MAE} follows the feature mask paradigm that masks input embeddings and reconstructs original time series via learnable mask token. But all of these methods follow the ``Pre-training + Fine-tuning'' paradigm, resulting in inconsistency of task objectives between pre-traing and fine-tuning.

\subsection{Prompt Learning}

Prompt learning is first proposed in natural language processing, aiming to adapt large pre-trained language models to downstream tasks \cite{Brown-2020,Jacob-2019}. Typically, prompt learning methods insert small learnable modules and fine-tune them with downstream tasks while freezing the pre-trained weights \cite{Prefix-Tuning,lester-2021,xiao-2022,Yusheng-2022,Tu-2022}. In addition to adding prompts in the text modal, \cite{Bahng-2022} and \cite{Jia-2022} propose visual prompt for adapting pre-trained vision transformer models. Notably, we are the first to utilize prompt learning in time series domain to bridge the gap between time series masked reconstruction and forecasting tasks.

\begin{figure*}[!htbp]
	\begin{center}
		\includegraphics[width=0.95\textwidth]{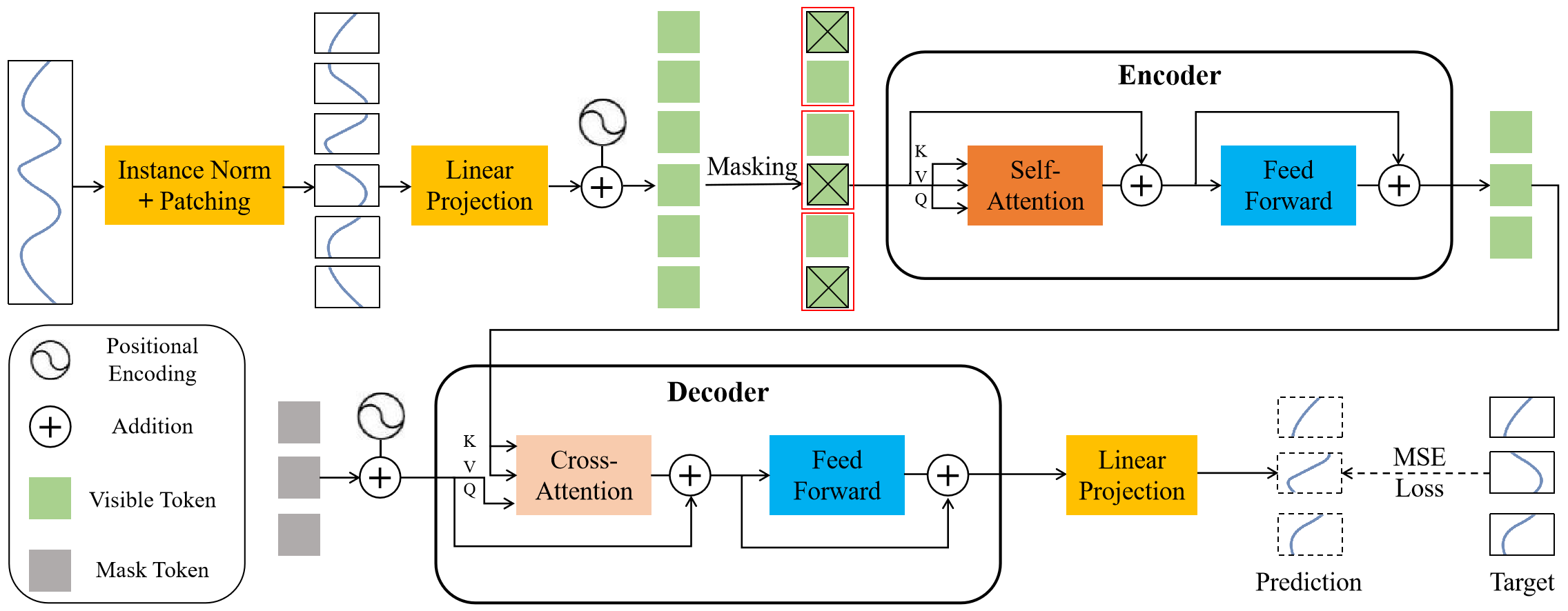}
	\end{center}
	\caption{The illustration of the proposed Cross-MAE. Three aspects are adjusted to enhance the ability of representation learning, including tokenziation, masking and decoder. We discard the layer normalization of transformer blocks for simplification.}
	\label{fig2}
\end{figure*}

\section{Gap Analysis between Time Series Masked Reconstruction and Forecasting}
Without loss of generality, we first introduce an enhanced masked reconstruction model for gap analysis between time series masked reconstruction and forecasting, then describe four kinds of experiment schemes based on the pre-trained model, finally point out the two critial factors for bridging the gap, i.e., unification of task objectives and adaptation for task difficulty. 

\subsection{Model Architecture}

To improve the ability of representation learning, we propose an enhanced masked reconstruction model, called Cross-MAE. The overall architecture of Cross-MAE for the pre-training stage is shown in Figure 2. We improve the original MAE-style framework in three aspects, including tokenziation, masking strategy and decoder.

\subsubsection{Tokenization.} Following PatchTST \cite{PatchTST}, We adopt the idea of patching that divides the time series into non-overlapped patches and maps them into tokens with a linear projection layer, which extracts local semantic information and reduces the time and space complexity simultaneously. And fixed sinusoidal positional embeddings are added to maintain the position information. Notably, instance normalization is also utilized to help mitigating the distribution shift effect between the training and testing data \cite{RevIN}. Besides, multivariate time series data is divided into different channels and shares the same model.

\subsubsection{Masking.} After tokenizing the original time seires into tokens, we adopt a more suitable masking strategy for the patch-level tokenization to learning better representation, called isometric sub-sequence masking, where entire token sequence is equally divided and masked positions are randomly sampled in the sub-sequence to main the same masking ratio. Notably, the patch-level reconstruction task is inherently more difficult than the point-level task. And the original random masking strategy enables the range of continous masking positions to be longer, which makes too much context information missing, resulting in more challenging reconstruction task. 

\subsubsection{Encoder.} Our encoder utilizes two layers of vanilla Transformer blocks with pre-norm in each block. And it is applied only on visible tokens after masking. By doing so, it makes the encoder more efficient during the pre-training stage.

\subsubsection{Decoder.} Our decoder also contains two layers of Transformer blocks. Differently, we adopt cross-attention instead of self-attention to alleviate the error accumulation problem. Specifically, the conventional decoder with self-attention performs information interaction on the union of the encoded visible tokens and learnable randomly initialized mask tokens. But the uncertainty mask tokens will bring in noises for both visible and mask tokens during self-attention process, resulting in error propogation. By decoupling these two kinds of tokens, the mask tokens only acquire information from the visible tokens, leading to better representation of time series.

\subsubsection{Reconstruction Target.} A linear predictor reconstructs the original time series patches by predicting all the values at masked positions based on the decoded tokens. The training loss computes the mean squared error between the reconstructed and original time series data over masking patches.

\subsection{Scheme Design}

Based on the pre-trained Cross-MAE, we design four kinds of experiment settings to investigate the existed gap between time series masked reconstruction and forecasting, including the conventional fine-tuning, direct forecasting, linear probing and mask token fine-tuning, as shown in Figure 3. Specifically, following other representation learning based methods \cite{Ti-MAE,SimMTM}, the conventional fine-tuning strategy freezes the pre-trained encoder and replaces the pre-trained decoder with linear predictor. Direct forecasting paradigm freezes all pre-trained parameters and predicts the future values based on the history values by extending the pre-trained mask token at future positions, which unifies the task objectives between masked reconstruction and forecasting. Moreover, linear probing follows the direct forecasting paradigm and only makes the pre-trained linear predictor tunable, while mask token fine-tuning only makes the pre-trained mask token tunable.

\begin{figure}[!ht]
	\begin{center}
		\includegraphics[width=0.45\textwidth]{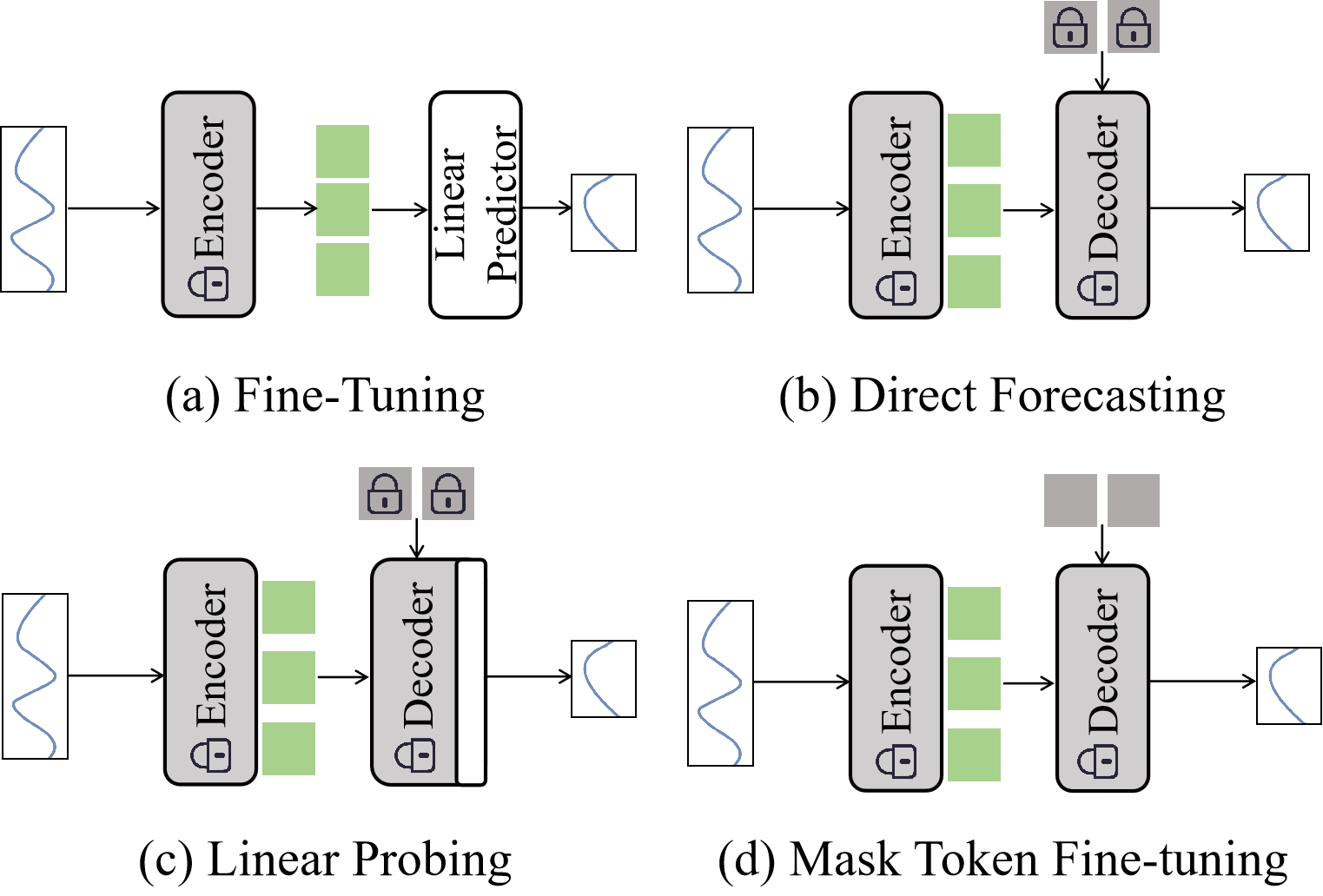}
	\end{center}
	\caption{The illustration of four kinds of experiment settings. The top two settings are responsible for unification analysis of task objectives. And the bottom two settings provide evidence for the necessity of adaption for task difficulty.}
	\label{fig3}
\end{figure}

\subsection{Gap Analysis}

\subsubsection{Results Analysis.} As shown in Figure 4, we report the forecasting results of four kinds of paradigms on the ETTh1 and ETTm1 datasets. Specifically, compared with the conventional fine-tuning paradigm, the direct forecasting paradigm achieves better performance consistently for the short-term prediction horizons (96 and 192) without any fine-tuning process. And the forecasting performance can be further improved by adopting the other two fine-tuning strategies, which proves the necessity and potentiality of unification for task objectives. However, there is a huge performance gap for the long-term prediction horizons (336 and 720), even with fine-tuning process. In our opinion, the task difficulty of masked reconstruction is different from masked forecasting. Because forecasting task can only make use of historical information to reconstruct future values while masked reconstruction can utilize contextual information, leading to inconsistency of task difficulty. Thereby, more suitable fune-tuning strategies need to be designed, making the pre-trained mask token compatible for both short-term and long-term prediction lengths.

\subsubsection{Representation Analysis.} To explore the reason why the time series forecasting can be taken as a special case of masked reconstruction, we also visualize the pre-trained feature distribution by using t-SNE \cite{TSNE}, as shown in Figure 5. It is obvious that the pre-trained mask token almost locates at the center of feature space, proving that it captures the clustered information from global perspective so that severe distribution shift problem can be alleviated. Hence, the pre-trained mask token is able to be generalized to different mask reconstruction cases, which explains the prediction performance of direct forecasting paradigm.

\subsubsection{Conclusion.} According to the experimental results and representation analysis, we indicate that the existed gaps between time series masked reconstruction and forecasting contain the unification of task objectives and adaptation for task difficulty, which is instructive for current representation learning based methods and deserves further study.

\begin{figure}[!t]
	\begin{center}
		\begin{minipage}{0.494\linewidth}
			\includegraphics[width=\linewidth]{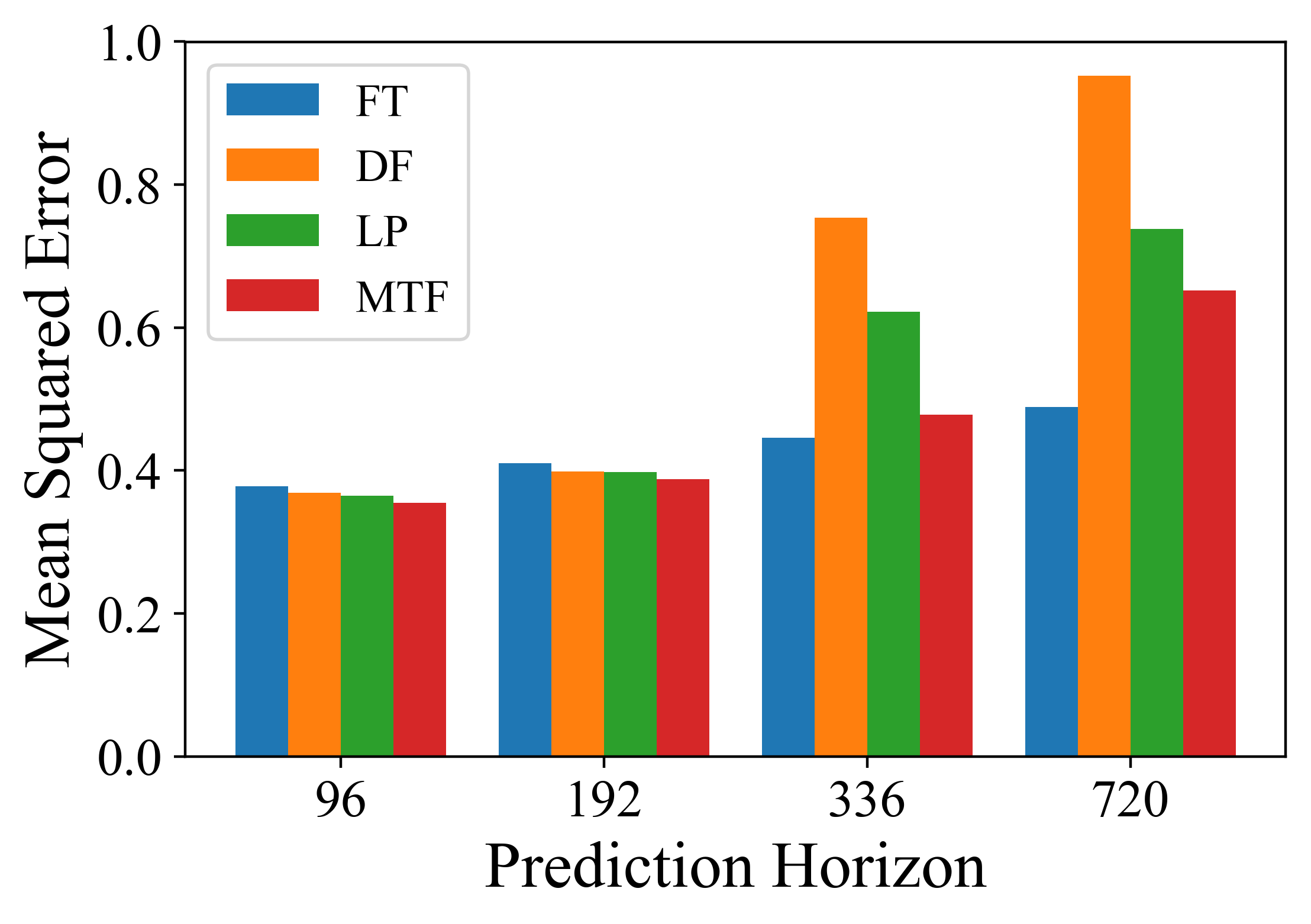}
		\end{minipage}
		\begin{minipage}{0.494\linewidth}
			\includegraphics[width=\linewidth]{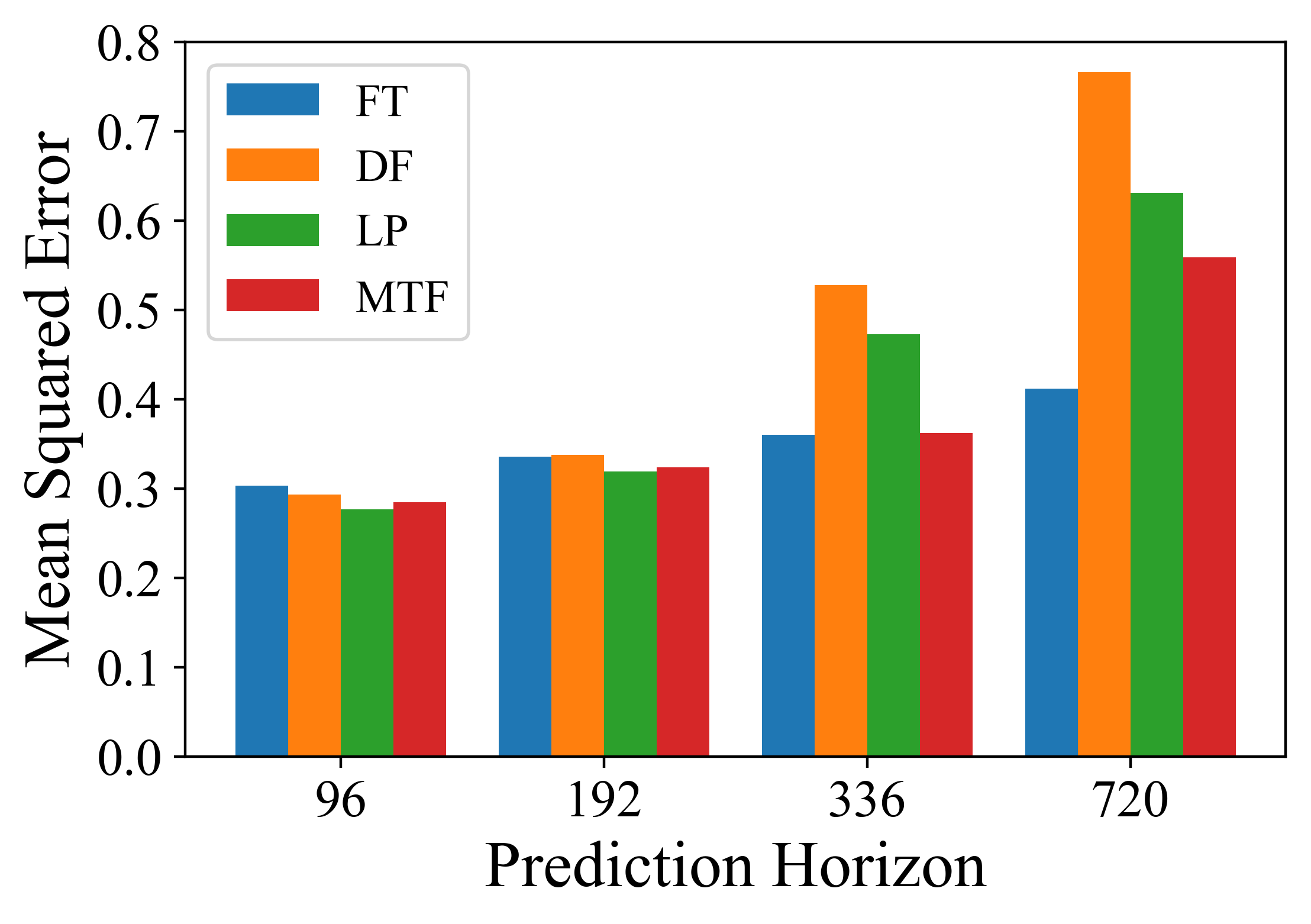}
		\end{minipage}
		\qquad			

		\begin{minipage}{0.494\linewidth}
			\includegraphics[width=\linewidth]{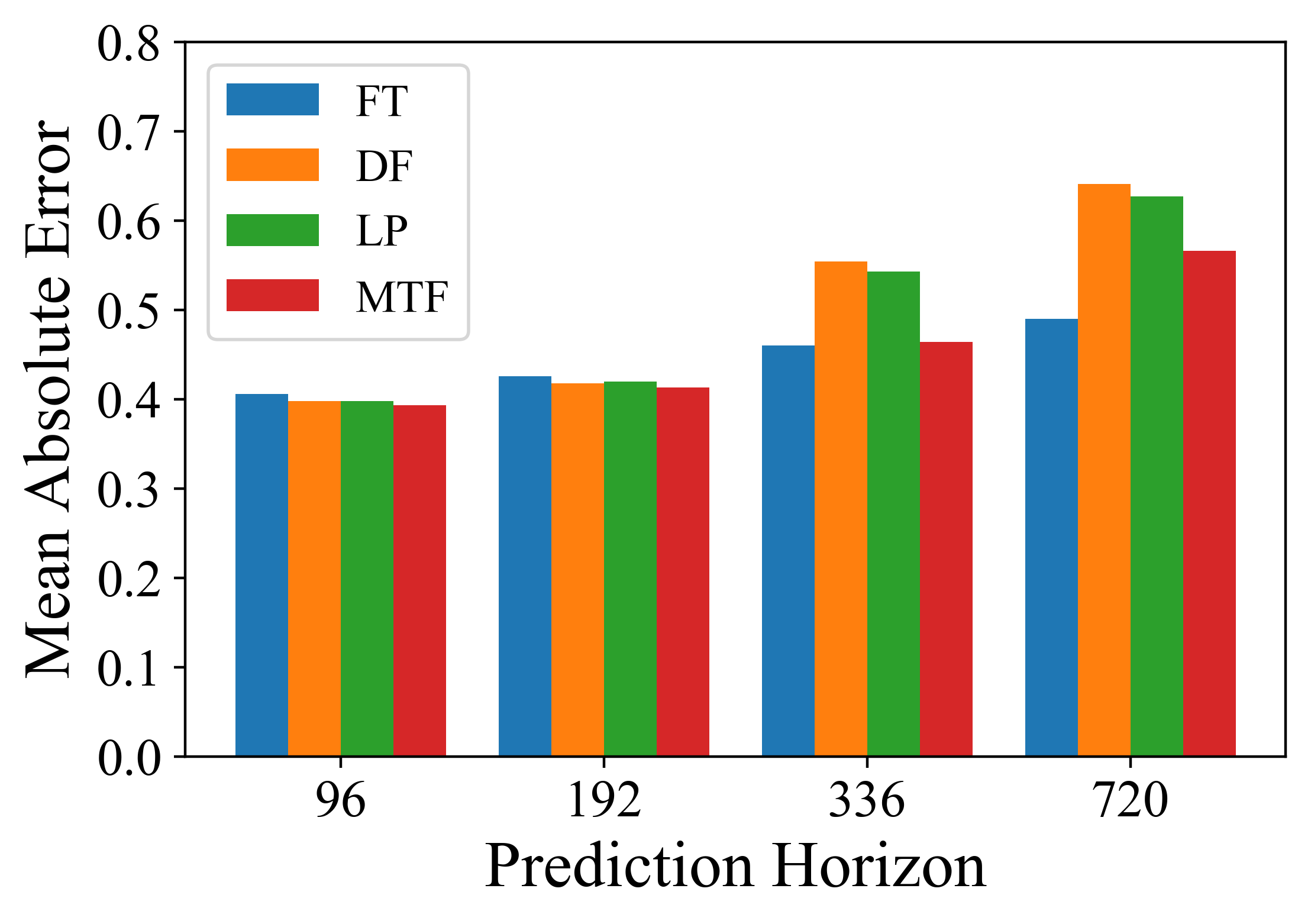}
			\centerline{(a) ETTh1}
		\end{minipage}
		\begin{minipage}{0.494\linewidth}
			\includegraphics[width=\linewidth]{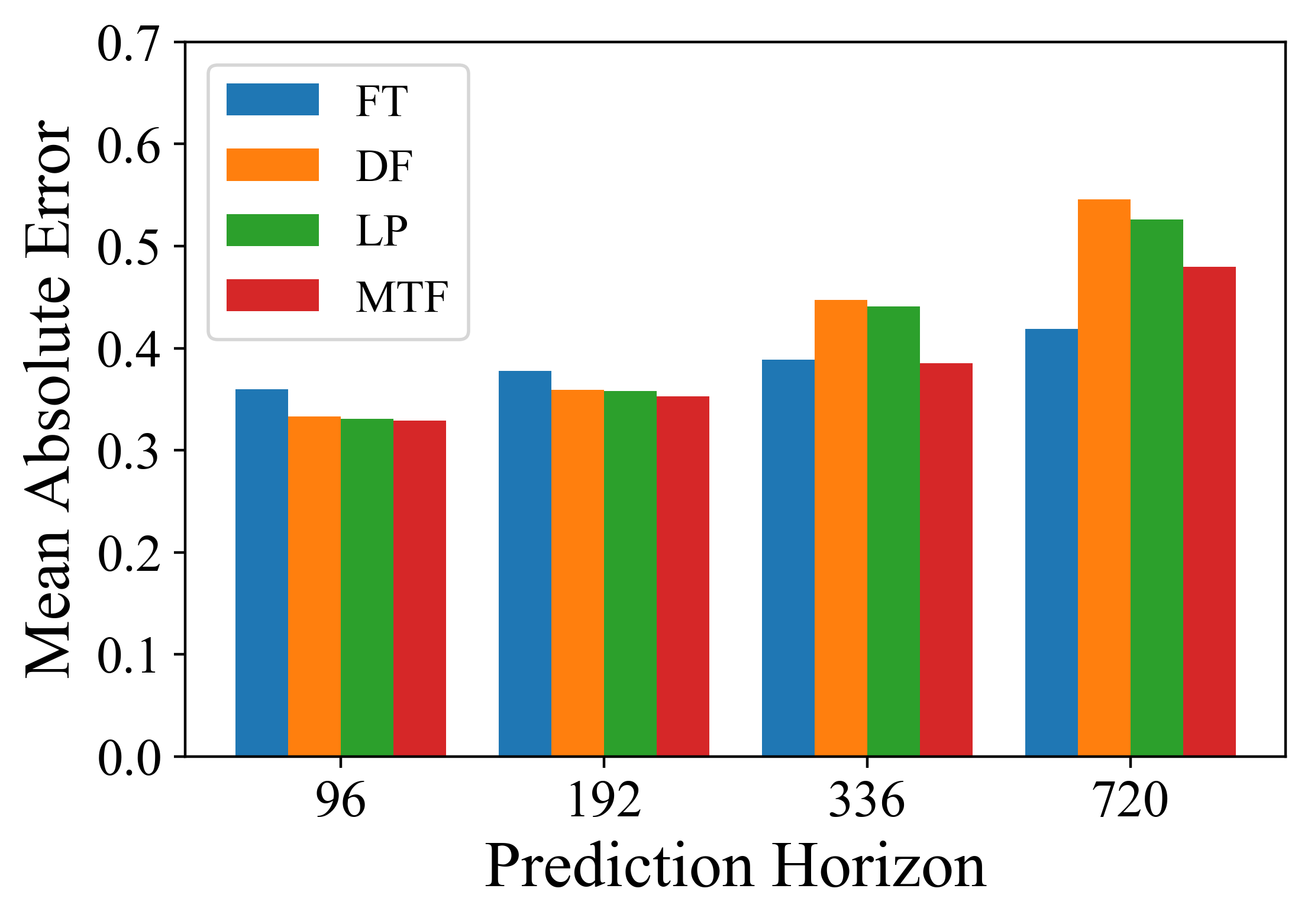}
			\centerline{(b) ETTm1}
		\end{minipage}
	\end{center}
	\caption{Unification effect of task objectives on ETTh1 and ETTm1 datasets. ``FT'', ``DF'', ``LP'' and ``MTF'' refers to fine-tuning, direct forecasting, linear probing and mask token fine-tuning respectively.}
	\label{fig3}
\end{figure}

\begin{figure}[!t]
	\begin{center}
		\begin{minipage}{0.494\linewidth}
			\includegraphics[width=\linewidth]{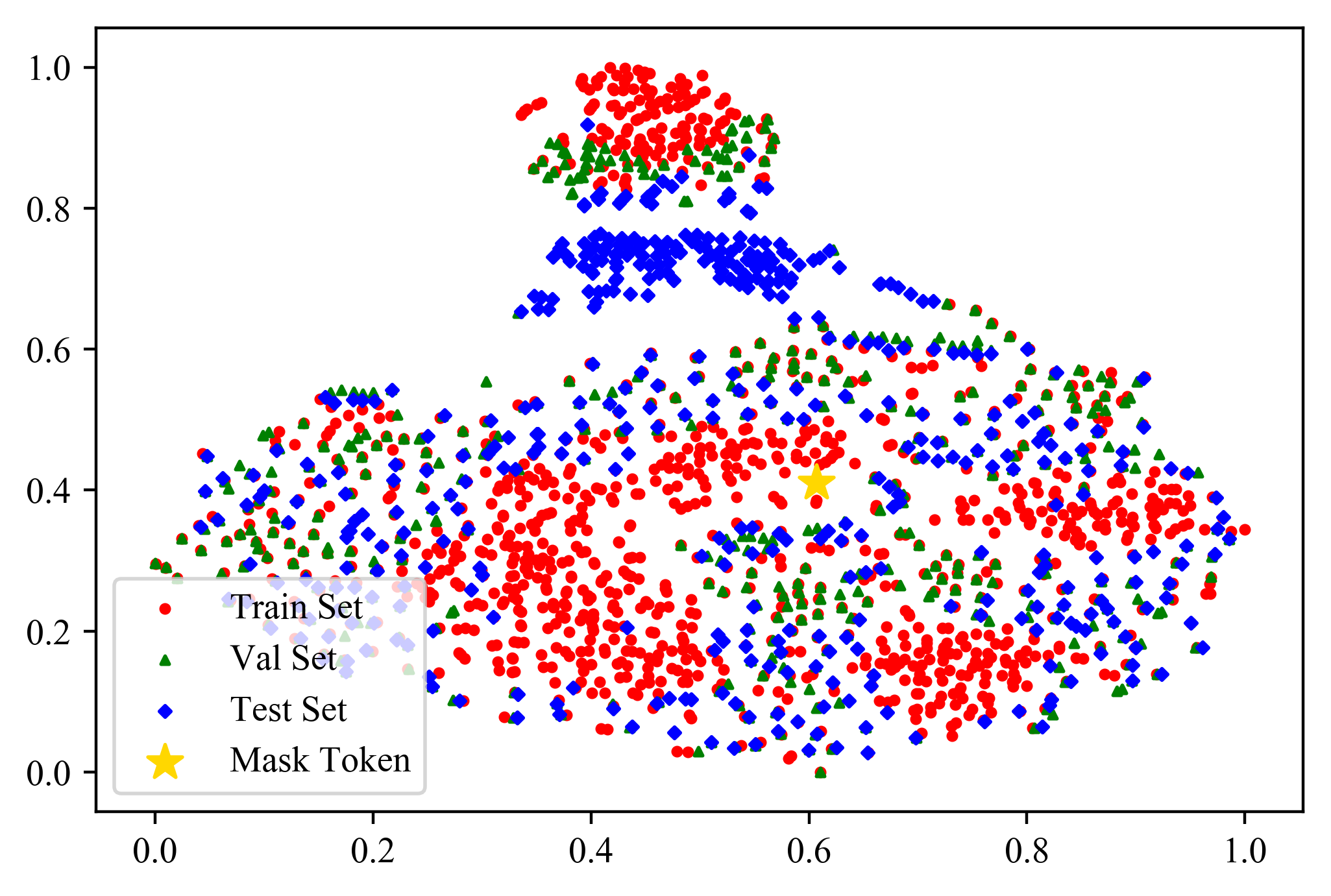}
			\centerline{(a) ETTh1}
		\end{minipage}
		\begin{minipage}{0.494\linewidth}
			\includegraphics[width=\linewidth]{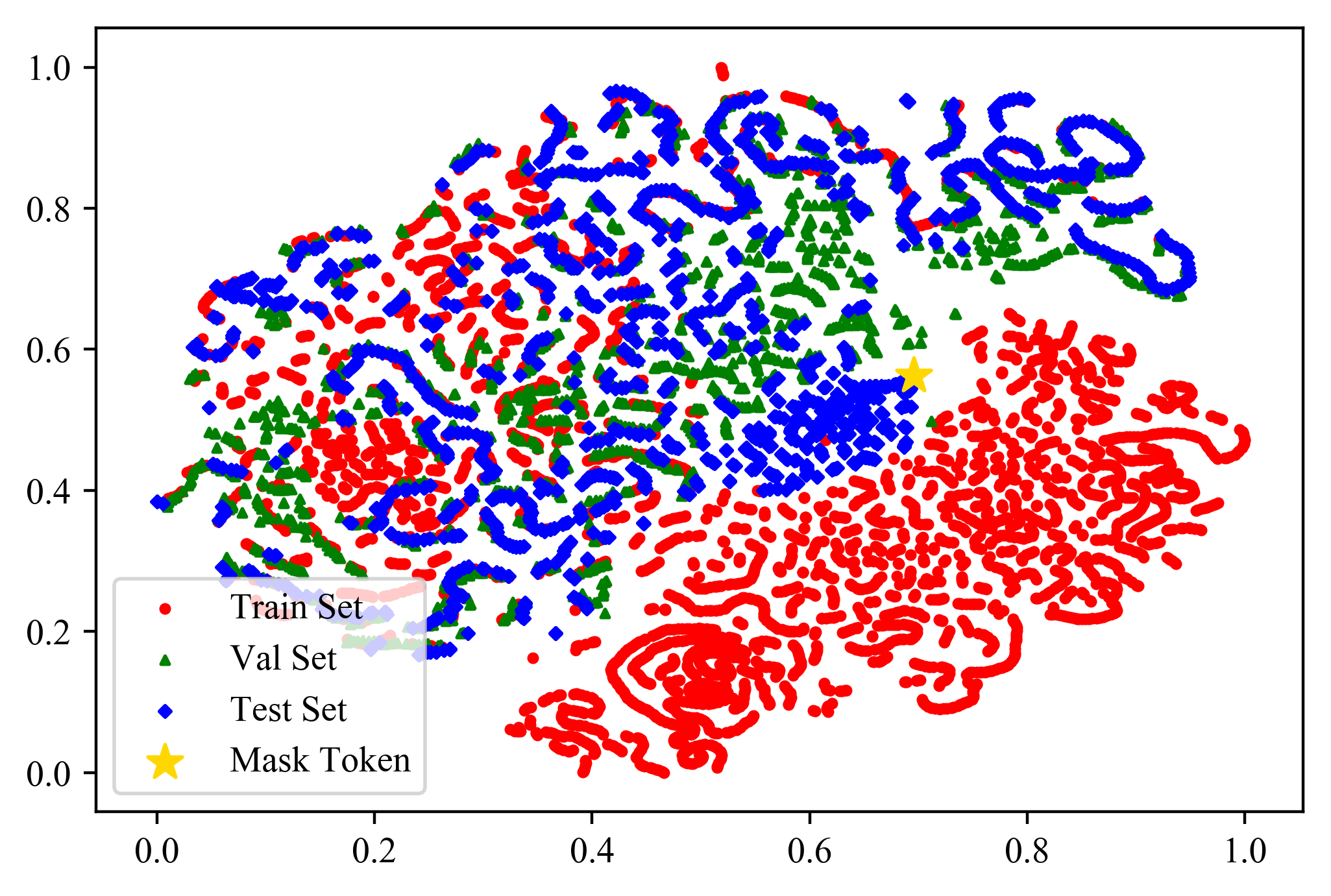}
			\centerline{(a) ETTm1}
		\end{minipage}
	\end{center}
	\caption{T-SNE visualization of encoded tokens and mask token for ETTh1 and ETTm1 datasets.}
	\label{fig4}
\end{figure}

\section{Prompt Token Tuning}

\begin{figure*}[!ht]
	\begin{center}
		\includegraphics[width=0.95\textwidth]{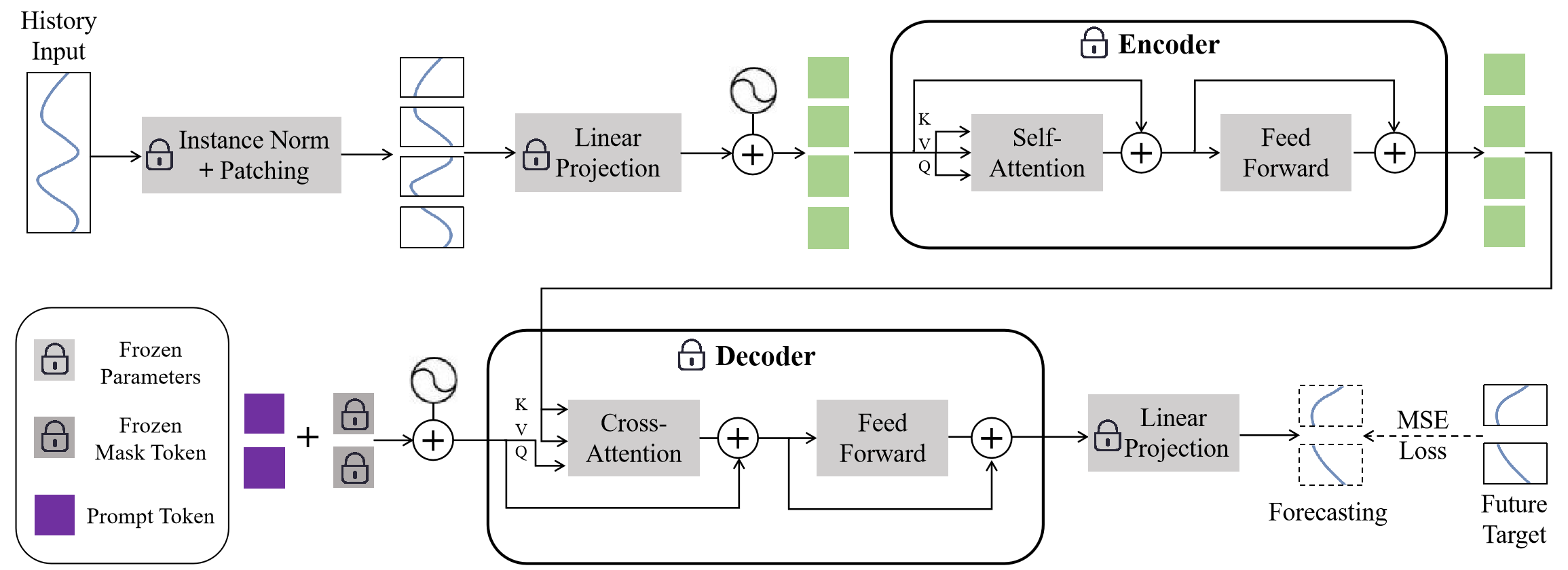}
	\end{center}
	\caption{The illustration of the proposed PT-Tuning paradigm. All pre-trained parameters are frozen and only a few trainable prompt tokens are added to extended mask tokens in element-wise manner.}
	\label{fig2}
\end{figure*}

According to the previous analysis, more suitable fune-tuning strategies need to be designed to make the pre-trained mask token compatible for more difficult forecasting task. Motivated by the success of prompt learning \cite{Prefix-Tuning,xiao-2022}, we propose a simple yet effective prompt token tuning (PT-Tuning) paradigm to further bridge the gap between time series masked reconstruction and forecasting. As shown in Figure 6, during the prompt token tuning stage, we freeze all pre-trained model parameters to unify the task objectives between masked reconstruction and forecasting. And  learnable prompt tokens are introduced to be adapted for inconsistency of task difficulty.

Specifically, a series of mask tokens are copied from the pre-trained mask token to be placeholders for future positions. And the same number of prompt tokens are directly added element-wise to those mask tokens. Moreover, we add fixed positional embeddings to provide the position information. So the future tokens are computed as follows: 
\begin{equation}
	F = M + P + E_p
\end{equation}
where $M$, $P$ and $E_p$ denotes mask tokens, prompt tokens and positional embeddings respectively. Then we generate a set of queries via linear projection of the future tokens. Keys and values are also calculated via linear projection of the history tokens $H$, which can be defined as follows: 
\begin{equation}
	Q = L_q(F), K = L_k(H), V = L_v(H)
\end{equation}
where $L_q$, $L_k$ and $L_v$ are MLP layers. Finally, we denote the cross attention by
\begin{equation}
	Attn(Q,K,V) = Softmax(\frac{QK^T}{\sqrt{d_k}})V
\end{equation}

By doing so, PT-Tuning mitigates the inconsistency problem of task difficuly and further improves the forecasting performance significantly. Moreover, only a small amount of trainable parameters are introduced, which reduces the risk of over-fitting during the fine-tuning stage. Besides, compared with the original fine-tuning paradigm, our paradigm only needs gradients for the prompt tokens and does not save any intermediate gradient results, making it memory-efficient. 

\section{Experiments}

\subsection{Experimental Setup}
\subsubsection{Implementation.} We conduct experiments on seven public benchmark datasets, including four Electricity Transformer Temperature (ETT) datasets, Weather, Electricity and Traffic. More details about the datasets, experiment configurations, model configurations, metrics and codes can be found in Appendix A.



\subsubsection{Baselines.} For representation learning methods, we select seven latest state-of-the-art models, including masked reconstruction \cite{SimMTM,Ti-MAE,George-2021} and contrastive learning based methods \cite{LaST,TF-C,Woo-2022,Zhihan-2021}. And for end-to-end supervised methods, other seven baselines are chosen, including Transformer-based \cite{PatchTST,Crossformer,tian-2021,Non-Stationary}, MLP-based \cite{DLinear} and CNN-based \cite{MICN,TimesNet} methods.

\begin{table}[!t]
\centering
\begin{tabular}{c|c|cc|cc}
\toprule[1.5pt]
\multicolumn{2}{c|}{Dataset} &  \multicolumn{2}{c|}{ETTh1}  &  \multicolumn{2}{c}{ETTm1}\\
\midrule[1pt]
\multicolumn{2}{c|}{Metric} & MSE & MAE & MSE & MAE \\
\midrule[1pt]
\multirow{2}{*}{BEiT-Style} & FT & 0.448 & 0.454 & 0.380 & 0.394 \\
& PT-T & \textbf{0.423} & \textbf{0.438} & \textbf{0.359} & \textbf{0.386} \\
\midrule[1pt]
\multirow{2}{*}{MAE-Style} & FT & 0.445 & 0.454 & 0.355 & 0.386 \\
& PT-T & \textbf{0.418} & \textbf{0.440} & \textbf{0.352} & \textbf{0.377} \\
\midrule[1pt]
\multirow{2}{*}{Cross-MAE} & FT & 0.431 & 0.446 & 0.353 & 0.387 \\
&PT-T & \textbf{0.394} & \textbf{0.418} & \textbf{0.340} & \textbf{0.364} \\
\bottomrule[1.5pt]
\end{tabular}
\caption{Performance by applying PT-Tuning to three pretraining frameworks. ``FT'' and ``PT-T'' refers to fine-tuning and prompt token tuning respectively.}
\label{table1}
\end{table}

\begin{table*}[!t]\footnotesize
\centering
\renewcommand\arraystretch{1.3}
\resizebox{\linewidth}{!}{
\begin{tabular}{c|cc|cc|cc|cc|cc|cc|cc|cc}
\toprule[1.2pt]
Method &  \multicolumn{2}{c|}{PT-Tuning}  &  \multicolumn{2}{c|}{SimMTM}  & \multicolumn{2}{c|}{Ti-MAE}  &  \multicolumn{2}{c|}{TST}  &  \multicolumn{2}{c|}{LaST} & \multicolumn{2}{c|}{TF-C} & \multicolumn{2}{c|}{CoST} &
\multicolumn{2}{c}{TS2Vec} \\
\midrule[1.2pt]
Metric &  MSE & MAE &  MSE & MAE  &  MSE & MAE  &  MSE & MAE   &  MSE & MAE & MSE & MAE  & MSE & MAE & MSE & MAE \\
\midrule[1.2pt]
ETTm1 & \textbf{0.340} & \textbf{0.364} & \textbf{0.340} & \underline{0.379} & 0.366 & 0.391 & 0.373 & 0.389 & 0.398 & 0.398 & 0.744 & 0.652 & 0.356 & 0.385 & 0.699 & 0.557 \\
\midrule[0.8pt]
ETTm2 & \textbf{0.251} & \textbf{0.306} & \underline{0.260} & \underline{0.318} & 0.264 & 0.320 & 0.297 & 0.347 & 0.255 & 0.326 & 1.755 & 0.947 & 0.314 & 0.365 & 0.326 & 0.361 \\
\midrule[0.8pt]
ETTh1 & \textbf{0.394} & \textbf{0.418} & \underline{0.404} & \underline{0.428} & 0.423 & 0.446 & 0.466 & 0.462 & 0.474 & 0.461 & 0.637 & 0.638 & 0.485 & 0.472 & 0.446 & 0.456 \\
\midrule[0.8pt]
ETTh2 & \textbf{0.337} & \textbf{0.381} & \underline{0.348} & \underline{0.391} & 0.380 & 0.386 & 0.404 & 0.421 & 0.499 & 0.497 & 0.398 & 0.398 & 0.399 & 0.427 & 0.417 & 0.468 \\
\midrule[0.8pt]
Weather & \textbf{0.227} & \textbf{0.261} & \underline{0.228} & \underline{0.267} & 0.230 & 0.265 & 0.239 & 0.276 & 0.232 & 0.261 & 0.286 & 0.349 & 0.324 & 0.329 & 0.233 & 0.267 \\
\midrule[0.8pt]
Electricity & \textbf{0.157} & \textbf{0.248} & \underline{0.162} & \underline{0.256} & 0.205 & 0.296 & 0.209 & 0.289 & 0.186 & 0.274 & 0.363 & 0.398 & 0.215 & 0.295 & 0.213 & 0.293 \\
\midrule[0.8pt]
Traffic & \underline{0.393} & \textbf{0.256} & \textbf{0.392} & \underline{0.264} & 0.475 & 0.310 & 0.586 & 0.362 & 0.713 & 0.397 & 0.717 & 0.456 & 0.435 & 0.362 & 0.470 & 0.350 \\
\midrule[1.2pt]
Avg & \textbf{0.300} & \textbf{0.319} & \underline{0.305} & \underline{0.329} & 0.335 & 0.345 & 0.368 & 0.364 & 0.395 & 0.373 & 0.700 & 0.548 & 0.361 & 0.376 & 0.401 & 0.393 \\
\bottomrule[1.2pt]
\end{tabular}}
\caption{Multivariate time series forecasting results compared with representation learning methods. All results are averaged from four prediction lengths. A lower MSE or MAE indicates a better prediction. Full results can be found in Appendix C.}
\label{table1}
\end{table*}

\subsection{Ablation Study}

In this subsection, ablation studies are performed on ETTh1 and ETTm1 datasets to demonstrate the effectiveness of the proposed Cross-MAE and PT-Tuning paradigm. We report the average MSE and MAE from all prediction lengths in \{96,192,336,720\}. All the detailed results for each forecasting horizon and more experiments are in Appendix B.

\subsubsection{Model Generality.} To verify the effectiveness of our prompt token tuning, we adopt three different pre-training frameworks that follow the feature mask paradigm, including BEiT-style, MAE-style and our Cross-MAE. From Table 2, we can find that PT-Tuning consistently improve the forecasting performance of diverse frameworks compared with the conventional fine-tuning paradigm. This generality also indicates that we can further improve the performance by employing better pre-training frameworks in the future.

\subsubsection{Prompt Strategy.} Note that the prompt tokens can be placed at different locations with different combination styles. So we explore the impact of prompt strategies by taking styles and loccations into consideration. Specifically, Prompt-C and Prompt-A denote concatenation and addition with prompt tokens respectively. Moreover, history and future means that prompt tokens are collaborative with history and future tokens respectively. As shown in Table 3, by adding prompt tokens to future tokens (extended mask tokens) in element-wise manner, the forecasting performance can be improved significantly.

\begin{table}[!t]
\centering
\begin{tabular}{c|c|cc|cc}
\toprule[1.5pt]
\multicolumn{2}{c|}{Dataset} &  \multicolumn{2}{c|}{ETTh1}  &  \multicolumn{2}{c}{ETTm1}\\
\midrule[1pt]
\multicolumn{2}{c|}{Metric} & MSE & MAE & MSE & MAE \\
\midrule[1pt]
Prompt-C & History & 0.449 & 0.451 & 0.346 & 0.373 \\
\midrule[1pt]
\multirow{2}{*}{Prompt-A} & History & 0.429 & 0.439 & 0.345 & 0.370 \\
&Future & \textbf{0.394} & \textbf{0.418} & \textbf{0.340} & \textbf{0.364} \\
\bottomrule[1.5pt]
\end{tabular}
\caption{The impact of different prompt strategies on ETTh1 and ETTm1. ``Prompt-C'' and ``Prompt-A'' refers to concatenation and addition respectively.}
\label{table1}
\end{table}

\subsubsection{Masking Strategy.} Table 4 studies the impact of different mask sampling strategies. The periodic masking strategy samples continuous tokens equidistantly to maintain the same masking ratio. And the continuous masking strategy tends to reconstruct the masked future values based on the historical values, which is the same as forecasting task. Notably, compared with random masking strategy, our masking strategy achieves the best results because it makes the difficulty of reconstruction task moderate at the patch-level.

\subsubsection{Masking Ratio.} Figure 7 shows the influence of different masking ratios. Similar to Ti-MAE \cite{Ti-MAE}, the masking ratio of 75\% achieves the best performance, proving that the information density of time series is lower. For lower ratios, masked values in time series can be reconstructed by simple average or interpolation, which creates a trivial self-supervised task and can not acquire useful information. And for higher ratios, too much contextual information are lost and temporal patterns can not be extracted.

\begin{table}[!t]
\centering
\begin{tabular}{c|cc|cc}
\toprule[1.5pt]
Dataset &  \multicolumn{2}{c|}{ETTh1}  &  \multicolumn{2}{c}{ETTm1}\\
\midrule[1pt]
Metric & MSE & MAE & MSE & MAE \\
\midrule[1pt]
Periodic & 0.569 & 0.509 & 0.393 & 0.407 \\
\midrule[1pt]
Continuous & 0.456 & 0.451 & 0.385 & 0.402 \\
\midrule[1pt]
Random & 0.438 & 0.449 & 0.343 & 0.366 \\
\midrule[1pt]
Isometric & \textbf{0.394} & \textbf{0.418} & \textbf{0.340} & \textbf{0.364} \\
\bottomrule[1.5pt]
\end{tabular}
\caption{The impact of different masking strategies with 75\% ratio on ETTh1 and ETTm1.}
\label{table1}
\end{table}

\begin{table*}[!t]\footnotesize
\centering
\renewcommand\arraystretch{1.3}
\resizebox{\linewidth}{!}{
\begin{tabular}{c|cc|cc|cc|cc|cc|cc|cc|cc}
\toprule[1.2pt]
Method &  \multicolumn{2}{c|}{PT-Tuning}  &  \multicolumn{2}{c|}{PatchTST}  & \multicolumn{2}{c|}{DLinear}  &  \multicolumn{2}{c|}{TimesNet}  &  \multicolumn{2}{c|}{MICN} & \multicolumn{2}{c|}{Crossformer} & \multicolumn{2}{c|}{FEDformer} &
\multicolumn{2}{c}{Stationary} \\
\midrule[1.2pt]
Metric &  MSE & MAE &  MSE & MAE  &  MSE & MAE  &  MSE & MAE   &  MSE & MAE & MSE & MAE  & MSE & MAE & MSE & MAE \\
\midrule[1.2pt]
ETTm1 & \textbf{0.340} & \textbf{0.364} & \underline{0.353} & 0.382 & 0.356 & \underline{0.380} & 0.400 & 0.406 & 0.387 & 0.411 & 0.424 & 0.439 & 0.448 & 0.452 & 0.481 & 0.456 \\
\midrule[0.8pt]
ETTm2 & \textbf{0.251} & \textbf{0.306} & \underline{0.256} & \underline{0.317} & 0.267 & 0.332 & 0.291 & 0.333 & 0.284 & 0.340 & 0.509 & 0.522 & 0.305 & 0.349 & 0.306 & 0.347 \\
\midrule[0.8pt]
ETTh1 & \textbf{0.394} & \textbf{0.418} & \underline{0.413} & \underline{0.434} & 0.423 & 0.437 & 0.458 & 0.450 & 0.440 & 0.462 & 0.437 & 0.461 & 0.440 & 0.460 & 0.570 & 0.537 \\
\midrule[0.8pt]
ETTh2 & \underline{0.337} & \textbf{0.381} & \textbf{0.331} & \textbf{0.381} & 0.431 & 0.447 & 0.414 & 0.427 & 0.402 &  0.437 & 0.454 & 0.446 & 0.437 & 0.449 & 0.526 & 0.516 \\
\midrule[0.8pt]
Weather & \underline{0.227} & \textbf{0.261} & \textbf{0.226} & \underline{0.264} & 0.246 & 0.300 & 0.259 & 0.287 & 0.243 & 0.299 & 0.232 & 0.295 & 0.309 & 0.360 & 0.288 & 0.314 \\
\midrule[0.8pt]
Electricity & \textbf{0.157} & \textbf{0.248} & \underline{0.159} & \underline{0.253} & 0.166 & 0.264 & 0.192 & 0.295 & 0.187 & 0.295 & 0.280 & 0.343 & 0.214 & 0.327 & 0.193 & 0.296 \\
\midrule[0.8pt]
Traffic &\underline{0.393} & \textbf{0.256} & \textbf{0.391} & \underline{0.264} & 0.434 & 0.295 & 0.620 & 0.336 & 0.542 & 0.316 & 0.534 & 0.304 & 0.610 & 0.376 & 0.624 & 0.340 \\
\midrule[1.2pt]
Avg & \textbf{0.300} & \textbf{0.319} & \underline{0.304} & \underline{0.328} & 0.332 & 0.351 & 0.376 & 0.362 & 0.355 & 0.366 & 0.410 & 0.401 & 0.395 & 0.396 & 0.427 & 0.401 \\
\bottomrule[1.2pt]
\end{tabular}}
\caption{Multivariate time series forecasting results compared with end-to-end methods. All results are averaged from four prediction lengths. A lower MSE or MAE indicates a better prediction. Full results can be found in Appendix C.}
\label{table1}
\end{table*}

\subsection{Comparison}
We summarize the results of proposed PT-Tuning compared with the representation learning and end-to-end methods in Table 2 and Table 5 respectively. As can be observed, PT-Tuning consistently outperforms other representation learning methods. On the average of all benchmarks, PT-Tuning achieves 1.6\% MSE reduction and 2.7\% MAE reduction compared to the masked reconstruction baseline SimMTM \cite{SimMTM}, 16.9\% MSE reduction and 15.2\% MAE reduction compared to the contrastive baseline CoST \cite{Woo-2022}.  As for end-to-end methods, our method also performs state-of-the-art performance in most cases, achieving 1.3\% MAE reduction and 2.7\% MSE reduction compared to the strong baseline PatchTST \cite{PatchTST}. All these results prove that the unification of task objectives and adaptation for task difficulty bridge the gap between time series masked reconstruction and forecasting.

\subsection{Model Efficiency}

To evaluate the model efficiency, we report the average parameters, memories and running times of the forecasting task on ETTh1 dataset. As shown in Table 6, unlike other representation learning and end-to-end methods, our model does not rely on the linear prediction head that projects the history feature into future values, resulting in the smallest number of parameters. But Cross-MAE performs comparable memory consumption and running speed compared to other strong baselines because it is still based on the self-attention. Overally, our method achieves the best balance by taking the forecasting performance into acount.

\begin{figure}[!t]
	\begin{center}
		\includegraphics[width=\linewidth]{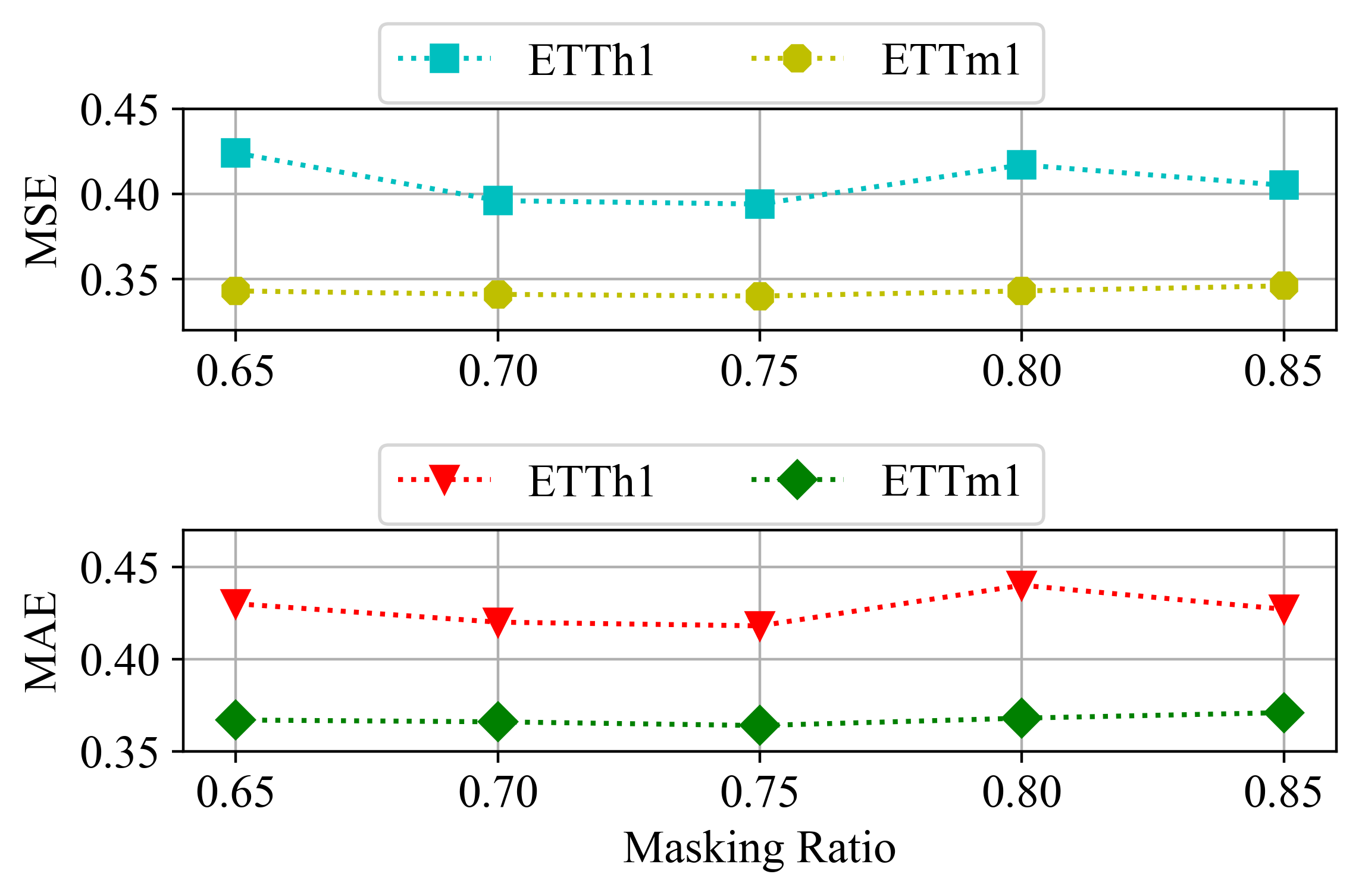}
	\end{center}
	\caption{The impact of different masking ratios on ETTh1 and ETTm1 datasets.}
	\label{fig4}
\end{figure}

\subsection{Discussion}
A recent work DLinear \cite{DLinear} challenges the effectiveness of the booming Transformers for the time series forecasting task. Because it surprisingly outperforms existing complex architectures by a large margin via only a single linear layer. From our perspective, although the Transformer-based methods make the model architecture more compatible for time series, they are still prone to be overfitting due to less inductive bias, leading to worse performance compared to MLP-based methods. However, with the help of pre-training, our model takes the overall distribution of datasets into consideration to alleviate the overfitting problem, which outperforms DLinear by a large margin. Also, according to the attention maps in Appendix D, we find that the pre-trained model can effectively capture the relationship between the input sequences and forecasting sequences.  Hence, it is deserved for researchers to perform more theoretical analysis to bridge the gap between the Transformer-based and MLP-based methods in the future.

\begin{table}[!t]
\centering
\begin{tabular}{c|c|c|c}
\toprule[1.5pt]
\multirow{2}{*}{Models} & Parameter & Memory & Running Time \\
& (MB) & (MiB) & (s/iter) \\
\midrule[1pt]
Cross-MAE & 0.183 & 1255 & 0.0196 \\
\midrule[1pt]
SimMTM & 1.812 & 1358 & 0.0177 \\
\midrule[1pt]
Ti-MAE & 1.812 & 1358 & 0.0177 \\
\midrule[1pt]
PatchTST & 0.359 & 1264 & 0.0184 \\
\midrule[1pt]
DLinear & 0.226 & 1241 & 0.0182 \\
\bottomrule[1.5pt]
\end{tabular}
\caption{Comparison of average model efficiency based on the forecasting task of ETTh1.}
\label{table1}
\end{table}

\section{Conclusion}
In this work, we point out the unification of task objectives and adaptation for task difficulty are critical for bridging the gap between time series masked reconstruction and forecasting tasks. To mitigate the inconsistency problem, we propose a simple yet effective prompt token tuning (PT-Tuning) paradigm, in which all pre-trained parameters are frozen and only a few trainable prompt tokens are added to extended mask tokens in element-wise manner. Extensive experiments on real-world datasets illustrate that our paradigm can achieve state-of-the-art performances compared to representation learning and end-to-end methods. However, our paradigm only support time series forecasting task and limits the scope of application. So we will extend PT-Tuning into a task-general paradigm for time series analysis, such as imputation, anomaly detection and so on.

\bibliography{aaai24}

\clearpage
\appendix
\section{Implementation Details}

\subsection{Datasets}
We conduct experiments on seven public benchmark datasets. (1) \textit{ETT}\footnote{https://github.com/zhouhaoyi/ETDataset} (Electricity Transformer Temperature) are collected from two different electric transformers with different resolutions, including two hourly-level datasets (ETTh1, ETTh2) and two 15-minute-level datasets (ETTm1, ETTm2), recording six power load features and oil temperature. (2) \textit{Weather}\footnote{https://www.bgc-jena.mpg.de/wetter/} dataset is recorded every 10 minutes for 2020 whole year from about 1,600 U.S. places, containing 21 meteorological indicators, such air temperature, humidity, etc. (3) \textit{Electricity}\footnote{https://archive.ics.uci.edu/ml/datasets} is a dataset that describes 321 customers' hourly electricity consumption. (4) \textit{Traffic}\footnote{https://pems.dot.ca.gov/} dataset records the road occupancy rates from different sensors on San Francisco freeways. We follow the same protocol as previous methods and split all datasets into training, validation and test set by the ratio of 6:2:2 for the ETT dataset and 7:1:2 for other datasets. 

\subsection{Experimental Settings}

We provide the model configurations in Table 7. All experiments are repeated three times, implemented in Pytorch and conducted on a single NVIDIA Tesla V100 32GB GPU. The Adam optimizer is adopted with the default hyperparameter configuration for (${\beta_1}$,${\beta_2}$) as (0.9,0.999). And all the training processes are early stopped within specific epochs. For the metrics, we also adopt the mean squared error (MSE) and mean absolute error (MAE) as previous methods. A lower MSE or MAE indicates a better prediction.

\begin{table*}[!t]\footnotesize
\centering
\renewcommand\arraystretch{1.3}
\resizebox{\linewidth}{!}{
\begin{tabular}{c|c|c|c|c|c|c|c|c|c|c|c|c}
\toprule[1.5pt]
Dataset & \#encoder & \#decoder & patch & model dim & \#head & FFN dim & dropout & LR & batch size & PT epoch & FT epoch & Ratio \\
\midrule[1pt]
ETTm1 & 2 & 2 & 8 & 64 & 4 & 256 & 0.05 & 0.001 & 32 & 20 & 20 & 0.75 \\
ETTm2 & 2 & 2 & 8 & 64 & 4 & 256 & 0.05 & 0.001 & 32 & 20 & 20 & 0.75 \\
ETTh1 & 2 & 2 & 8 & 64 & 4 & 256 & 0.05 & 0.001 & 32 & 20 & 20 & 0.75 \\
ETTh2 & 2 & 2 & 8 & 64 & 4 & 256 & 0.05 & 0.001 & 32 & 100 & 20 & 0.75 \\
Weather & 2 & 2 & 8 & 64 & 4 & 256 & 0.05 & 0.001 & 32 & 20 & 20 & 0.75 \\
Electricity & 2 & 2 & 8 & 64 & 4 & 256 & 0.05 & 0.001 & 32 & 100 & 20 & 0.75 \\
Traffic & 2 & 2 & 8 & 64 & 4 & 256 & 0.05 & 0.001 & 32 & 100 & 20 & 0.75 \\

\bottomrule[1.5pt]
\end{tabular}}
\caption{Experiment configurations of proposed PT-Tuning.}
\label{table1}
\end{table*}

\section{Ablation Stuides}

In this section, we report the full results of all ablation studies in the main text. And more experiments are conducted to verify the hyperparameter sensitivity of our proposed method on ETTh1 and ETTm1 datasets, including pre-training length, look-back window and patch size.

\subsection{Full Results}
Due to the space limitation of the main text, we place all the full results of ablation studies in the following: model generality in Table 8, prompt strategy in Table 9, masking strategy in Table 10 and masking ratio in Table 11.

\begin{table*}[!t]\footnotesize
\centering
\renewcommand\arraystretch{1.3}
\resizebox{\linewidth}{!}{
\begin{tabular}{c|c|cc|cc|cc|cc|cc|cc|cc|cc}
\toprule[1.5pt]
\multicolumn{2}{c|}{Dataset} &  \multicolumn{8}{c|}{ETTh1}  &  \multicolumn{8}{c}{ETTm1}\\
\midrule[1pt]
\multicolumn{2}{c|}{Length} & \multicolumn{2}{c|}{96}  &  \multicolumn{2}{c|}{192} & \multicolumn{2}{c|}{336}  &  \multicolumn{2}{c|}{720} & \multicolumn{2}{c|}{96}  &  \multicolumn{2}{c|}{192} & \multicolumn{2}{c|}{336}  &  \multicolumn{2}{c}{720} \\
\midrule[1pt]
\multicolumn{2}{c|}{Metric} & MSE & MAE & MSE & MAE & MSE & MAE & MSE & MAE & MSE & MAE & MSE & MAE & MSE & MAE & MSE & MAE \\
\midrule[1pt]
\multirow{2}{*}{BEiT-Style} & FT & 0.408 & 0.426 & 0.442 & 0.444 & 0.444 & 0.454 & 0.501 & 0.493 & 0.329 & 0.368 & 0.362 & 0.384 & 0.391 & 0.400 & 0.436 & 0.425 \\
& PT-T & \textbf{0.391} & \textbf{0.417} & \textbf{0.414} & \textbf{0.431} & \textbf{0.433} & \textbf{0.444} & \textbf{0.455} & \textbf{0.461} & \textbf{0.311} & \textbf{0.356} & \textbf{0.343} & \textbf{0.375} & \textbf{0.367} & \textbf{0.392} & \textbf{0.417} & \textbf{0.421} \\
\midrule[1pt]
\multirow{2}{*}{MAE-Style} & FT & \textbf{0.373} & \textbf{0.402} & 0.423 & 0.435 & 0.444 & 0.452 & 0.558 & 0.528 & 0.303 & 0.357 & 0.340 & 0.378 & 0.363 & 0.389 & \textbf{0.414} & 0.420  \\
& PT-T & 0.376 & 0.408 & \textbf{0.401} & \textbf{0.426} & \textbf{0.421} & \textbf{0.444} & \textbf{0.474} & \textbf{0.484} & \textbf{0.298} & \textbf{0.343} & \textbf{0.335} & \textbf{0.366} & \textbf{0.362} & \textbf{0.385} & \textbf{0.414} & \textbf{0.416} \\
\midrule[1pt]
\multirow{2}{*}{Cross-MAE} & FT & 0.378 & 0.406 & 0.410 & 0.426 & 0.446 & 0.460 & 0.489 & 0.490 & 0.303 & 0.360 & 0.336 & 0.378 & 0.360 & 0.389 & 0.412 & 0.419  \\
&PT-T & \textbf{0.348} & \textbf{0.388} & \textbf{0.382} & \textbf{0.410} & \textbf{0.406} & \textbf{0.423} & \textbf{0.442} & \textbf{0.451} & \textbf{0.281} & \textbf{0.327} & \textbf{0.322} & \textbf{0.352} & \textbf{0.350} & \textbf{0.372} & \textbf{0.407} & \textbf{0.406}  \\
\bottomrule[1.5pt]
\end{tabular}}
\caption{Full results by applying PT-Tuning to three pretraining frameworks.}
\label{table1}
\end{table*}

\begin{table*}[!t]\footnotesize
\centering
\renewcommand\arraystretch{1.3}
\resizebox{\linewidth}{!}{
\begin{tabular}{c|c|cc|cc|cc|cc|cc|cc|cc|cc}
\toprule[1.5pt]
\multicolumn{2}{c|}{Dataset} &  \multicolumn{8}{c|}{ETTh1}  &  \multicolumn{8}{c}{ETTm1}\\
\midrule[1pt]
\multicolumn{2}{c|}{Length} & \multicolumn{2}{c|}{96}  &  \multicolumn{2}{c|}{192} & \multicolumn{2}{c|}{336}  &  \multicolumn{2}{c|}{720} & \multicolumn{2}{c|}{96}  &  \multicolumn{2}{c|}{192} & \multicolumn{2}{c|}{336}  &  \multicolumn{2}{c}{720} \\
\midrule[1pt]
\multicolumn{2}{c|}{Metric} & MSE & MAE & MSE & MAE & MSE & MAE & MSE & MAE & MSE & MAE & MSE & MAE & MSE & MAE & MSE & MAE \\
\midrule[1pt]
Prompt-C & History & 0.363 & 0.396 & 0.394 & 0.417 & 0.455 & 0.450 & 0.583 & 0.539 & 0.280 & 0.333 & 0.320 & 0.357 & 0.355 & 0.380 & 0.428 & 0.421 \\
\midrule[1pt]
\multirow{2}{*}{Prompt-A} & History & 0.351 & \textbf{0.386} & 0.388 & \textbf{0.410} & 0.424 & 0.434 & 0.554 & 0.524 & \textbf{0.279} & \textbf{0.327} & \textbf{0.322} & 0.355 & 0.358 & 0.380 & 0.422 & 0.418  \\
&Future & \textbf{0.348} & 0.388 & \textbf{0.382} & \textbf{0.410} & \textbf{0.406} & \textbf{0.423} & \textbf{0.442} & \textbf{0.451} & 0.281 & \textbf{0.327} & \textbf{0.322} & \textbf{0.352} & \textbf{0.350} & \textbf{0.372} & \textbf{0.407} & \textbf{0.406}  \\
\bottomrule[1.5pt]
\end{tabular}}
\caption{Full results of different prompt strategies on ETTh1 and ETTm1.}
\label{table1}
\end{table*}

\begin{table*}[!t]\footnotesize
\centering
\renewcommand\arraystretch{1.3}
\resizebox{\linewidth}{!}{
\begin{tabular}{c|cc|cc|cc|cc|cc|cc|cc|cc}
\toprule[1.5pt]
Dataset &  \multicolumn{8}{c|}{ETTh1}  &  \multicolumn{8}{c}{ETTm1}\\
\midrule[1pt]
Length & \multicolumn{2}{c|}{96}  &  \multicolumn{2}{c|}{192} & \multicolumn{2}{c|}{336}  &  \multicolumn{2}{c|}{720} & \multicolumn{2}{c|}{96}  &  \multicolumn{2}{c|}{192} & \multicolumn{2}{c|}{336}  &  \multicolumn{2}{c}{720} \\
\midrule[1pt]
Metric & MSE & MAE & MSE & MAE & MSE & MAE & MSE & MAE & MSE & MAE & MSE & MAE & MSE & MAE & MSE & MAE \\
\midrule[1pt]
Periodic & 0.533 & 0.475 & 0.565 & 0.498 & 0.579 & 0.515 & 0.599 & 0.550 & 0.346 & 0.377 & 0.376 & 0.397 & 0.401 & 0.413 & 0.448 & 0.441 \\
\midrule[1pt]
Continuous & 0.422 & 0.428 & 0.450 & 0.444 & 0.461 & 0.454 & 0.493 & 0.480 & 0.344 & 0.379 & 0.371 & 0.393 & 0.391 & 0.406 & 0.436 & 0.431 \\
\midrule[1pt]
Random & 0.393 & 0.421 & 0.422 & 0.438 & 0.442 & 0.451 & 0.497 & 0.486 & 0.284 & 0.328 & 0.324 & 0.353 & 0.353 & 0.374 & 0.411 & 0.407 \\
\midrule[1pt]
Isometric & \textbf{0.348} & \textbf{0.388} & \textbf{0.382} & \textbf{0.410} & \textbf{0.406} & \textbf{0.423} & \textbf{0.442} & \textbf{0.451} & \textbf{0.281} & \textbf{0.327} & \textbf{0.322} & \textbf{0.352} & \textbf{0.350} & \textbf{0.372} & \textbf{0.407} & \textbf{0.406}  \\
\bottomrule[1.5pt]
\end{tabular}}
\caption{Full results of different masking strategies with 75\% ratio on ETTh1 and ETTm1 datasets.}
\label{table1}
\end{table*}

\begin{table*}[!t]\footnotesize
\centering
\renewcommand\arraystretch{1.3}
\resizebox{\linewidth}{!}{
\begin{tabular}{c|cc|cc|cc|cc|cc|cc|cc|cc}
\toprule[1.5pt]
Dataset &  \multicolumn{8}{c|}{ETTh1}  &  \multicolumn{8}{c}{ETTm1}\\
\midrule[1pt]
Length & \multicolumn{2}{c|}{96}  &  \multicolumn{2}{c|}{192} & \multicolumn{2}{c|}{336}  &  \multicolumn{2}{c|}{720} & \multicolumn{2}{c|}{96}  &  \multicolumn{2}{c|}{192} & \multicolumn{2}{c|}{336}  &  \multicolumn{2}{c}{720} \\
\midrule[1pt]
Metric & MSE & MAE & MSE & MAE & MSE & MAE & MSE & MAE & MSE & MAE & MSE & MAE & MSE & MAE & MSE & MAE \\
\midrule[1pt]
0.65 & 0.378 & 0.399 & 0.413 & 0.420 & 0.434 & 0.434 & 0.470 & 0.466 & 0.285 & 0.331 & 0.326 & 0.355 & 0.353 & 0.375 & 0.408 & 0.408 \\
\midrule[1pt]
0.7 & 0.352 & 0.390 & 0.383 & 0.412 & \textbf{0.403} & 0.424 & 0.445 & 0.453 & 0.282 & 0.329 & 0.323 & 0.353 & 0.351 & 0.374 & 0.409 & 0.409 \\
\midrule[1pt]
0.75 & \textbf{0.348} & \textbf{0.388} & \textbf{0.382} & \textbf{0.410} & 0.406 & \textbf{0.423} & \textbf{0.442} & \textbf{0.451} & \textbf{0.281} & \textbf{0.327} & \textbf{0.322} & \textbf{0.352} & \textbf{0.350} & \textbf{0.372} & \textbf{0.407} & \textbf{0.406}  \\
\midrule[1pt]
0.8 & 0.383 & 0.415 & 0.409 & 0.433 & 0.422 & 0.443 & 0.456 & 0.467 & 0.288 & 0.332 & 0.325 & 0.356 & 0.352 & 0.376 & 0.408 & 0.409 \\
\midrule[1pt]
0.85 & 0.366 & 0.401 & 0.394 & 0.421 & 0.408 & 0.431 & 0.451 & 0.456 & 0.288 & 0.335 & 0.327 & 0.359 & 0.355 & 0.378 & 0.413 & 0.412 \\
\bottomrule[1.5pt]
\end{tabular}}
\caption{Full results of different masking strategies with 75\% ratio on ETTh1 and ETTm1 datasets.}
\label{table1}
\end{table*}

\subsection{Pre-training Length}

We compare different lengths of input sequence for the pre-training stage in Table 12. It can be observed that the performance of our model is significantly improved with the increase of per-training lengths because longer sequences can introduce more long-range contextual information. Notably, too long pre-training sequences dose not bring too much information and increases the computational burden. So the length around 720 balances the tradeoff more appropriate, which works best for our paradigm.

\begin{table*}[!t]\footnotesize
\centering
\renewcommand\arraystretch{1.3}
\resizebox{\linewidth}{!}{
\begin{tabular}{c|cc|cc|cc|cc|cc|cc|cc|cc}
\toprule[1.5pt]
Dataset &  \multicolumn{8}{c|}{ETTh1}  &  \multicolumn{8}{c}{ETTm1}\\
\midrule[1pt]
Length & \multicolumn{2}{c|}{96}  &  \multicolumn{2}{c|}{192} & \multicolumn{2}{c|}{336}  &  \multicolumn{2}{c|}{720} & \multicolumn{2}{c|}{96}  &  \multicolumn{2}{c|}{192} & \multicolumn{2}{c|}{336}  &  \multicolumn{2}{c}{720} \\
\midrule[1pt]
Metric & MSE & MAE & MSE & MAE & MSE & MAE & MSE & MAE & MSE & MAE & MSE & MAE & MSE & MAE & MSE & MAE \\
\midrule[1pt]
336 & 0.424 & 0.432 & 0.439 & 0.442 & 0.448 & 0.451 & 0.452 & 0.463 & 0.357 & 0.388 & 0.379 & 0.397 & 0.402 & 0.408 & 0.447 & 0.432 \\
\midrule[1pt]
512 & 0.355 & 0.392 & 0.385 & 0.412 & \textbf{0.403} & \textbf{0.423} & \textbf{0.439} & \textbf{0.444} & 0.292 & 0.337 & 0.328 & 0.359 & 0.357 & 0.378 & 0.416 & 0.411 \\
\midrule[1pt]
720 & \textbf{0.348} & 0.388 & \textbf{0.382} & 0.410 & 0.406 & \textbf{0.423} & 0.442 & 0.451 & \textbf{0.281} & \textbf{0.327} & \textbf{0.322} & \textbf{0.352} & \textbf{0.350} & \textbf{0.372} & \textbf{0.407} & \textbf{0.406}  \\
\midrule[1pt]
912 & 0.350 & \textbf{0.387} & 0.386 & \textbf{0.409} & 0.412 & \textbf{0.423} & 0.462 & 0.454 & 0.286 & 0.334 & 0.327 & 0.359 & 0.355 & 0.379 & 0.409 & 0.412 \\
\bottomrule[1.5pt]
\end{tabular}}
\caption{Full results of different pre-training lengths with 75\% ratio on ETTh1 and ETTm1 datasets.}
\label{table1}
\end{table*}

\subsection{Look-back Window}

The size of the look-back window has a high impact on forecasting performance because it determines how much model can learn from historical data. To study the impact of look-back window sizes, we conduct experiments with different window sizes \{96,192,336,512,720\} to forecast different horizons \{96,192,336,720\}. As shown in Table 13, the performance of PT-Tuning is significantly boosted with the increase of look-back window size, which proves the effectiveness of our model in learning temproal information from longer receptive field. Notably, the correlation of future information to far-history observations is usually smaller than that to near-history observations, implying that far-history observations have higher variance. So it can be observed that the performance achieves the best when the size of look-back window is 512 and starts to decrease under the setting of 720 look-back window.

\begin{table*}[!t]\footnotesize
\centering
\renewcommand\arraystretch{1.3}
\resizebox{\linewidth}{!}{
\begin{tabular}{c|cc|cc|cc|cc|cc|cc|cc|cc}
\toprule[1.5pt]
Dataset &  \multicolumn{8}{c|}{ETTh1}  &  \multicolumn{8}{c}{ETTm1}\\
\midrule[1pt]
Length & \multicolumn{2}{c|}{96}  &  \multicolumn{2}{c|}{192} & \multicolumn{2}{c|}{336}  &  \multicolumn{2}{c|}{720} & \multicolumn{2}{c|}{96}  &  \multicolumn{2}{c|}{192} & \multicolumn{2}{c|}{336}  &  \multicolumn{2}{c}{720} \\
\midrule[1pt]
Metric & MSE & MAE & MSE & MAE & MSE & MAE & MSE & MAE & MSE & MAE & MSE & MAE & MSE & MAE & MSE & MAE \\
\midrule[1pt]
96 & 0.375 & 0.394 & 0.428 & 0.424 & 0.470 & 0.444 & 0.477 & 0.466 & 0.332 & 0.352 & 0.368 & 0.374 & 0.400 & 0.396 & 0.465 & 0.432 \\
\midrule[1pt]
192 & 0.368 & 0.392 & 0.415 & 0.419 & 0.447 & 0.434 & 0.455 & 0.458 & 0.296 & 0.332 & 0.331 & 0.357 & 0.363 & 0.379 & 0.429 & 0.416 \\
\midrule[1pt]
336 & 0.363 & 0.394 & 0.401 & 0.416 & 0.424 & 0.428 & 0.440 & 0.451 & 0.283 & 0.330 & 0.323 & 0.356 & 0.356 & 0.377 & 0.419 & 0.411 \\
\midrule[1pt]
512 & \textbf{0.348} & \textbf{0.388} & \textbf{0.382} & \textbf{0.410} & \textbf{0.406} & \textbf{0.423} & \textbf{0.442} & \textbf{0.451} & \textbf{0.281} & \textbf{0.327} & \textbf{0.322} & \textbf{0.352} & \textbf{0.350} & \textbf{0.372} & \textbf{0.407} & \textbf{0.406}  \\
\midrule[1pt]
720 & 0.356 & 0.397 & 0.388 & 0.416 & 0.411 & 0.429 & 0.453 & 0.460 & 0.289 & 0.335 & 0.323 & 0.358 & 0.351 & 0.377 & \textbf{0.407} & 0.408 \\
\bottomrule[1.5pt]
\end{tabular}}
\caption{Full results of different look-back windows with 75\% ratio on ETTh1 and ETTm1 datasets.}
\label{table1}
\end{table*}

\subsection{Patch Size}

To study the effect of patch sizes to the forecasting performance, we fix the look-back window to be 512 and vary the patch sizes with \{4,6,8,12,16\}. As shown in Table 14, MSE and MAE scores vary significantly with different choices of patch size on ETTh1 dataset while ETTm1 dataset is opposite, which indicates the importance of suitable patch size for different datasets. Overall, PT-Tuning benefits from increased patch length, not only in forecasting performance but also in the computational reduction. And it achieves the best balance when the patch size is 8.

\begin{table*}[!t]\footnotesize
\centering
\renewcommand\arraystretch{1.3}
\resizebox{\linewidth}{!}{
\begin{tabular}{c|cc|cc|cc|cc|cc|cc|cc|cc}
\toprule[1.5pt]
Dataset &  \multicolumn{8}{c|}{ETTh1}  &  \multicolumn{8}{c}{ETTm1}\\
\midrule[1pt]
Length & \multicolumn{2}{c|}{96}  &  \multicolumn{2}{c|}{192} & \multicolumn{2}{c|}{336}  &  \multicolumn{2}{c|}{720} & \multicolumn{2}{c|}{96}  &  \multicolumn{2}{c|}{192} & \multicolumn{2}{c|}{336}  &  \multicolumn{2}{c}{720} \\
\midrule[1pt]
Metric & MSE & MAE & MSE & MAE & MSE & MAE & MSE & MAE & MSE & MAE & MSE & MAE & MSE & MAE & MSE & MAE \\
\midrule[1pt]
4 & 0.353 & 0.393 & 0.390 & 0.419 & 0.414 & 0.436 & 0.474 & 0.468 & 0.295 & 0.337 & 0.334 & 0.364 & 0.366 & 0.384 & 0.423 & 0.417 \\
\midrule[1pt]
6 & 0.440 & 0.443 & 0.467 & 0.460 & 0.487 & 0.471 & 0.531 & 0.504 & 0.291 & 0.330 & 0.335 & 0.357 & 0.361 & 0.377 & 0.424 & 0.411 \\
\midrule[1pt]
8 & \textbf{0.348} & \textbf{0.388} & \textbf{0.382} & \textbf{0.410} & \textbf{0.406} & \textbf{0.423} & \textbf{0.442} & \textbf{0.451} & \textbf{0.281} & \textbf{0.327} & \textbf{0.322} & \textbf{0.352} & \textbf{0.350} & \textbf{0.372} & \textbf{0.407} & \textbf{0.406}  \\
\midrule[1pt]
12 & 0.720 & 0.555 & 0.734 & 0.568 & 0.732 & 0.575 & 0.766 & 0.608 & 0.321 & 0.357 & 0.353 & 0.378 & 0.375 & 0.394 & 0.426 & 0.423 \\
\midrule[1pt]
16 & 0.396 & 0.418 & 0.427 & 0.438 & 0.448 & 0.453 & 0.490 & 0.486 & 0.289 & 0.334 & 0.330 & 0.360 & 0.357 & 0.379 & 0.417 & 0.412 \\
\bottomrule[1.5pt]
\end{tabular}}
\caption{Full results of different patch sizes with 75\% ratio on ETTh1 and ETTm1 datasets.}
\label{table1}
\end{table*}

\section{Comparison}

We provide the full results in the following: multivariate time series forecasting results compared with representation learning methods in Table 15, forecasting results compared with end-to-end methods in Table 16. And our state-of-the-art results are based on the 512 look-back window. For representation learning methods, the full results of SimMTM \cite{SimMTM}, Ti-MAE \cite{Ti-MAE}, TST \cite{George-2021}, LaST \cite{LaST}, TF-C \cite{TF-C}, CoST \cite{Woo-2022} and TS2Vec \cite{Zhihan-2021} are originated from \cite{SimMTM}. For end-to-end methods, the full results of PatchTST, DLinear are from \cite{PatchTST}, TimesNet \cite{TimesNet}, FEDformemr \cite{tian-2021} and Stationary \cite{Non-Stationary} are from \cite{TimesNet}, and MICN \cite{MICN} and Crossformer \cite{Crossformer} are based on our reproduction.

\begin{table*}\footnotesize
\centering
\renewcommand\arraystretch{1.3}
\resizebox{\linewidth}{!}{
\begin{tabular}{c|c|cc|cc|cc|cc|cc|cc|cc|cc}
\toprule
\multicolumn{2}{c|}{Method} &  \multicolumn{2}{c|}{PT-Tuning}  &  \multicolumn{2}{c|}{SimMTM}  & \multicolumn{2}{c|}{Ti-MAE}  &  \multicolumn{2}{c|}{TST}  &  \multicolumn{2}{c|}{LaST} & \multicolumn{2}{c|}{TF-C} & \multicolumn{2}{c|}{CoST} &
\multicolumn{2}{c}{TS2Vec} \\
\midrule
\multicolumn{2}{c|}{Metric} &  MSE & MAE &  MSE & MAE  &  MSE & MAE  &  MSE & MAE   &  MSE & MAE & MSE & MAE  & MSE & MAE & MSE & MAE \\
\midrule
\multirow{5}{*}{\rotatebox{90}{ETTm1}} & 96 & \textbf{0.281} & \textbf{0.327} & \underline{0.289} & \underline{0.343} & 0.305 & 0.351 & 0.310 & 0.348 & 0.316 & 0.355 & 0.671 & 0.601 & 0.291 & \underline{0.343} & 0.681 & 0.689 \\
 & 192 & \textbf{0.322} & \textbf{0.352} & \underline{0.323} & 0.369 & 0.343 & 0.374 & 0.362 & 0.380 & 0.349 & \underline{0.366} & 0.719 & 0.638 & 0.330 & 0.370 & 0.689 & 0.551 \\
 & 336 & \underline{0.350} & \textbf{0.372} & \textbf{0.349} & \underline{0.385} & 0.387 & 0.407 & 0.389 & 0.402 & 0.429 & 0.407 & 0.743 & 0.659 & 0.382 & 0.401 & 0.704 & 0.559 \\
 & 720 & \underline{0.407} & \textbf{0.406} & \textbf{0.399} & \underline{0.418} & 0.428 & 0.432 & 0.433 & 0.427 & 0.496 & 0.464 & 0.842 & 0.708 & 0.422 & 0.425 & 0.721 & 0.571 \\
\cline{2-18}
 & Avg & \textbf{0.340} & \textbf{0.364} & \textbf{0.340} & \underline{0.379} & 0.366 & 0.391 & 0.373 & 0.389 & 0.398 & 0.398 & 0.744 & 0.652 & 0.356 & 0.385 & 0.699 & 0.557 \\
\hline
\multirow{5}{*}{\rotatebox{90}{ETTm2}} & 96 & \underline{0.162} & \textbf{0.246} & 0.166 & 0.257 & 0.173 & 0.259 & 0.215 & 0.296 & \textbf{0.160} & \underline{0.254} & 0.401 & 0.490 & 0.242 & 0.333 & 0.224 & 0.303 \\
 & 192 & \textbf{0.218} & \textbf{0.284} & \underline{0.223} & \underline{0.295} & 0.259 & 0.301 & 0.259 & 0.323 & 0.225 & 0.300 & 0.822 & 0.677 & 0.283 & 0.345 & 0.273 & 0.331 \\
 & 336 & \underline{0.271} & \underline{0.321} & 0.282 & 0.334 & 0.272 & \textbf{0.319} & 0.319 & 0.364 & \textbf{0.239} & 0.366 & 1.214 & 0.908 & 0.303 & 0.349 & 0.399 & 0.402 \\
 & 720 & \underline{0.353} & \textbf{0.374} & 0.370 & 0.385 & \textbf{0.351} & 0.400 & 0.395 & 0.405 & 0.397 & \underline{0.382} & 4.584 & 1.711 & 0.431 & 0.431 & 0.406 & 0.408 \\
\cline{2-18}
 & Avg & \textbf{0.251} & \textbf{0.306} & 0.260 & \underline{0.318} & 0.264 & 0.320 & 0.297 & 0.347 & \underline{0.255} & 0.326 & 1.755 & 0.947 & 0.314 & 0.365 & 0.326 & 0.361 \\
 \hline
\multirow{5}{*}{\rotatebox{90}{ETTh1}} & 96  & \textbf{0.348}  & \textbf{0.388} & 0.367 & \underline{0.402} & \underline{0.356} & 0.420 & 0.401 & 0.425 & 0.399 & 0.412 & 0.665 & 0.604 & 0.422 & 0.436 & 0.392 & 0.420 \\
 & 192 & \textbf{0.382} & \textbf{0.410} & \underline{0.403} & \underline{0.425} & 0.421 & 0.434 & 0.427 & 0.432 & 0.484 & 0.468 & 0.630 & 0.640 & 0.520 & 0.487 & 0.445 & 0.452 \\
 & 336 & \textbf{0.406} & \textbf{0.423} & \underline{0.415} & \underline{0.430} & 0.447 & 0.446 & 0.519 & 0.487 & 0.580 & 0.533 & 0.605 & 0.645 & 0.472 & 0.462 & 0.453 & 0.455 \\
 & 720 & \underline{0.442} & \textbf{0.451} & \textbf{0.430} & \underline{0.453} & 0.469 & 0.482 & 0.515 & 0.504 & 0.432 & 0.432 & 0.647 & 0.662 & 0.525 & 0.501 & 0.495 & 0.496 \\
\cline{2-18}
 & Avg & \textbf{0.394} & \textbf{0.418} & \underline{0.404} & \underline{0.428} & 0.423 & 0.446 & 0.466 & 0.462 & 0.474 & 0.461 & 0.637 & 0.638 & 0.485 & 0.472 & 0.446 & 0.456 \\
\hline
\multirow{5}{*}{\rotatebox{90}{ETTh2}} & 96 & \textbf{0.274} & \textbf{0.333} & \underline{0.288} & \underline{0.347} & 0.339 & 0.378 & 0.322 & 0.358 & 0.331 & 0.390 & 1.663 & 1.021 & 0.321 & 0.374 & 0.365 & 0.509 \\
 & 192 & \textbf{0.336} & \textbf{0.374} & \underline{0.346} & \underline{0.385} & 0.380 & 0.402 & 0.448 & 0.435 & 0.751 & 0.612 & 3.525 & 1.561 & 0.380 & 0.403 & 0.396 & 0.422  \\
 & 336 & \textbf{0.354} & \textbf{0.394} & \underline{0.363} & \underline{0.401} & 0.388 & 0.423 & 0.420 & 0.440 & 0.460 & 0.478 & 3.283 & 1.500 & 0.430 & 0.451 & 0.399 & 0.436 \\
 & 720 & \textbf{0.384} & \textbf{0.426} & \underline{0.396} & \underline{0.431} & 0.414 & 0.442 & 0.424 & 0.452 & 0.552 & 0.509 & 2.930 & 1.316 & 0.466 & 0.480 & 0.508 & 0.503 \\
\cline{2-18}
 & Avg & \textbf{0.337} & \textbf{0.381} & \underline{0.348} & \underline{0.391} & 0.380 & 0.386 & 0.404 & 0.421 & 0.499 & 0.497 & 2.850 & 1.349 & 0.399 & 0.427 & 0.417 & 0.468 \\
\hline
\multirow{5}{*}{\rotatebox{90}{Weather}} & 96 & 0.155 & \underline{0.201} & \underline{0.151} & 0.202 & \textbf{0.143} & \textbf{0.195} & 0.162 & 0.214 & 0.153 & 0.211 & 0.215 & 0.296 & 0.216 & 0.280 & 0.154 & 0.205 \\
 & 192 & \underline{0.197} & \textbf{0.240} & \textbf{0.195} & \underline{0.243} & 0.204 & 0.253 & 0.203 & 0.252 & 0.207 & 0.250 & 0.267 & 0.345 & 0.303 & 0.335 & 0.200 & \underline{0.243} \\
 & 336 & \textbf{0.243} & 0.277 & \underline{0.246} & 0.283 & 0.241 & \underline{0.274} & 0.260 & 0.297 & 0.249 & \textbf{0.264} & 0.299 & 0.360 & 0.351 & 0.358 & 0.252 & 0.286 \\
 & 720 & \textbf{0.313} & \underline{0.328} & 0.320 & 0.338 & 0.322 & 0.339 & 0.330 & 0.342 & \underline{0.319} & \textbf{0.320} & 0.361 & 0.395 & 0.425 & 0.343 & 0.324 & 0.335 \\
\cline{2-18}
 & Avg & \textbf{0.227} & \textbf{0.261} & \underline{0.228} & 0.267 & 0.230 & 0.265 & 0.239 & 0.276 & 0.232 & \textbf{0.261} & 0.286 & 0.349 & 0.324 & 0.329 & 0.233 & 0.267 \\
\hline
\multirow{5}{*}{\rotatebox{90}{Electricity}} & 96 & \textbf{0.128} & \textbf{0.221} & \underline{0.133} & \underline{0.223} & 0.163 & 0.255 & 0.186 & 0.268 & 0.166 & 0.254 & 0.366 & 0.436 & 0.197 & 0.277 & 0.195 & 0.275 \\
 & 192 & \textbf{0.144} & \textbf{0.235} & \underline{0.147} & \underline{0.237} & 0.194 & 0.288 & 0.193 & 0.276 & 0.178 & 0.278 & 0.366 & 0.433 & 0.197 & 0.279 & 0.195 & 0.277 \\
 & 336 & \textbf{0.160} & \textbf{0.252} & \underline{0.166} & \underline{0.265} & 0.201 & 0.298 & 0.206 & 0.289 & 0.186 & 0.275 & 0.358 & 0.428 & 0.211 & 0.295 & 0.210 & 0.294 \\
 & 720 & \textbf{0.199} & \textbf{0.285} & \underline{0.203} & 0.297 & 0.263 & 0.343 & 0.250 & 0.324 & 0.213 & \underline{0.288} & 0.363 & 0.431 & 0.255 & 0.330 & 0.252 & 0.327 \\
\cline{2-18}
 & Avg & \textbf{0.157} & \textbf{0.248} & \underline{0.162} & \underline{0.256} & 0.205 & 0.296 & 0.209 & 0.289 & 0.186 & 0.274 & 0.363 & 0.398 & 0.215 & 0.295 & 0.213 & 0.293 \\
\hline
\multirow{5}{*}{\rotatebox{90}{Traffic}} & 96 & \textbf{0.365} & \textbf{0.243} & \underline{0.368} & \underline{0.262} & 0.448 & 0.298 & 0.595 & 0.360 & 0.706 & 0.385 & 0.613 & 0.340 & 0.378 & 0.365 & 0.480 & 0.357 \\
 & 192 & 0.382 & \textbf{0.250} & \underline{0.373} & \underline{0.251} & 0.445 & 0.301 & 0.576 & 0.353 & 0.709 & 0.388 & 0.619 & 0.516 & \textbf{0.371} & 0.352 & 0.439 & 0.336 \\
 & 336 & \textbf{0.395} & \underline{0.257} & \textbf{0.395} & \textbf{0.254} & 0.492 & 0.320 & 0.569 & 0.362 & 0.714 & 0.394 & 0.785 & 0.497 & 0.467 & 0.354 & 0.460 & 0.344 \\
 & 720 & \textbf{0.431} & \textbf{0.277} & \underline{0.432} & \underline{0.290} & 0.514 & 0.321 & 0.603 & 0.372 & 0.723 & 0.421 & 0.850 & 0.472 & 0.525 & 0.378 & 0.499 & 0.364  \\
\cline{2-18}
 & Avg & \underline{0.393} & \textbf{0.256} & \textbf{0.392} & \underline{0.264} & 0.475 & 0.310 & 0.586 & 0.362 & 0.713 & 0.397 & 0.717 & 0.456 & 0.435 & 0.362 & 0.470 & 0.350 \\
\hline
\multicolumn{2}{c|}{${1^{st}}$ Count} &  \multicolumn{2}{c|}{\textbf{55}}  &  \multicolumn{2}{c|}{\underline{8}}  & \multicolumn{2}{c|}{4}  &  \multicolumn{2}{c|}{0}  &  \multicolumn{2}{c|}{5} & \multicolumn{2}{c|}{0} & \multicolumn{2}{c|}{1} &
\multicolumn{2}{c}{0} \\
\bottomrule
\end{tabular}}
\caption{Multivariate time series forecasting results compared with representation learning methods. We compare extensive competitive models under different prediction lengths. The best results are highlighted in bold and the second best results are highlighted with underline. The standard deviations of PT-Tuning are within 0.003 for MSE and within 0.002 for MAE.}
\label{table1}
\end{table*}

\begin{table*}\footnotesize
\centering
\renewcommand\arraystretch{1.3}
\resizebox{\linewidth}{!}{
\begin{tabular}{c|c|cc|cc|cc|cc|cc|cc|cc|cc}
\toprule
\multicolumn{2}{c|}{Method} &  \multicolumn{2}{c|}{PT-Tuning}  &  \multicolumn{2}{c|}{PatchTST}  & \multicolumn{2}{c|}{DLinear}  &  \multicolumn{2}{c|}{TimesNet}  &  \multicolumn{2}{c|}{MICN} & \multicolumn{2}{c|}{Crossformer} & \multicolumn{2}{c|}{FEDformer} &
\multicolumn{2}{c}{Stationary} \\
\midrule
\multicolumn{2}{c|}{Metric} &  MSE & MAE &  MSE & MAE  &  MSE & MAE  &  MSE & MAE   &  MSE & MAE & MSE & MAE  & MSE & MAE & MSE & MAE \\
\midrule
\multirow{5}{*}{\rotatebox{90}{ETTm1}} & 96 & \textbf{0.281} & \textbf{0.327} & \underline{0.293} & 0.346 & 0.299 & \underline{0.343} & 0.338 & 0.375 & 0.316 & 0.364 & 0.320 & 0.373 & 0.379 & 0.419 & 0.386 & 0.398 \\
 & 192 & \textbf{0.322} & \textbf{0.352} & \underline{0.333} & 0.370 & 0.335 & \underline{0.365} & 0.374 & 0.387 & 0.363 & 0.390 & 0.372 & 0.411 & 0.426 & 0.441 & 0.459 & 0.444 \\
 & 336 & \textbf{0.350} & \textbf{0.372} & 0.369 & \underline{0.392} & \underline{0.366} & \underline{0.392} & 0.410 & 0.411 & 0.408 & 0.426 & 0.429 & 0.441 & 0.445 & 0.459 & 0.495 & 0.464 \\
 & 720 & \textbf{0.407} & \textbf{0.406} & \underline{0.416} & \underline{0.420} & 0.425 & 0.421 & 0.478 & 0.450 & 0.459 & 0.464 & 0.573 & 0.531 & 0.543 & 0.490 & 0.585 & 0.516 \\
\cline{2-18}
 & Avg & \textbf{0.340} & \textbf{0.364} & \underline{0.353} & 0.382 & 0.356 & \underline{0.380} & 0.400 & 0.406 & 0.387 & 0.411 & 0.424 & 0.439 & 0.448 & 0.452 & 0.481 & 0.456 \\
\hline
\multirow{5}{*}{\rotatebox{90}{ETTm2}} & 96 & \textbf{0.162} & \textbf{0.246} & \underline{0.166} & \underline{0.256} & 0.167 & 0.260 & 0.187 & 0.267 & 0.179 & 0.275 & 0.254 & 0.348 & 0.203 & 0.287 & 0.192 & 0.274 \\
 & 192 & \textbf{0.218} & \textbf{0.284} & \underline{0.223} & \underline{0.296} & 0.224 & 0.303 & 0.249 & 0.309 & 0.262 & 0.326 & 0.370 & 0.433 & 0.269 & 0.328 & 0.280 & 0.339 \\
 & 336 & \textbf{0.271} & \textbf{0.321} & \underline{0.274} & \underline{0.329} & 0.281 & 0.342 & 0.321 & 0.351 & 0.305 & 0.353 & 0.511 & 0.527 & 0.325 & 0.366 & 0.334 & 0.361 \\
 & 720 & \textbf{0.353} & \textbf{0.374} & \underline{0.362} & \underline{0.385} & 0.397 & 0.421 & 0.408 & 0.403 & 0.389 & 0.407 & 0.901 & 0.689 & 0.421 & 0.415 & 0.417 & 0.413 \\
\cline{2-18}
 & Avg & \textbf{0.251} & \textbf{0.306} & \underline{0.256} & \underline{0.317} & 0.267 & 0.332 & 0.291 & 0.333 & 0.284 & 0.340 & 0.509 & 0.522 & 0.305 & 0.349 & 0.306 & 0.347 \\
 \hline
\multirow{5}{*}{\rotatebox{90}{ETTh1}} & 96  & \textbf{0.348}  & \textbf{0.388} & \underline{0.370} & 0.400 & 0.375 & \underline{0.399} & 0.384 & 0.402 & 0.398 & 0.427 & 0.377 & 0.419 & 0.376 & 0.419 & 0.513 & 0.491 \\
 & 192 & \textbf{0.382} & \textbf{0.410} & 0.413 & 0.429 & \underline{0.405} & \underline{0.416} & 0.436 & 0.429 & 0.430 & 0.453 & 0.410 & 0.439 & 0.420 & 0.448 & 0.534 & 0.504 \\
 & 336 & \textbf{0.406} & \textbf{0.423} & \underline{0.422} & \underline{0.440} & 0.439 & 0.443 & 0.491 & 0.469 & 0.440 & 0.460 & 0.440 & 0.461 & 0.459 & 0.465 & 0.588 & 0.535 \\
 & 720 & \textbf{0.442} & \textbf{0.451} & \underline{0.447} & \underline{0.468} & 0.472 & 0.490 & 0.521 & 0.500 & 0.491 & 0.509 & 0.519 & 0.524 & 0.506 & 0.507 & 0.643 & 0.616 \\
\cline{2-18}
 & Avg & \textbf{0.394} & \textbf{0.418} & \underline{0.413} & \underline{0.434} & 0.423 & 0.437 & 0.458 & 0.450 & 0.440 & 0.462 & 0.437 & 0.461 & 0.440 & 0.460 & 0.570 & 0.537 \\
\hline
\multirow{5}{*}{\rotatebox{90}{ETTh2}} & 96 & \textbf{0.274} & \textbf{0.333} & \textbf{0.274} & \underline{0.337} & 0.289 & 0.353 & 0.340 & 0.374 & 0.299 & 0.364 & 0.770 & 0.589 & 0.358 & 0.397 & 0.476 & 0.458 \\
 & 192 & \textbf{0.336} & \textbf{0.374} & \underline{0.341} & \underline{0.382} & 0.383 & 0.418 & 0.402 & 0.414 & 0.422 & 0.441 & 0.848 & 0.657 & 0.429 & 0.439 & 0.512 & 0.493  \\
 & 336 & \underline{0.354} & \underline{0.394} & \textbf{0.329} & \textbf{0.384} & 0.448 & 0.465 & 0.452 & 0.452 & 0.447 & 0.474 & 0.859 & 0.674 & 0.496 & 0.487 & 0.552 & 0.551 \\
 & 720 & \underline{0.384} & \underline{0.426} & \textbf{0.379} & \textbf{0.422} & 0.605 & 0.551 & 0.462 & 0.468 & 0.442 & 0.467 & 1.221 & 0.825 & 0.463 & 0.474 & 0.562 & 0.560 \\
\cline{2-18}
 & Avg & \underline{0.337} & \textbf{0.381} & \textbf{0.331} & \textbf{0.381} & 0.431 & 0.447 & 0.414 & 0.427 & 0.402 &  0.437 & 0.454 & 0.446 & 0.437 & 0.449 & 0.526 & 0.516 \\
\hline
\multirow{5}{*}{\rotatebox{90}{Weather}} & 96 & 0.155 & \underline{0.201} & \textbf{0.149} & \textbf{0.198} & 0.176 & 0.237 & 0.172 & 0.220 & 0.161 & 0.229 & 0.145 & 0.211 & 0.217 & 0.296 & 0.173 & 0.223 \\
 & 192 & \underline{0.197} & \textbf{0.240} & \textbf{0.194} & \underline{0.241} & 0.220 & 0.282 & 0.219 & 0.261 & 0.220 & 0.281 & 0.190 & 0.259 & 0.276 & 0.336 & 0.245 & 0.285 \\
 & 336 & \textbf{0.243} & \textbf{0.277} & \underline{0.245} & \underline{0.282} & 0.265 & 0.319 & 0.280 & 0.306 & 0.278 & 0.331 & 0.259 & 0.326 & 0.339 & 0.380 & 0.321 & 0.338 \\
 & 720 & \textbf{0.313} & \textbf{0.328} & \underline{0.314} & \underline{0.334} & 0.323 & 0.362 & 0.365 & 0.359 & 0.311 & 0.356 & 0.332 & 0.382 & 0.403 & 0.428 & 0.414 & 0.410 \\
\cline{2-18}
 & Avg & \underline{0.227} & \textbf{0.261} & \textbf{0.226} & \underline{0.264} & 0.246 & 0.300 & 0.259 & 0.287 & 0.243 & 0.299 & 0.232 & 0.295 & 0.309 & 0.360 & 0.288 & 0.314 \\
\hline
\multirow{5}{*}{\rotatebox{90}{Electricity}} & 96 & \textbf{0.128} & \textbf{0.221} & \underline{0.129} & \underline{0.222} & 0.140 & 0.237 & 0.168 & 0.272 & 0.164 & 0.269 & 0.186 & 0.281 & 0.193 & 0.308 & 0.169 & 0.273 \\
 & 192 & \textbf{0.144} & \textbf{0.235} & \underline{0.147} & \underline{0.240} & 0.153 & 0.249 & 0.184 & 0.289 & 0.177 & 0.285 & 0.208 & 0.300 & 0.201 & 0.315 & 0.182 & 0.286 \\
 & 336 & \textbf{0.160} & \textbf{0.252} & \underline{0.163} & \underline{0.259} & 0.169 & 0.267 & 0.198 & 0.300 & 0.193 & 0.304 & 0.323 & 0.369 & 0.214 & 0.329 & 0.200 & 0.304  \\
 & 720 & \underline{0.199} & \textbf{0.285} & \textbf{0.197} & \underline{0.290} & 0.203 & 0.301 & 0.220 & 0.320 & 0.212 & 0.321 & 0.404 & 0.423 & 0.246 & 0.355 & 0.222 & 0.321 \\
\cline{2-18}
 & Avg & \textbf{0.157} & \textbf{0.248} & \underline{0.159} & \underline{0.253} & 0.166 & 0.264 & 0.192 & 0.295 & 0.187 & 0.295 & 0.280 & 0.343 & 0.214 & 0.327 & 0.193 & 0.296 \\
\hline
\multirow{5}{*}{\rotatebox{90}{Traffic}} & 96 & \underline{0.365} & \textbf{0.243} & \textbf{0.360} & \underline{0.249} & 0.410 & 0.282 & 0.593 & 0.321 & 0.519 & 0.309 & 0.511 & 0.292 & 0.587 & 0.366 & 0.612 & 0.338 \\
 & 192 & \underline{0.382} & \textbf{0.250} & \textbf{0.379} & \underline{0.256} & 0.423 & 0.287 & 0.617 & 0.336 & 0.537 & 0.315 & 0.523 & 0.311 & 0.604 & 0.373 & 0.613 & 0.340 \\
 & 336 & \underline{0.395} & \textbf{0.257} & \textbf{0.392} & \underline{0.264} & 0.436 & 0.296 & 0.629 & 0.336 & 0.534 & 0.313 & 0.530 & 0.300 & 0.621 & 0.383 & 0.618 & 0.328 \\
 & 720 & \textbf{0.431} & \textbf{0.277} & \underline{0.432} & \underline{0.286} & 0.466 & 0.315 & 0.640 & 0.350 & 0.577 & 0.325 & 0.573 & 0.313 & 0.626 & 0.382 & 0.653 & 0.355  \\
\cline{2-18}
 & Avg & \underline{0.393} & \textbf{0.256} & \textbf{0.391} & \underline{0.264} & 0.434 & 0.295 & 0.620 & 0.336 & 0.542 & 0.316 & 0.534 & 0.304 & 0.610 & 0.376 & 0.624 & 0.340 \\
\hline
\multicolumn{2}{c|}{${1^{st}}$ Count} &  \multicolumn{2}{c|}{\textbf{56}}  &  \multicolumn{2}{c|}{\underline{16}}  & \multicolumn{2}{c|}{0}  &  \multicolumn{2}{c|}{0}  &  \multicolumn{2}{c|}{0} & \multicolumn{2}{c|}{0} & \multicolumn{2}{c|}{0} &
\multicolumn{2}{c}{0} \\
\bottomrule
\end{tabular}}
\caption{Multivariate time series forecasting results compared with end-to-end methods. We compare extensive competitive models under different prediction lengths. The best results are highlighted in bold and the second best results are highlighted with underline. The standard deviations of PT-Tuning are within 0.003 for MSE and within 0.002 for MAE.}
\label{table1}
\end{table*}

\section{Visualization}

\subsection{Forecasting}
To provide a clear comparison among different models, we plot the prediction results on four selected time series datasets with end-to-end and masked reconstruction methods. As shown in Figure 8-Figure 11, we set the look-back window as 512 and the prediction horizon as 336. It can be observed that PT-Tuning predicts more accurate future trends compared with other strong baselines.

\subsection{Attention Map}

We also report the attention maps of four heads in the last cross-attention layers on four datasets. The look-back window is set as 512 and the prediction horizon is set as 336. And the model utilize patch size 8 to convert history and future time steps into 64 and 42 tokens respectively. As shown in Figure 12, we find that our model can effectively capture the relationship between the input sequences and forecasting sequences, leading to good forecasting performance.

\begin{figure*}[!t]
	\begin{center}
		\begin{minipage}{0.245\linewidth}
			\centerline{\ \ \ \ (a) DLinear}
			\includegraphics[width=\linewidth,height=3.05cm]{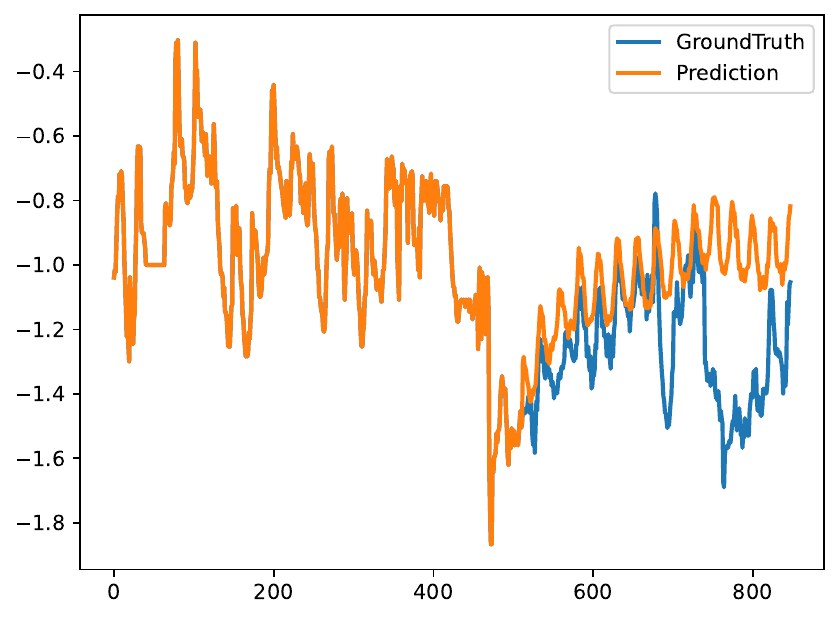}
		\end{minipage}
		\begin{minipage}{0.245\linewidth}
			\centerline{\ \ \ \ (b) PatchTST}
			\includegraphics[width=\linewidth,height=3.05cm]{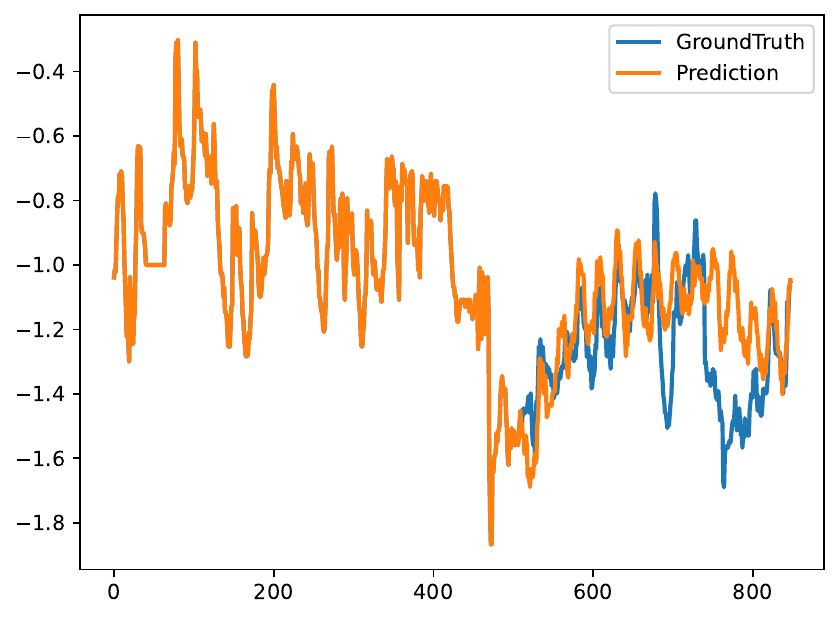}
		\end{minipage}
		\begin{minipage}{0.245\linewidth}
			\centerline{\ \ \ \ (c) Ti-MAE}
			\includegraphics[width=\linewidth,height=3.05cm]{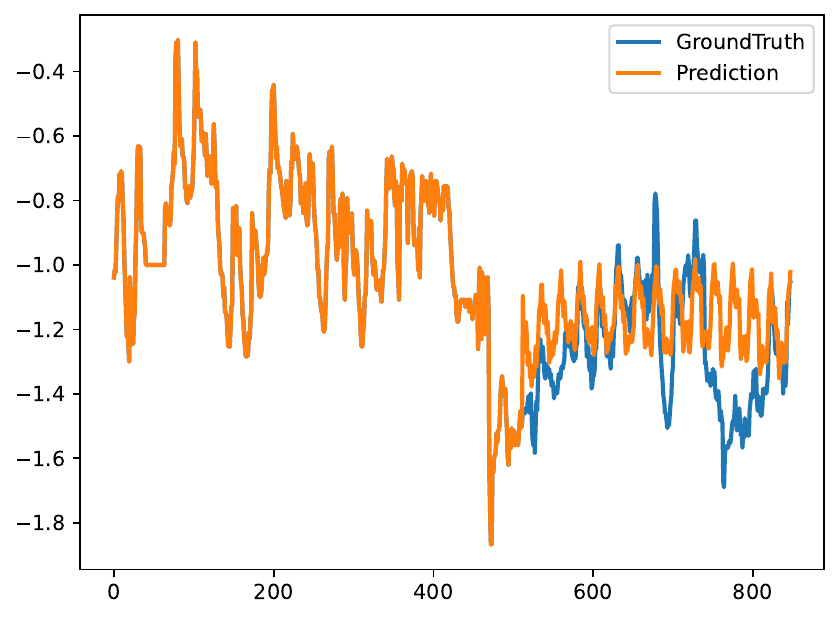}
		\end{minipage}
		\begin{minipage}{0.245\linewidth}
			\centerline{\ \ \ \ \ \ (d) PT-Tuning}
			\includegraphics[width=\linewidth,height=3.05cm]{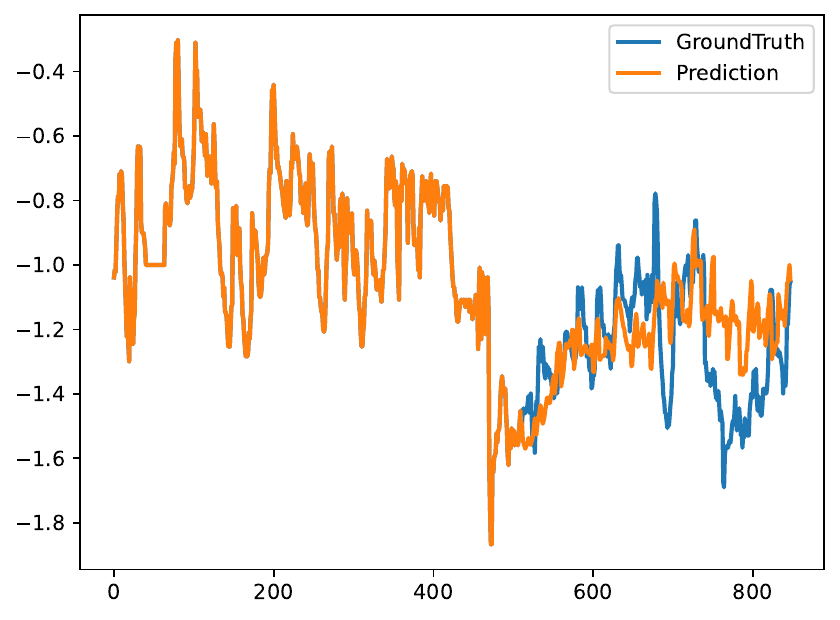}
		\end{minipage}
		\qquad			

		\begin{minipage}{0.245\linewidth}
			\includegraphics[width=\linewidth,height=3.05cm]{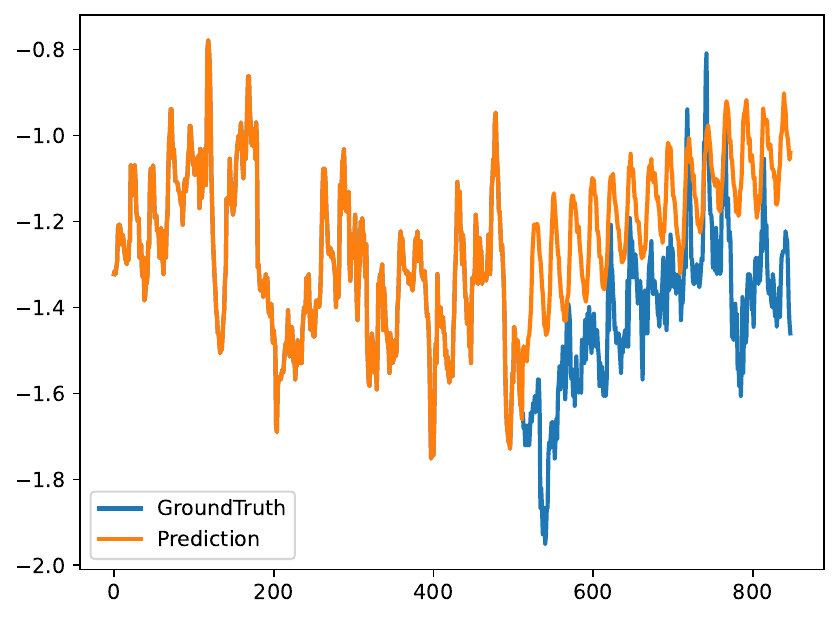}
		\end{minipage}
		\begin{minipage}{0.245\linewidth}
			\includegraphics[width=\linewidth,height=3.05cm]{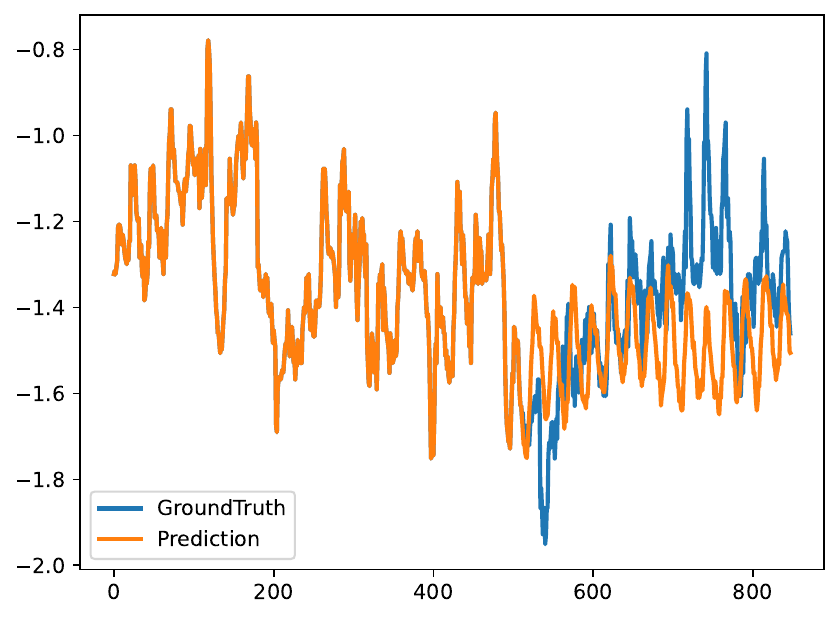}
		\end{minipage}
		\begin{minipage}{0.245\linewidth}
			\includegraphics[width=\linewidth,height=3.05cm]{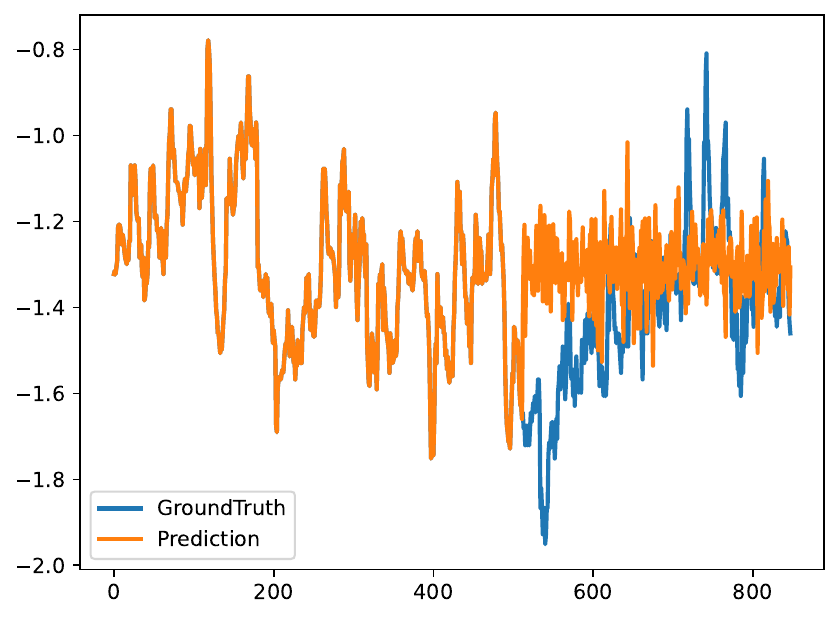}
		\end{minipage}
		\begin{minipage}{0.245\linewidth}
			\includegraphics[width=\linewidth,height=3.05cm]{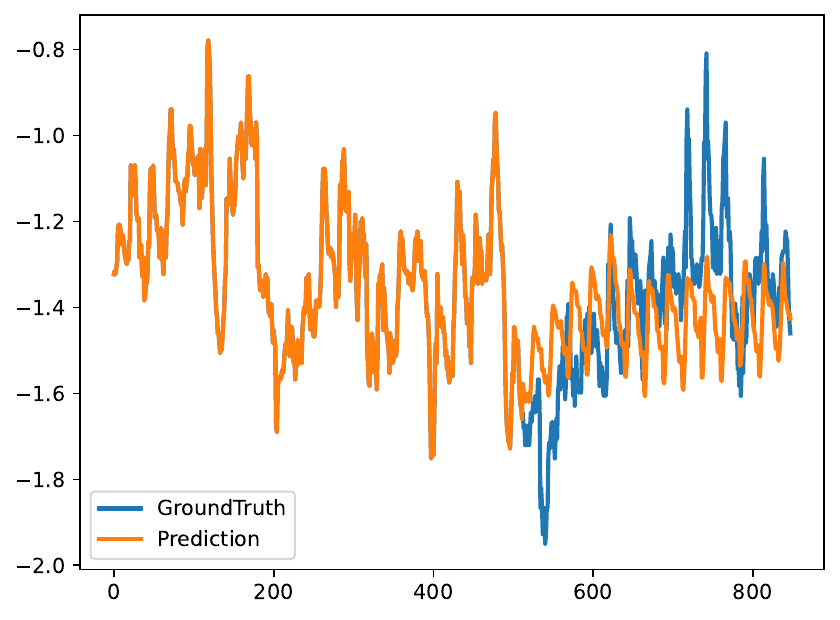}
		\end{minipage}
	\end{center}
	\caption{Illustration of the forecasting outputs of four models with an input length 512 and output length 336 on ETTh1.}
	\label{fig3}
\end{figure*}

\begin{figure*}[!t]
	\begin{center}
		\begin{minipage}{0.245\linewidth}
			\centerline{\ \ \ \ (a) DLinear}
			\includegraphics[width=\linewidth,height=3.05cm]{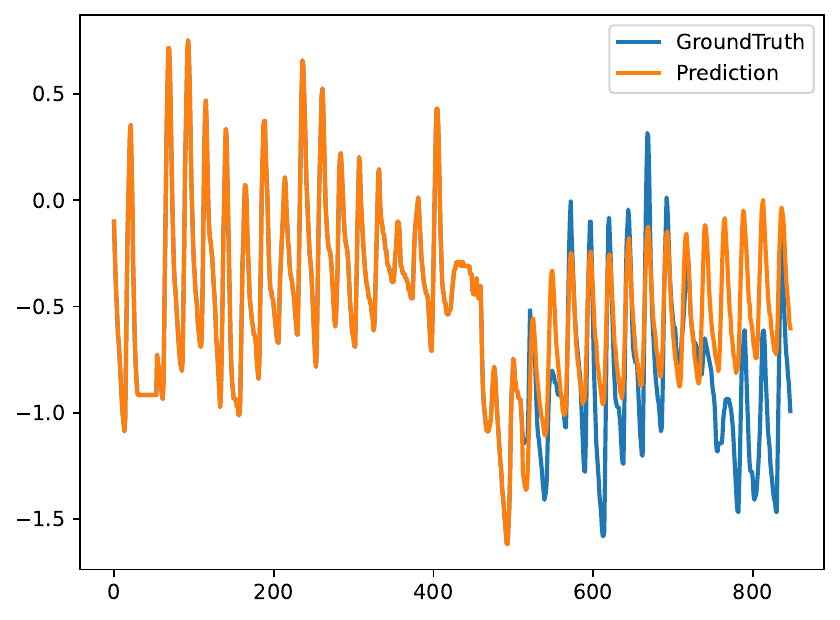}
		\end{minipage}
		\begin{minipage}{0.245\linewidth}
			\centerline{\ \ \ \ (b) PatchTST}
			\includegraphics[width=\linewidth,height=3.05cm]{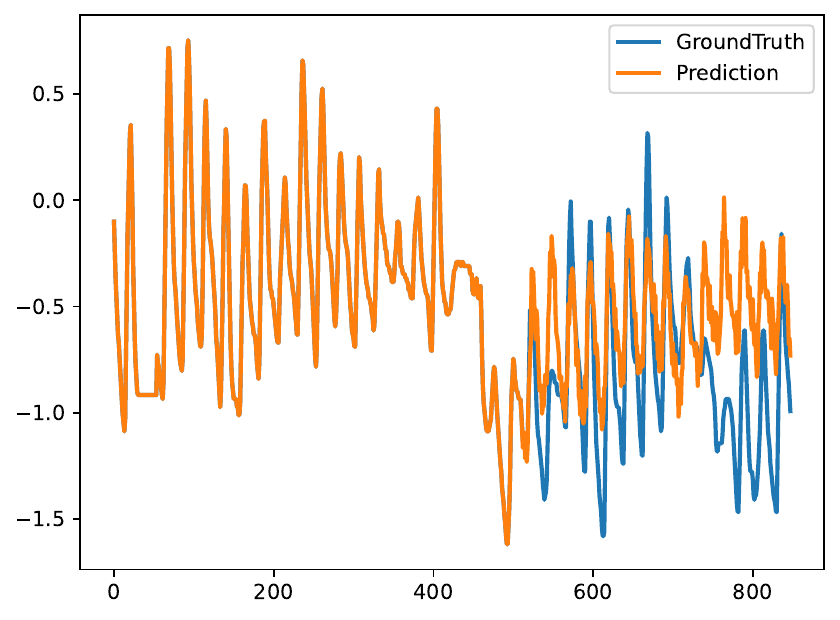}
		\end{minipage}
		\begin{minipage}{0.245\linewidth}
			\centerline{\ \ \ \ (c) Ti-MAE}
			\includegraphics[width=\linewidth,height=3.05cm]{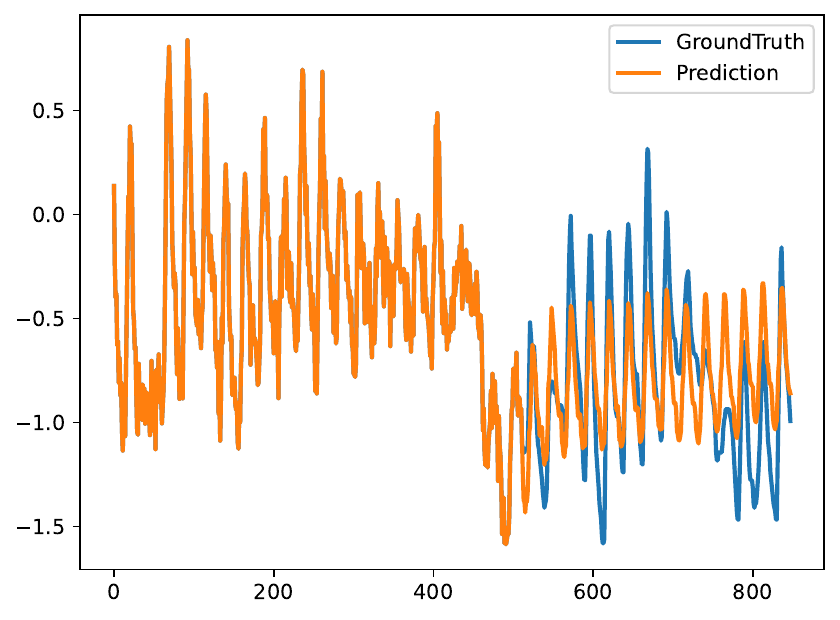}
		\end{minipage}
		\begin{minipage}{0.245\linewidth}
			\centerline{\ \ \ \ \ \ (d) PT-Tuning}
			\includegraphics[width=\linewidth,height=3.05cm]{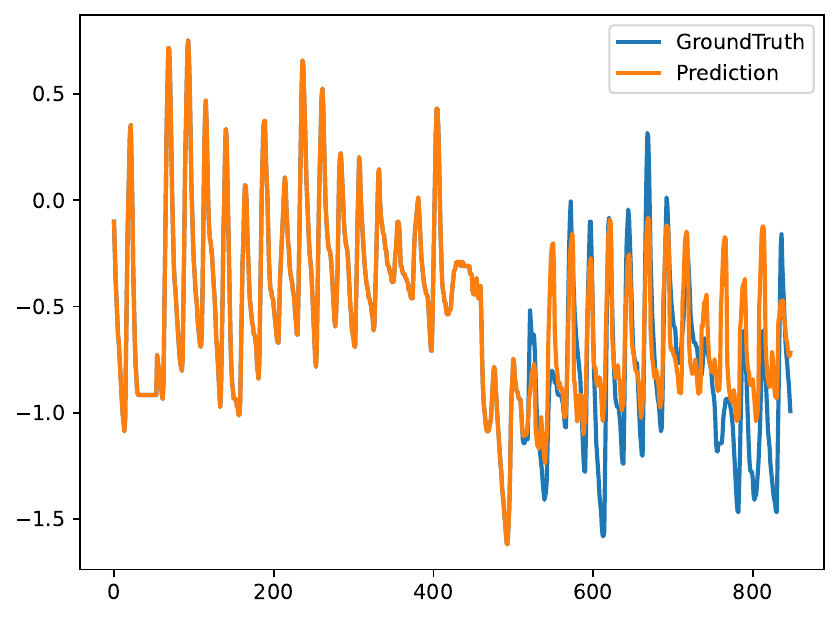}
		\end{minipage}
		\qquad			

		\begin{minipage}{0.245\linewidth}
			\includegraphics[width=\linewidth,height=3.05cm]{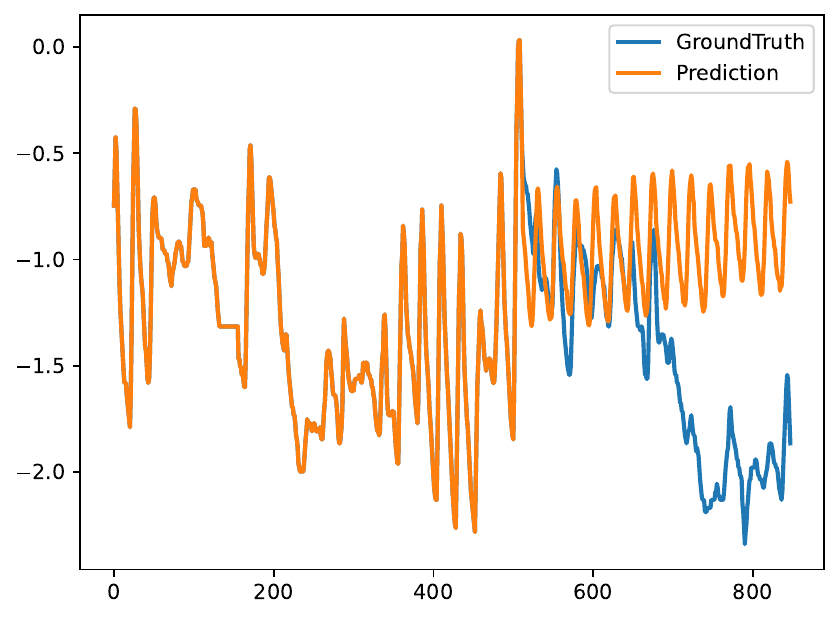}
		\end{minipage}
		\begin{minipage}{0.245\linewidth}
			\includegraphics[width=\linewidth,height=3.05cm]{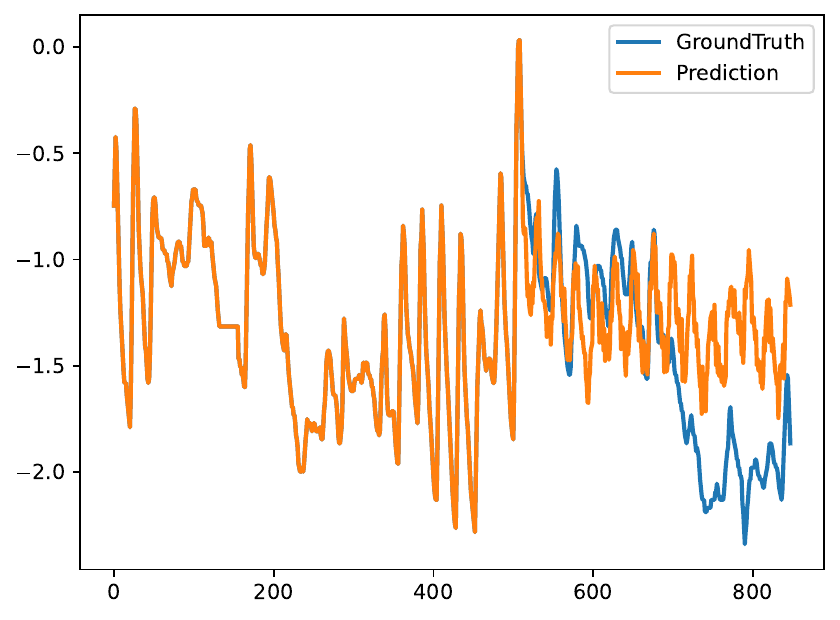}
		\end{minipage}
		\begin{minipage}{0.245\linewidth}
			\includegraphics[width=\linewidth,height=3.05cm]{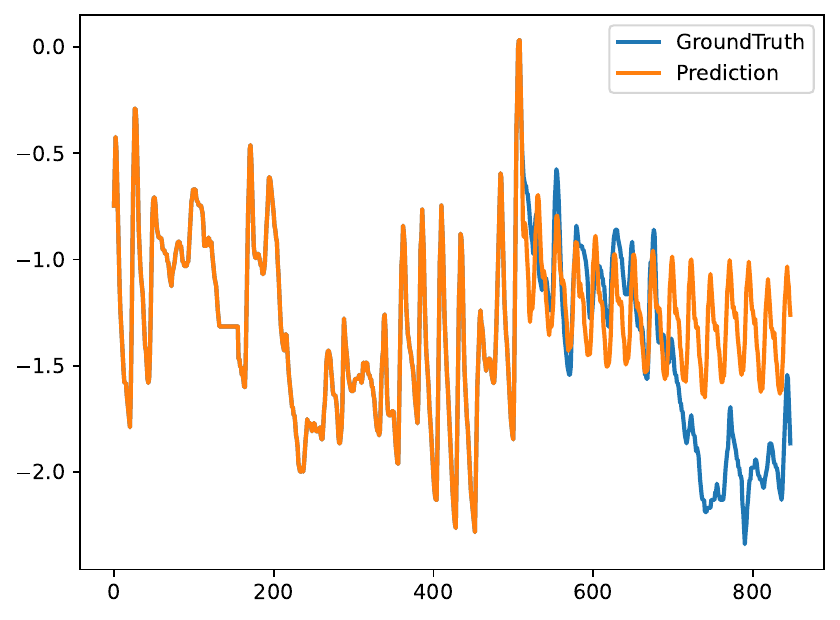}
		\end{minipage}
		\begin{minipage}{0.245\linewidth}
			\includegraphics[width=\linewidth,height=3.05cm]{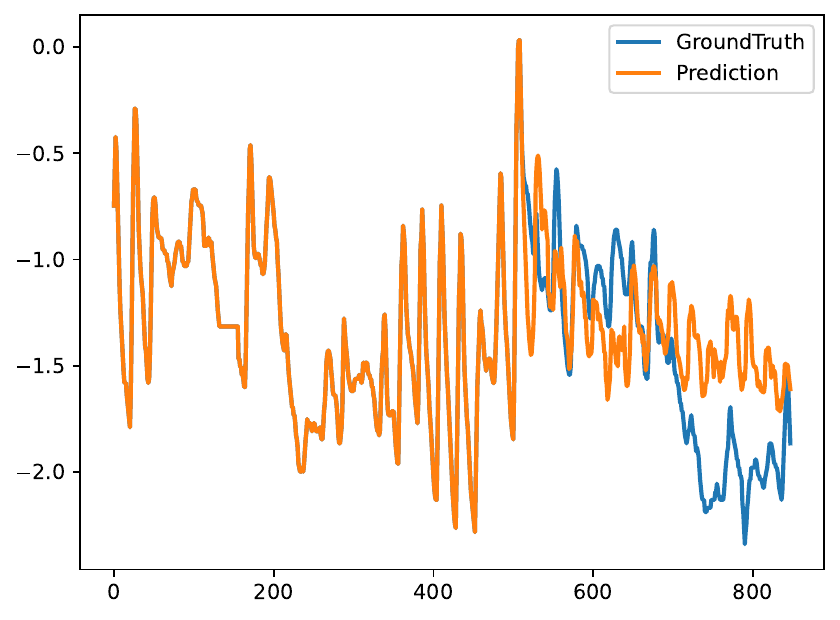}
		\end{minipage}
	\end{center}
	\caption{Illustration of the forecasting outputs of four models with an input length 512 and output length 336 on ETTh2.}
	\label{fig3}
\end{figure*}

\begin{figure*}[!t]
	\begin{center}
		\begin{minipage}{0.245\linewidth}
			\centerline{\ \ \ \ (a) DLinear}
			\includegraphics[width=\linewidth,height=3.05cm]{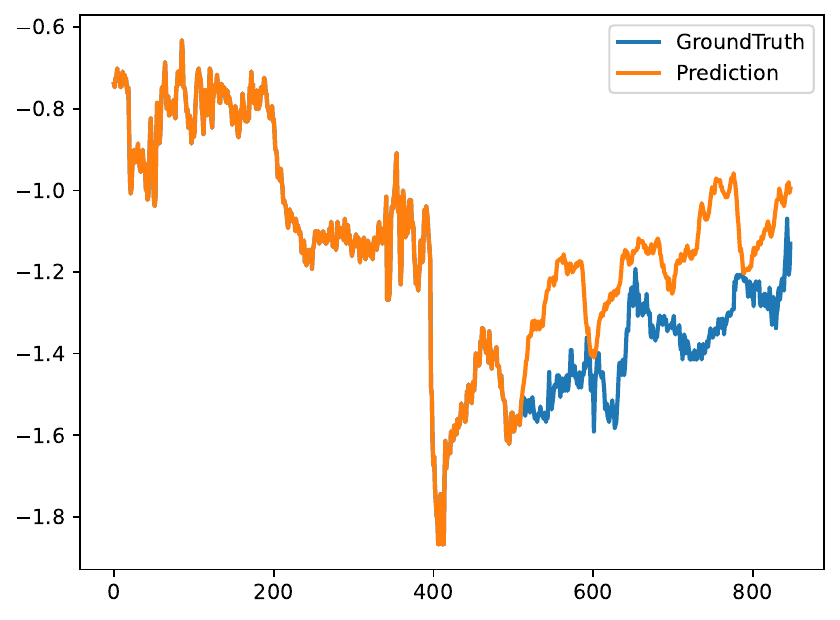}
		\end{minipage}
		\begin{minipage}{0.245\linewidth}
			\centerline{\ \ \ \ (b) PatchTST}
			\includegraphics[width=\linewidth,height=3.05cm]{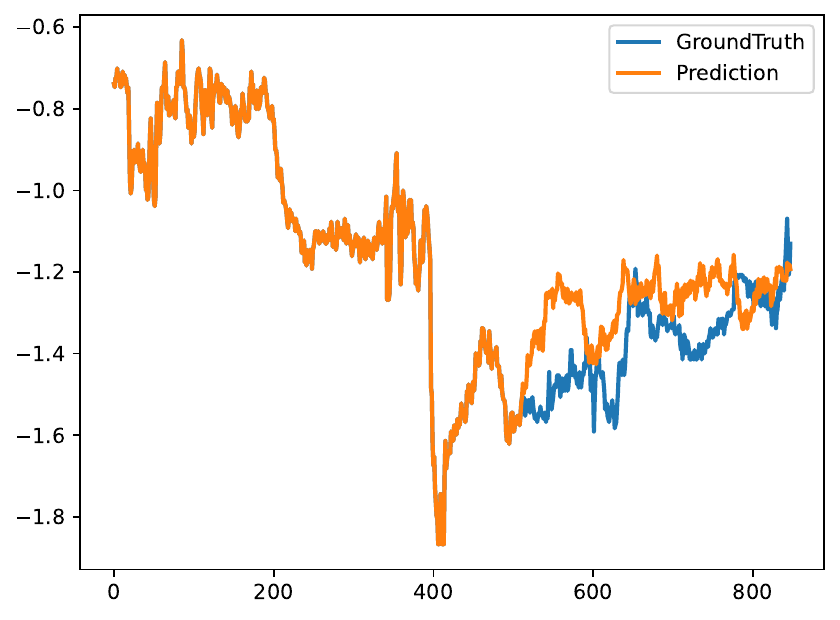}
		\end{minipage}
		\begin{minipage}{0.245\linewidth}
			\centerline{\ \ \ \ (c) Ti-MAE}
			\includegraphics[width=\linewidth,height=3.05cm]{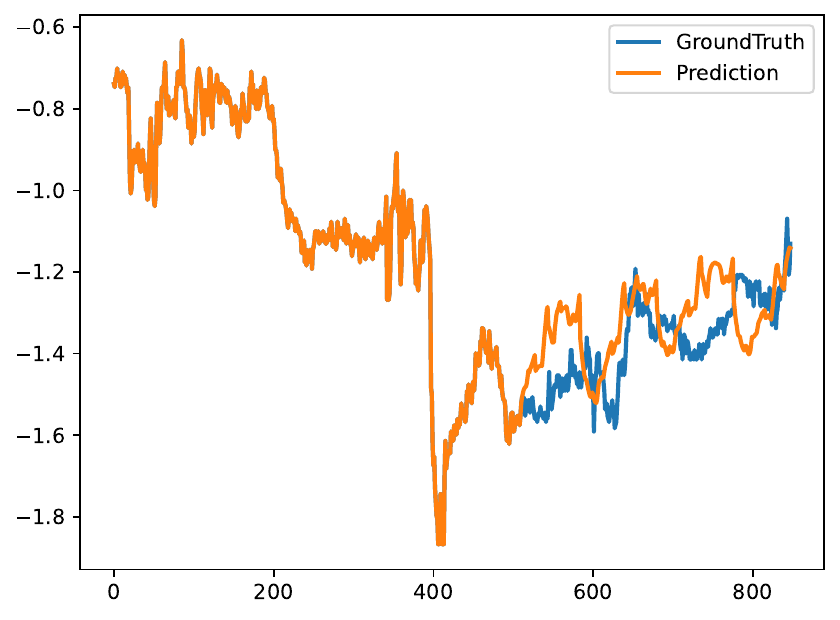}
		\end{minipage}
		\begin{minipage}{0.245\linewidth}
			\centerline{\ \ \ \ \ \ (d) PT-Tuning}
			\includegraphics[width=\linewidth,height=3.05cm]{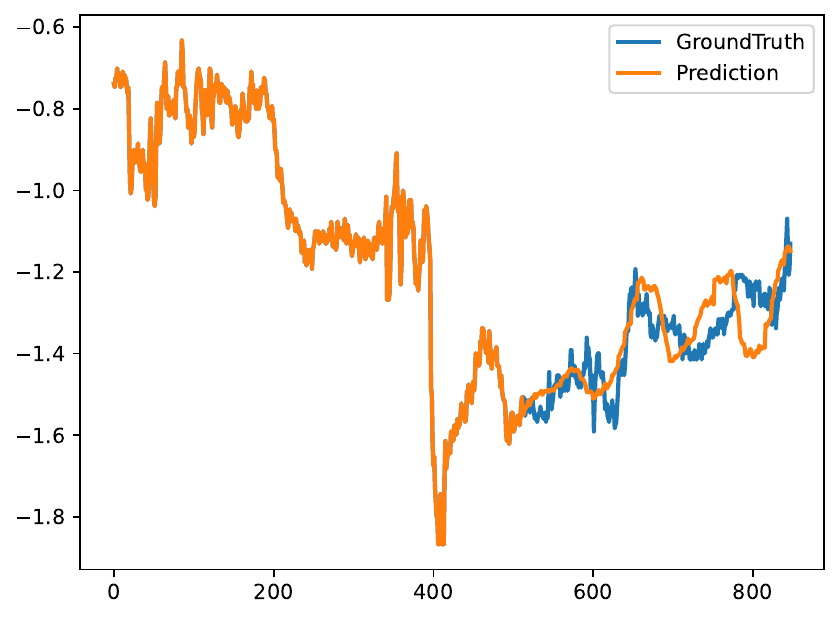}
		\end{minipage}
		\qquad			

		\begin{minipage}{0.245\linewidth}
			\includegraphics[width=\linewidth,height=3.05cm]{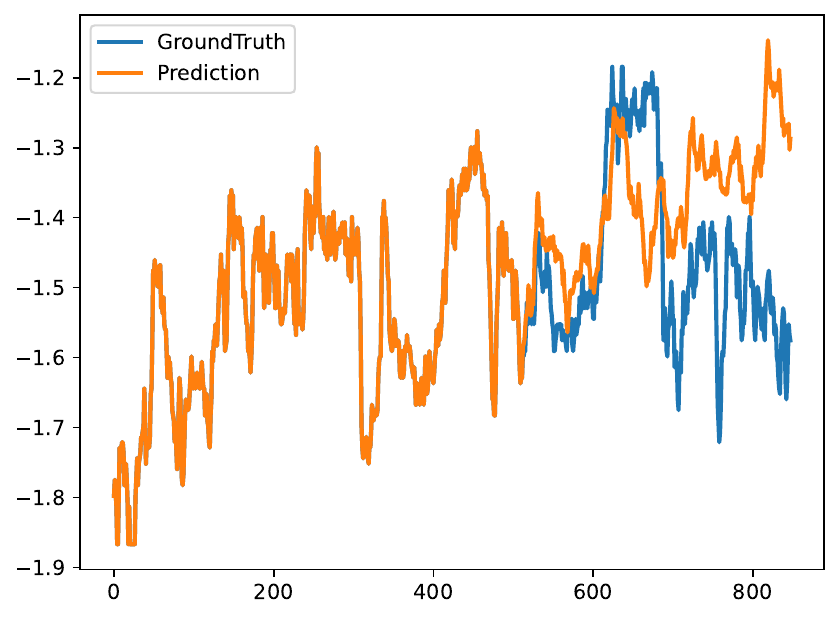}
		\end{minipage}
		\begin{minipage}{0.245\linewidth}
			\includegraphics[width=\linewidth,height=3.05cm]{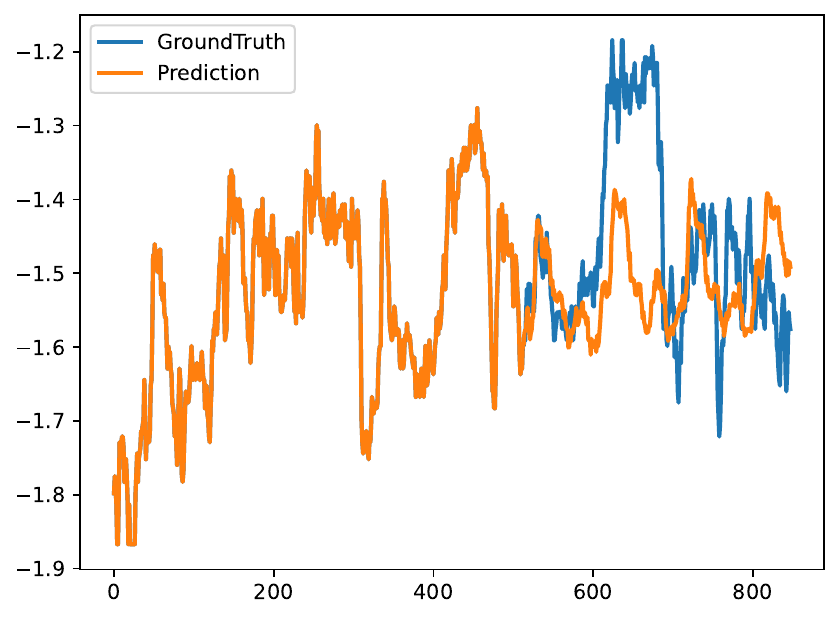}
		\end{minipage}
		\begin{minipage}{0.245\linewidth}
			\includegraphics[width=\linewidth,height=3.05cm]{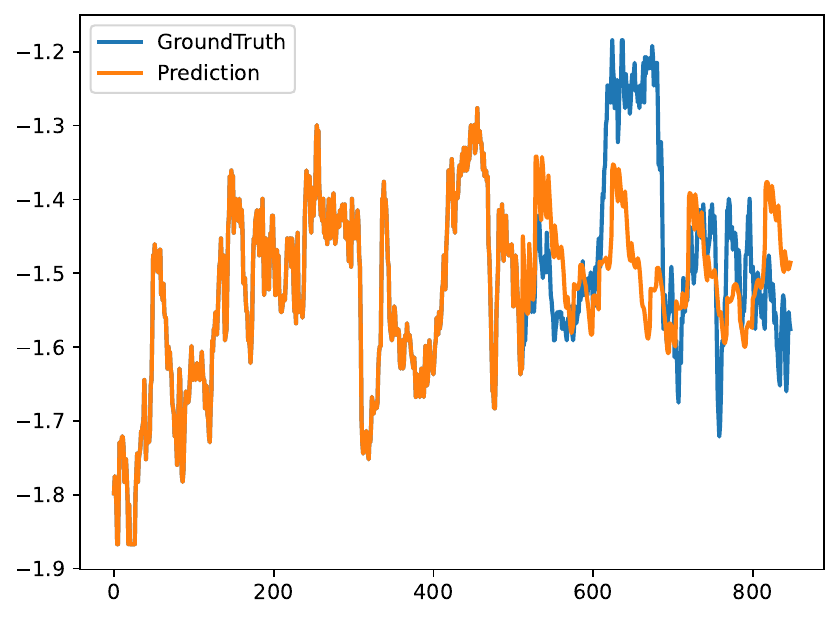}
		\end{minipage}
		\begin{minipage}{0.245\linewidth}
			\includegraphics[width=\linewidth,height=3.05cm]{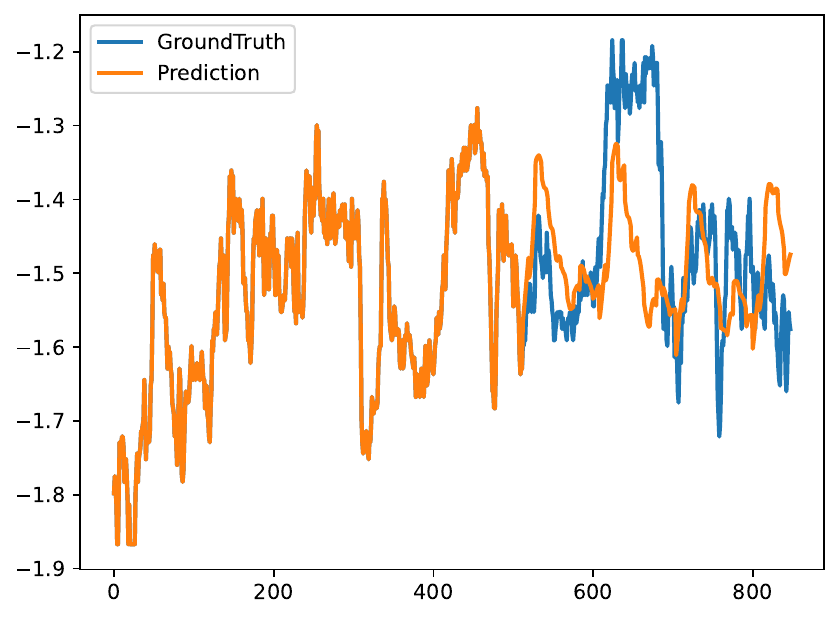}
		\end{minipage}
	\end{center}
	\caption{Illustration of the forecasting outputs of four models with an input length 512 and output length 336 on ETTm1.}
	\label{fig3}
\end{figure*}

\begin{figure*}[!t]
	\begin{center}
		\begin{minipage}{0.245\linewidth}
			\centerline{\ \ \ \ (a) DLinear}
			\includegraphics[width=\linewidth,height=3.05cm]{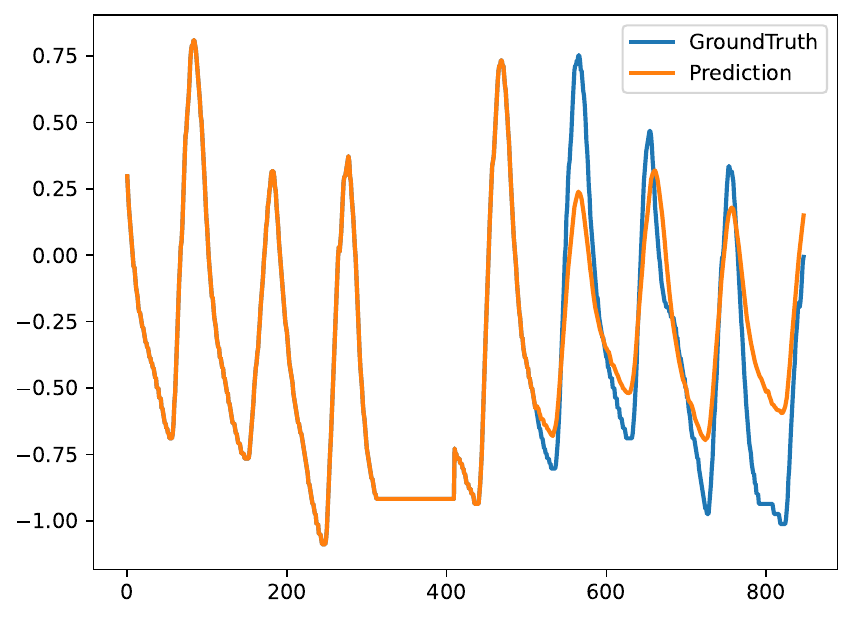}
		\end{minipage}
		\begin{minipage}{0.245\linewidth}
			\centerline{\ \ \ \ (b) PatchTST}
			\includegraphics[width=\linewidth,height=3.05cm]{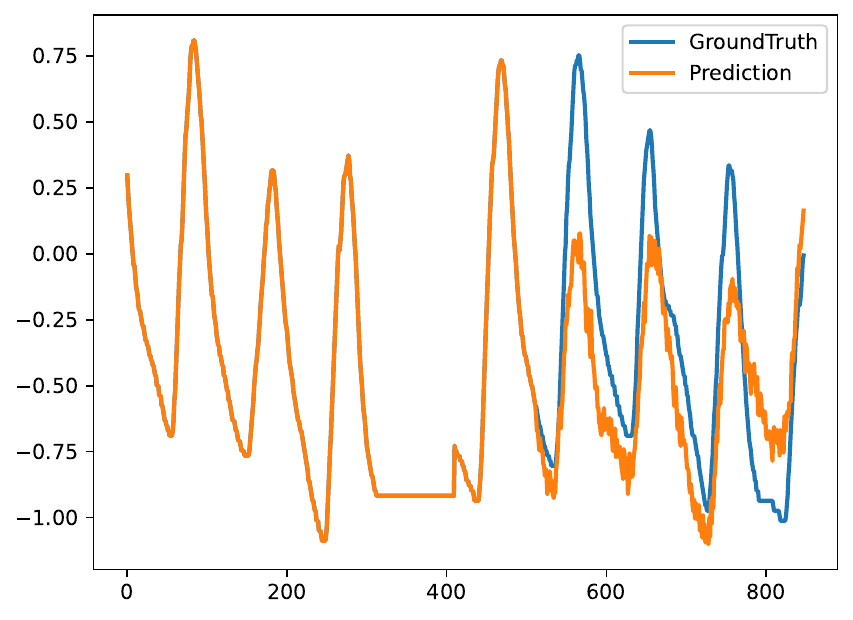}
		\end{minipage}
		\begin{minipage}{0.245\linewidth}
			\centerline{\ \ \ \ (c) Ti-MAE}
			\includegraphics[width=\linewidth,height=3.05cm]{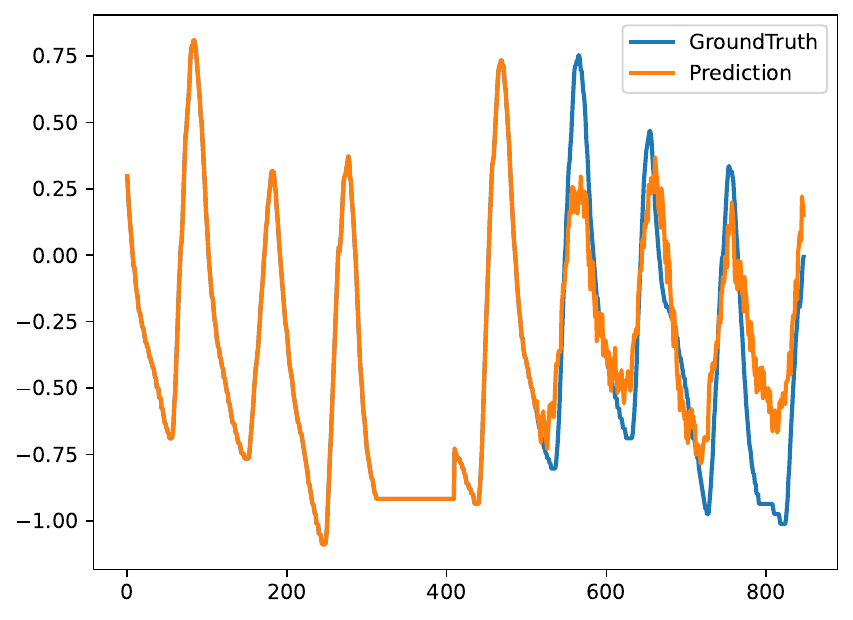}
		\end{minipage}
		\begin{minipage}{0.245\linewidth}
			\centerline{\ \ \ \ \ \ (d) PT-Tuning}
			\includegraphics[width=\linewidth,height=3.05cm]{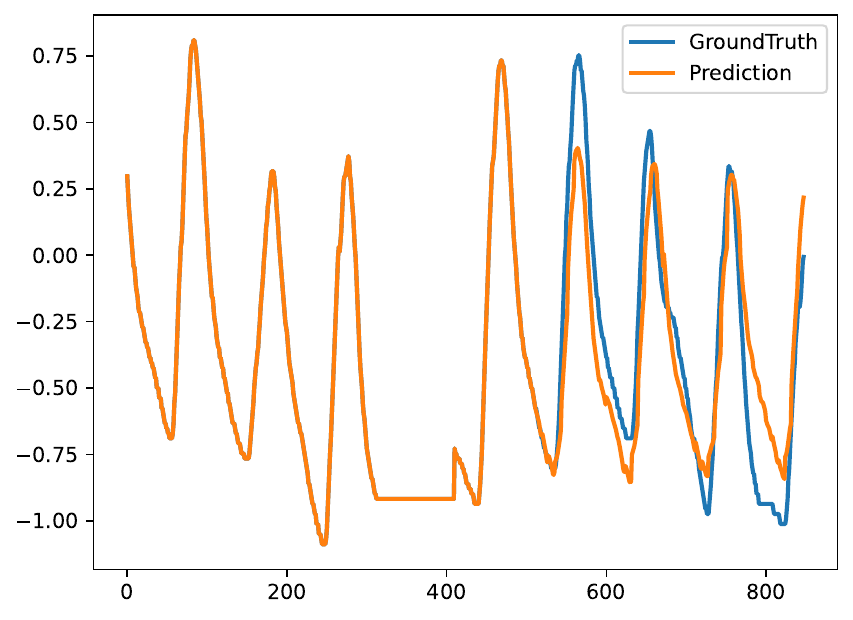}
		\end{minipage}
		\qquad			

		\begin{minipage}{0.245\linewidth}
			\includegraphics[width=\linewidth,height=3.05cm]{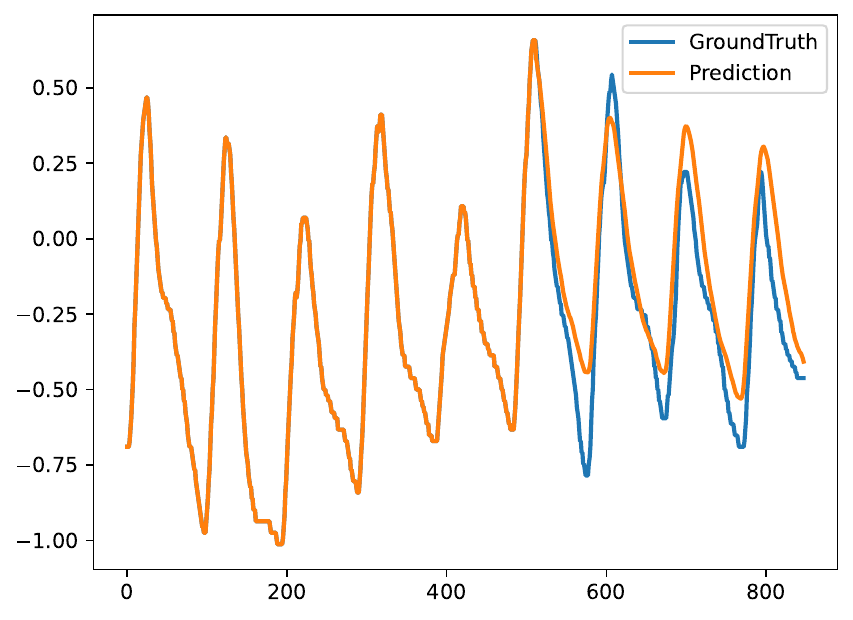}
		\end{minipage}
		\begin{minipage}{0.245\linewidth}
			\includegraphics[width=\linewidth,height=3.05cm]{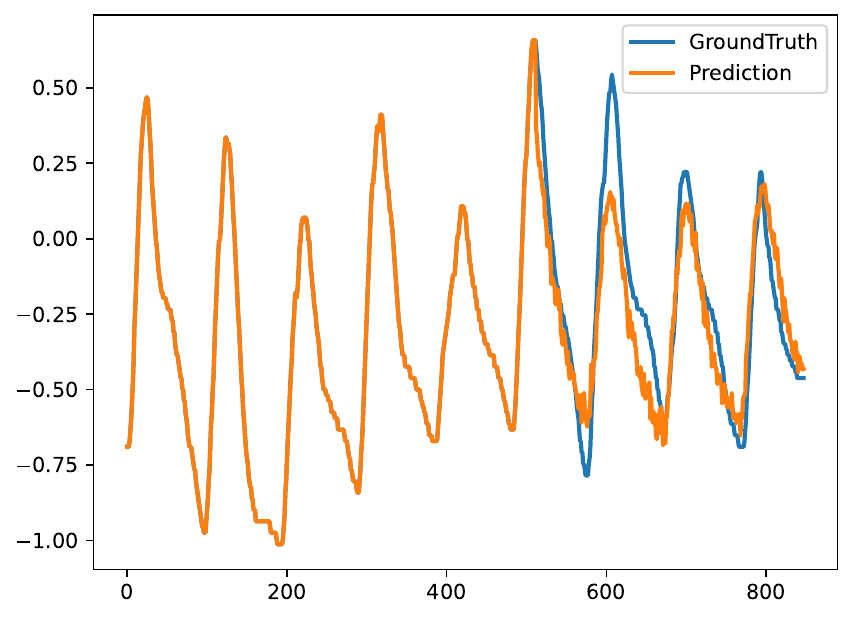}
		\end{minipage}
		\begin{minipage}{0.245\linewidth}
			\includegraphics[width=\linewidth,height=3.05cm]{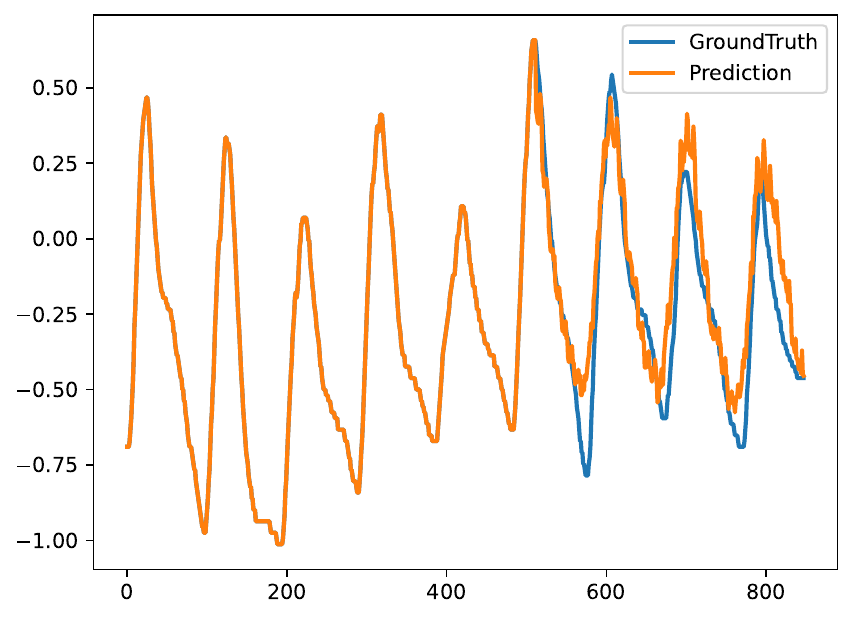}
		\end{minipage}
		\begin{minipage}{0.245\linewidth}
			\includegraphics[width=\linewidth,height=3.05cm]{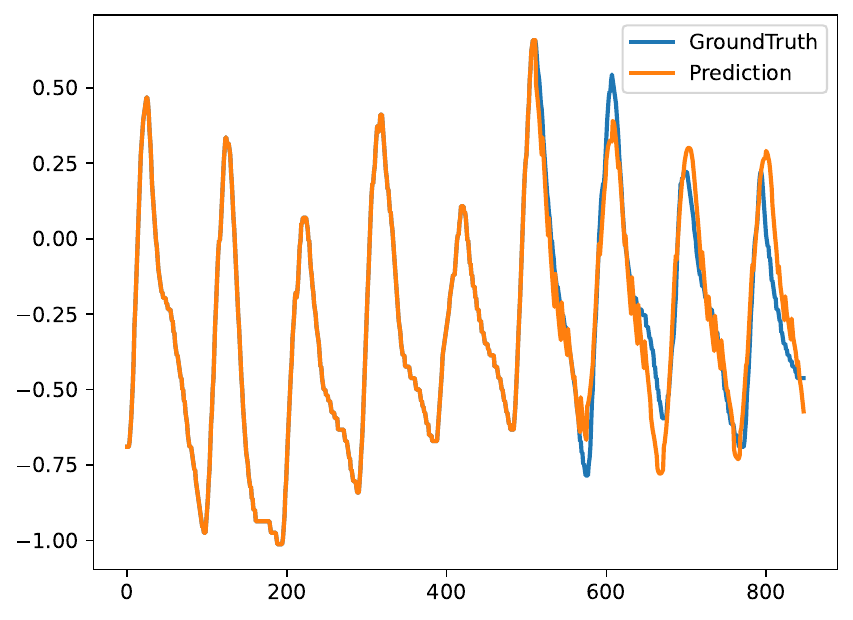}
		\end{minipage}
	\end{center}
	\caption{Illustration of the forecasting outputs of four models with an input length 512 and output length 336 on ETTm2.}
	\label{fig3}
\end{figure*}

\begin{figure*}[!t]
	\begin{center}
		\begin{minipage}{0.195\linewidth}
			\centerline{\ \ \ \ (a) Prediction}
		\end{minipage}
		\begin{minipage}{0.195\linewidth}
			\ \ \ \ \ \ \ \ \ (b) Head 1
		\end{minipage}
		\begin{minipage}{0.195\linewidth}
			\ \ \ \ \ \ \ \ \ (c) Head 2
		\end{minipage}
		\begin{minipage}{0.195\linewidth}
			\ \ \ \ \ \ \ \ \ (d) Head 3
		\end{minipage}
		\begin{minipage}{0.195\linewidth}
			\ \ \ \ \ \ \ \ \ (e) Head 4
		\end{minipage}

		\enspace

		\begin{minipage}{0.195\linewidth}
			\includegraphics[width=\linewidth]{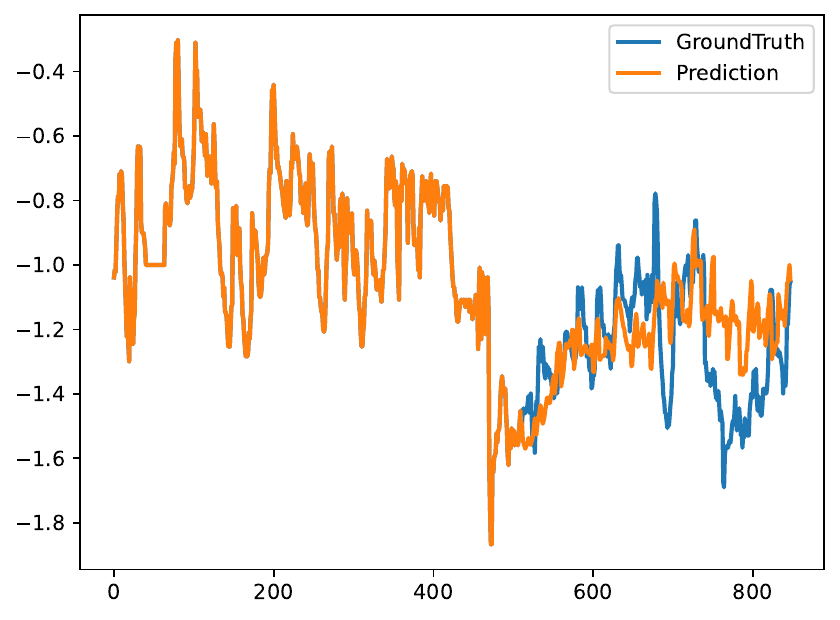}
		\end{minipage}
		\begin{minipage}{0.195\linewidth}
			\includegraphics[width=\linewidth,height=3.05cm]{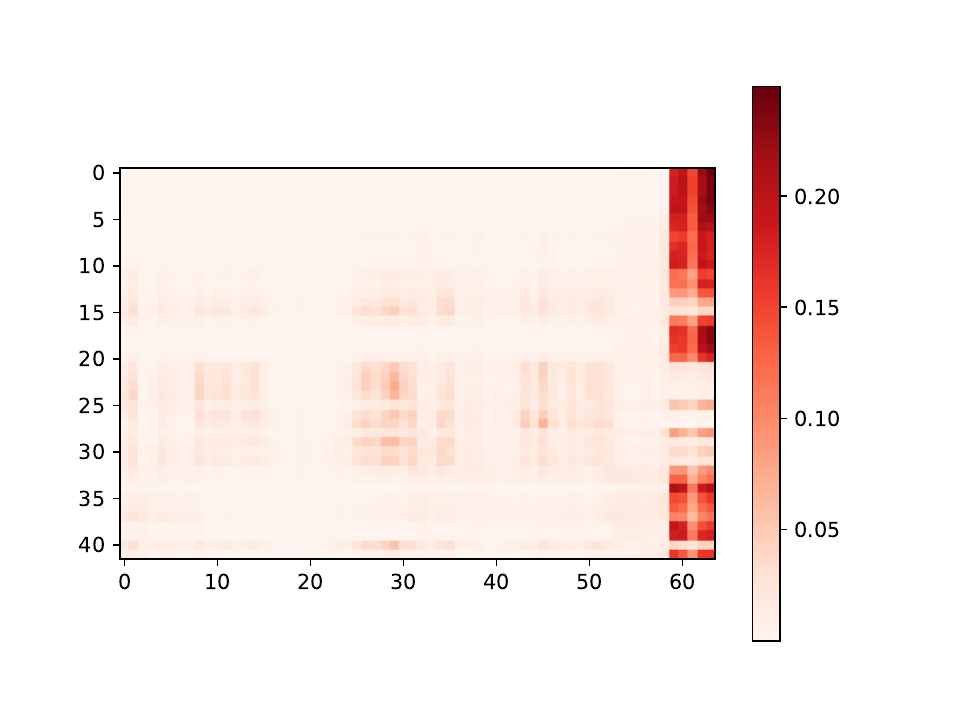}
		\end{minipage}
		\begin{minipage}{0.195\linewidth}
			\includegraphics[width=\linewidth,height=3.05cm]{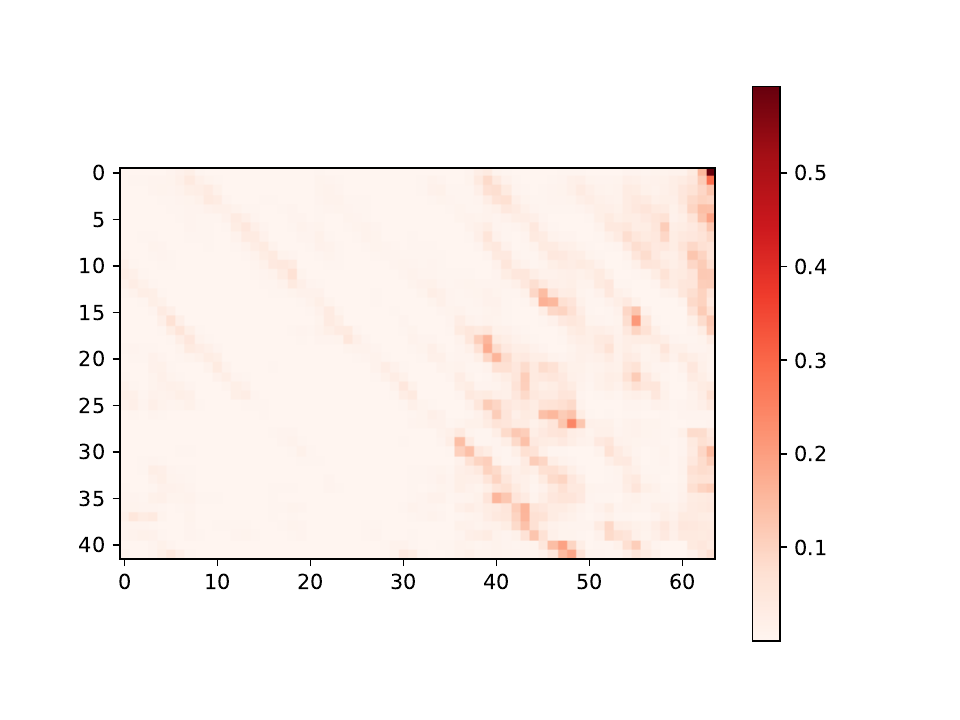}
		\end{minipage}
		\begin{minipage}{0.195\linewidth}
			\includegraphics[width=\linewidth,height=3.05cm]{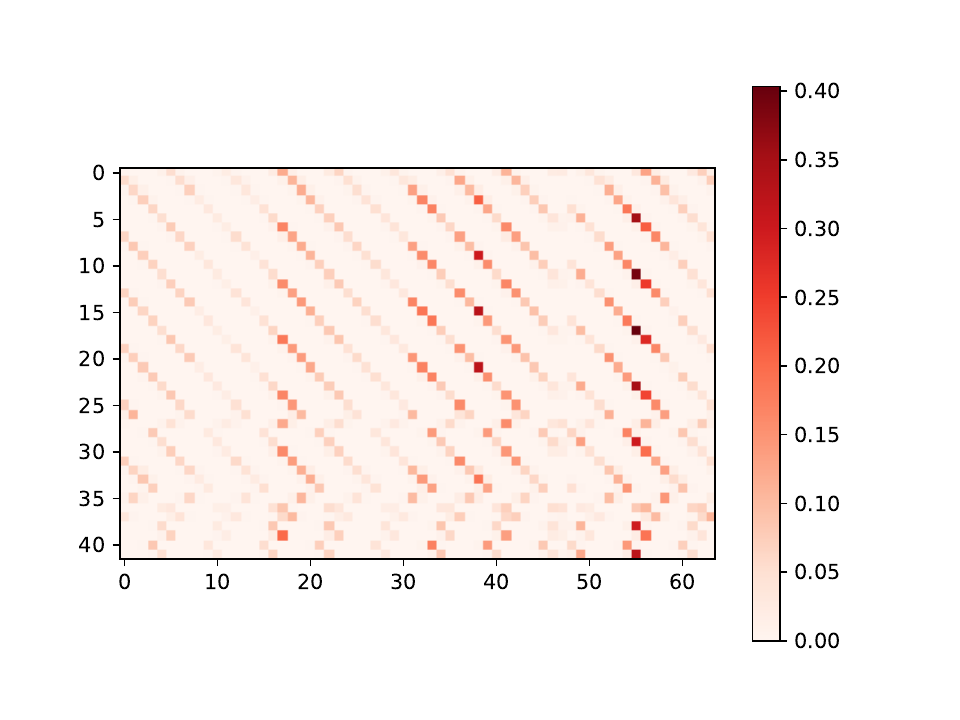}
		\end{minipage}
		\begin{minipage}{0.195\linewidth}
			\includegraphics[width=\linewidth,height=3.05cm]{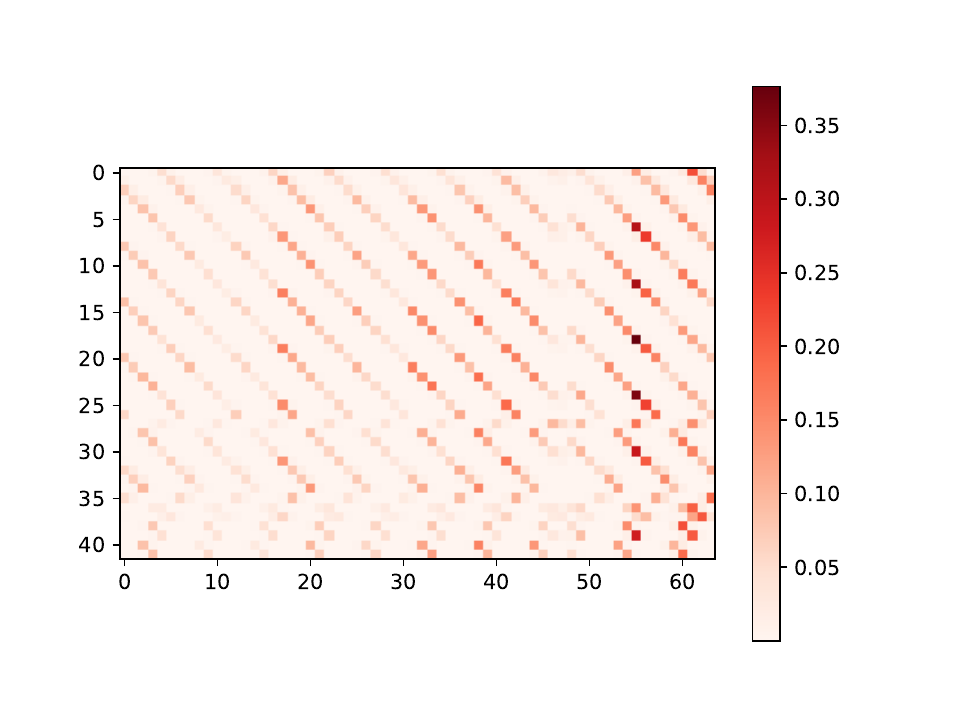}
		\end{minipage}

		\qquad			

		\begin{minipage}{0.195\linewidth}
			\includegraphics[width=\linewidth]{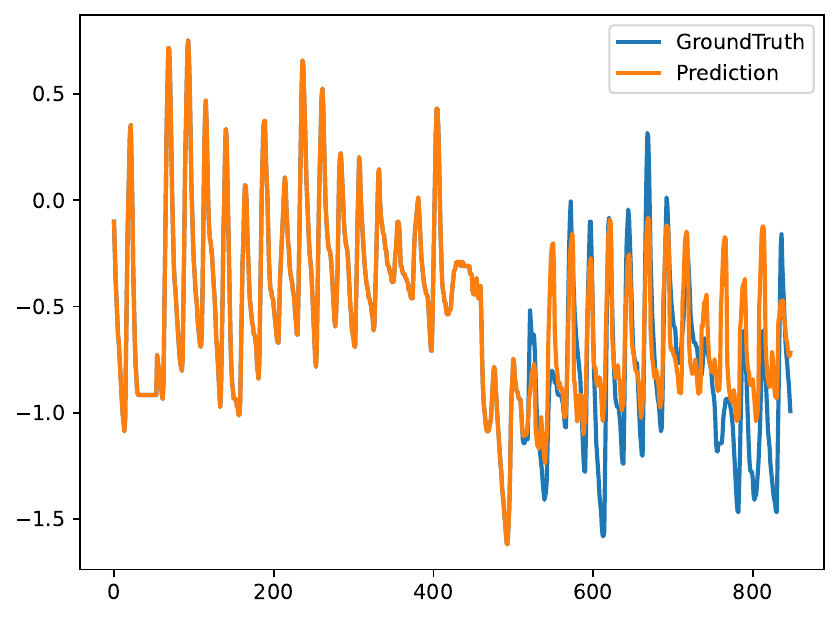}
		\end{minipage}
		\begin{minipage}{0.195\linewidth}
			\includegraphics[width=\linewidth,height=3.05cm]{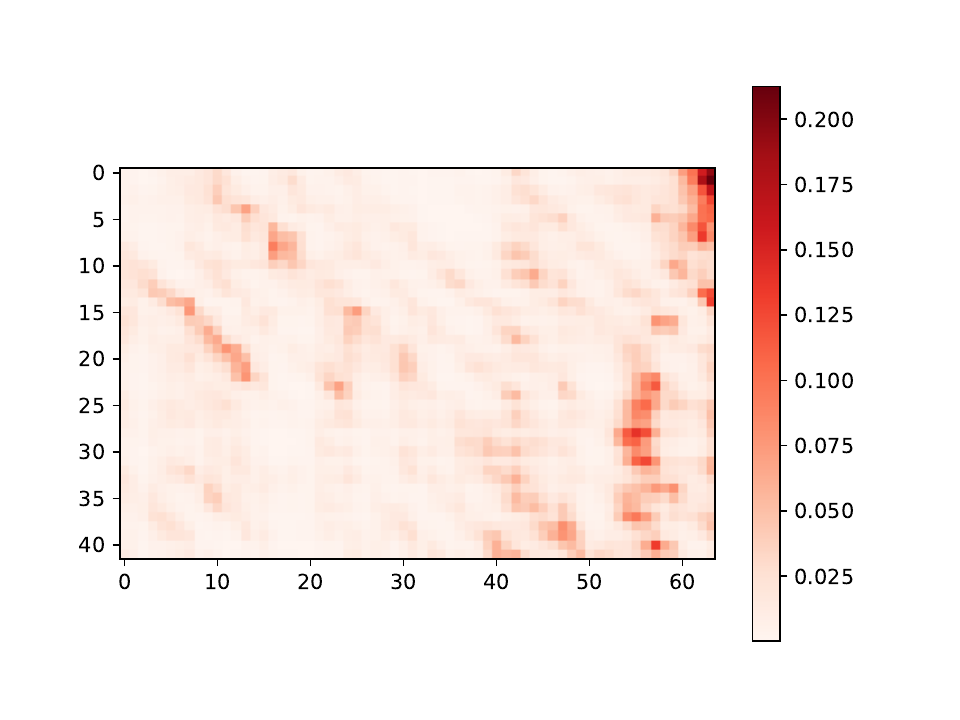}
		\end{minipage}
		\begin{minipage}{0.195\linewidth}
			\includegraphics[width=\linewidth,height=3.05cm]{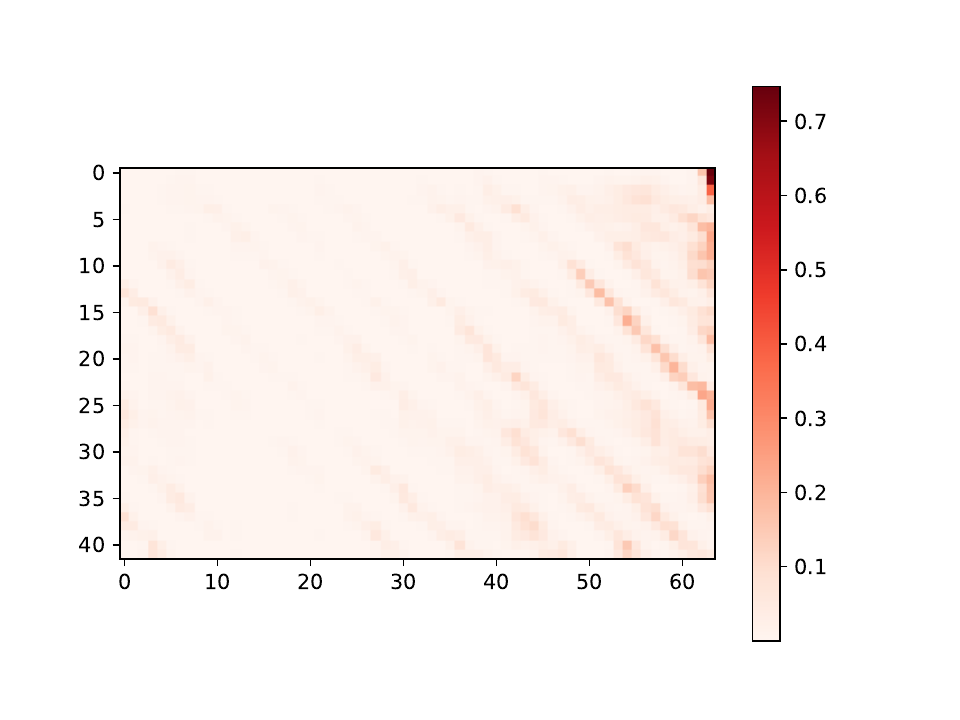}
		\end{minipage}
		\begin{minipage}{0.195\linewidth}
			\includegraphics[width=\linewidth,height=3.05cm]{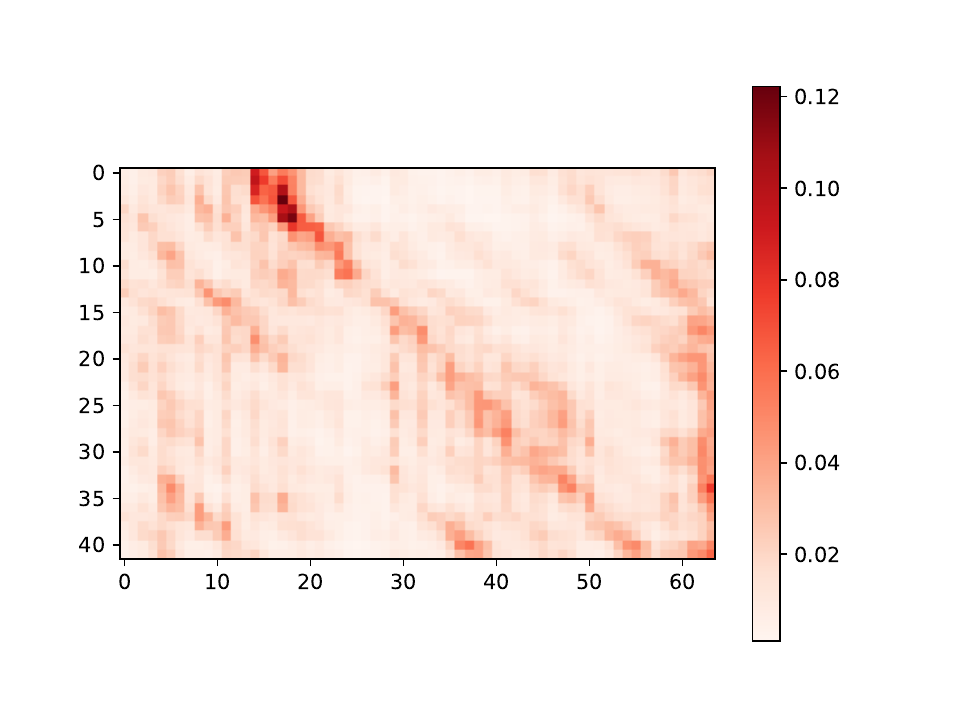}
		\end{minipage}
		\begin{minipage}{0.195\linewidth}
			\includegraphics[width=\linewidth,height=3.05cm]{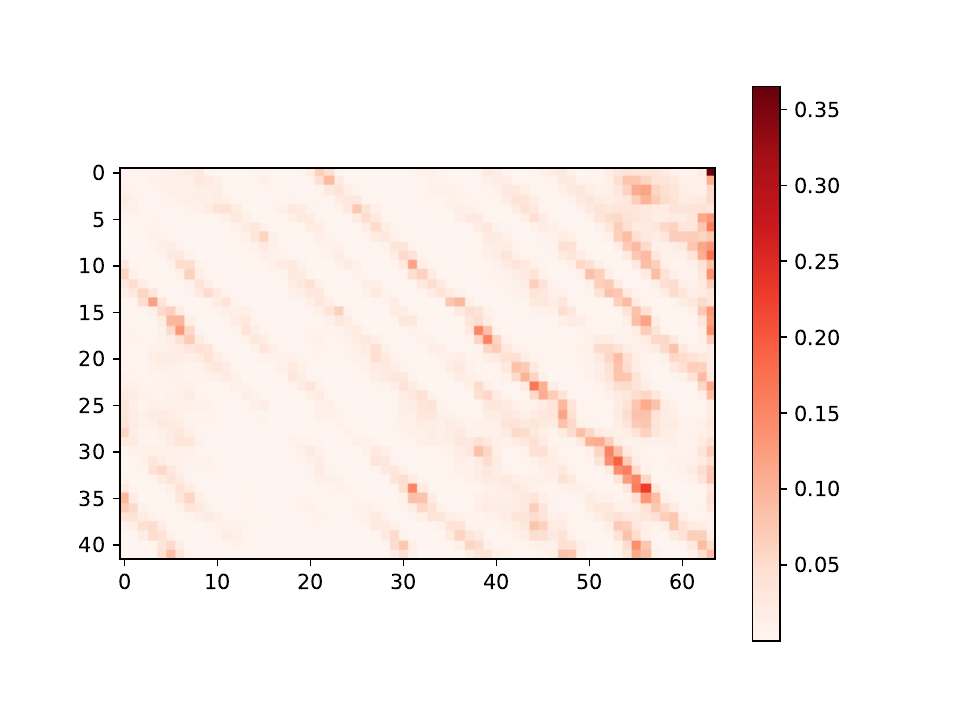}
		\end{minipage}
		\qquad			

		\begin{minipage}{0.195\linewidth}
			\includegraphics[width=\linewidth]{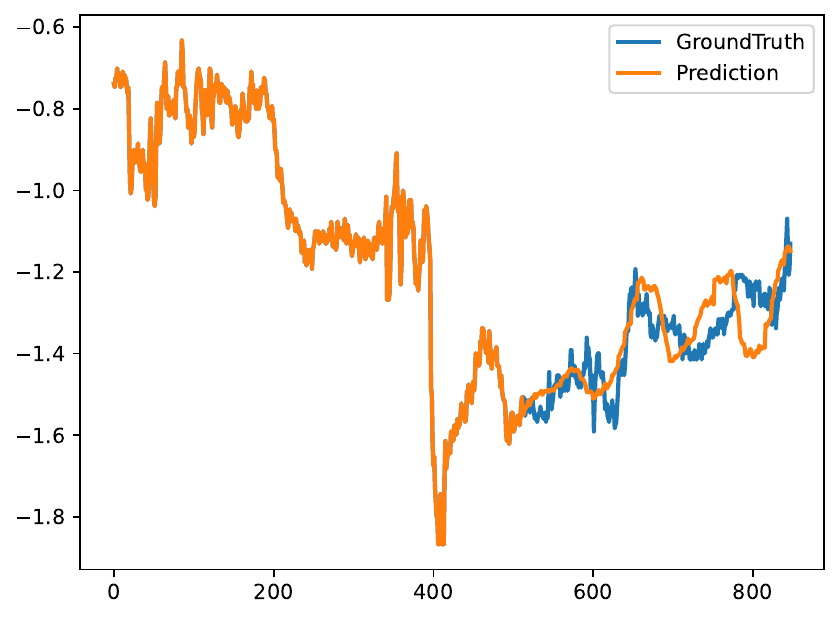}
		\end{minipage}
		\begin{minipage}{0.195\linewidth}
			\includegraphics[width=\linewidth,height=3.05cm]{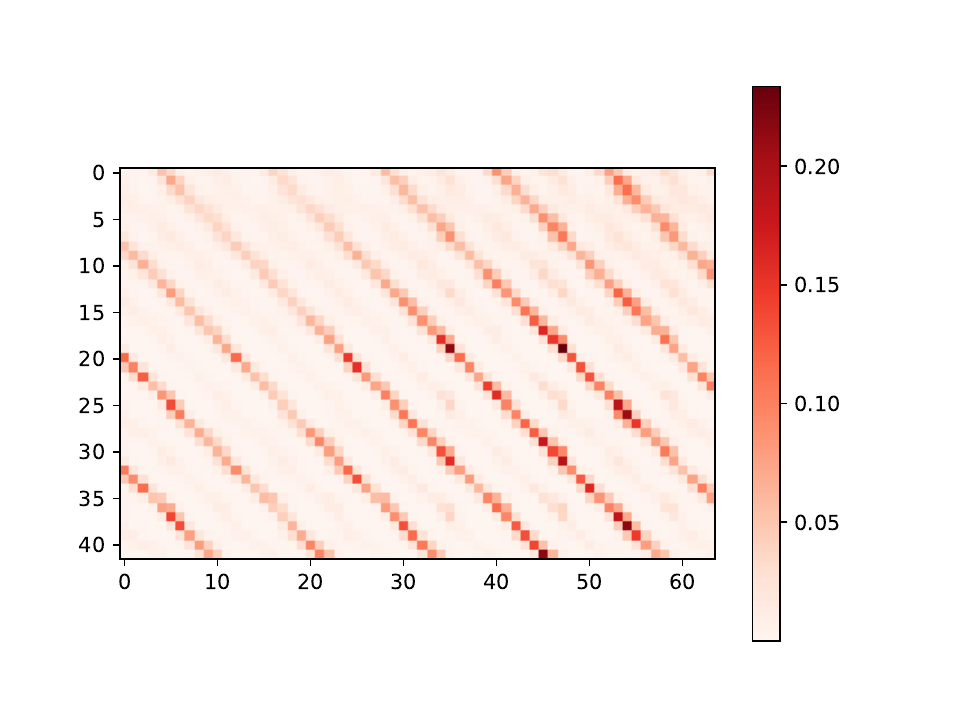}
		\end{minipage}
		\begin{minipage}{0.195\linewidth}
			\includegraphics[width=\linewidth,height=3.05cm]{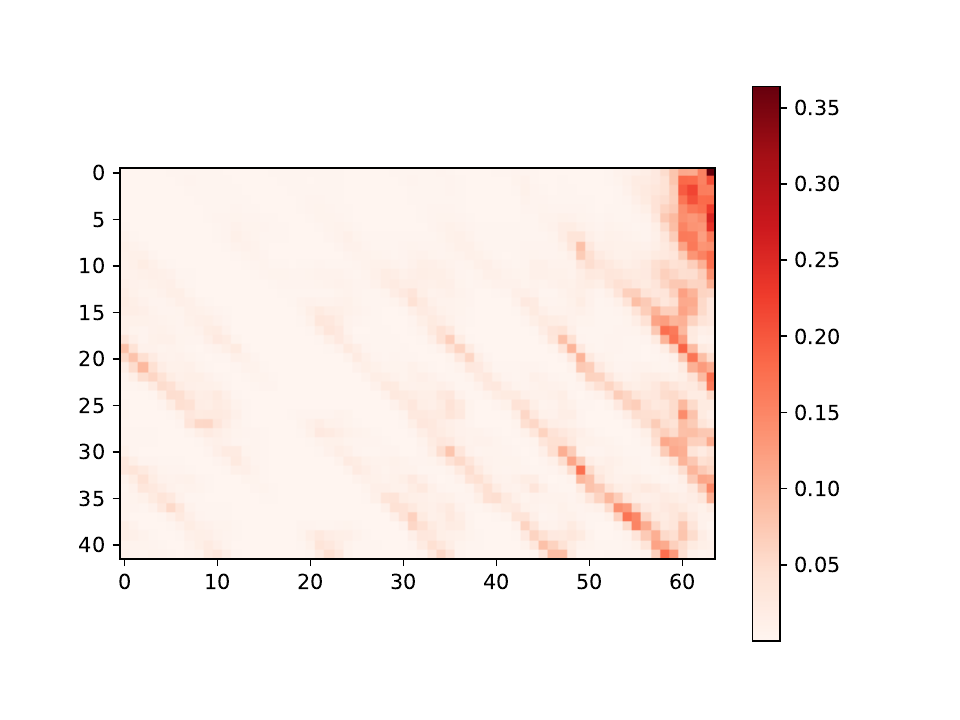}
		\end{minipage}
		\begin{minipage}{0.195\linewidth}
			\includegraphics[width=\linewidth,height=3.05cm]{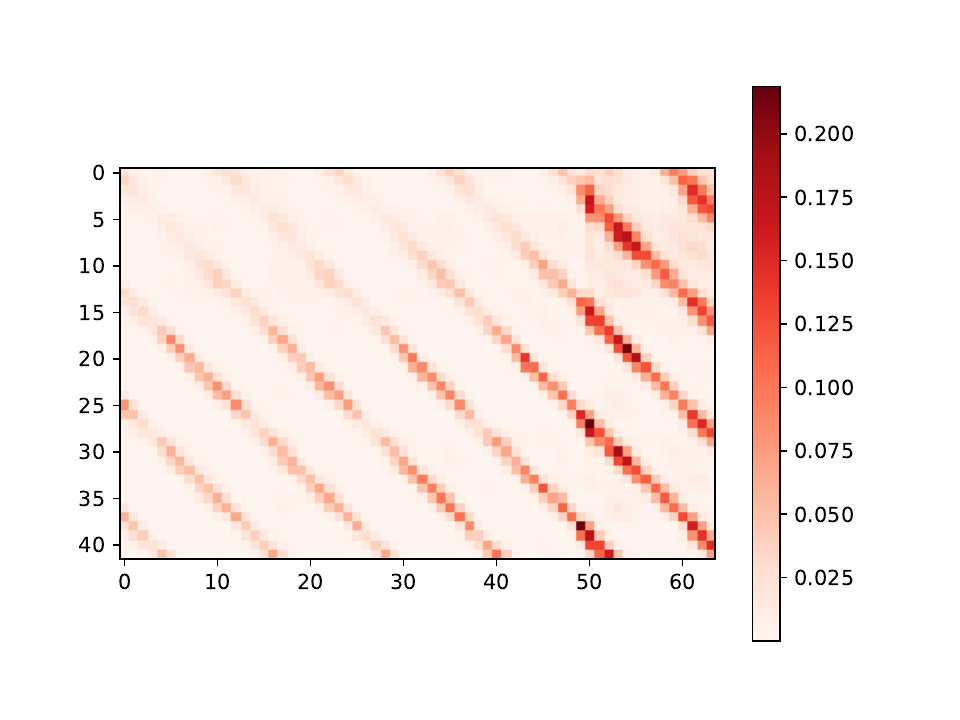}
		\end{minipage}
		\begin{minipage}{0.195\linewidth}
			\includegraphics[width=\linewidth,height=3.05cm]{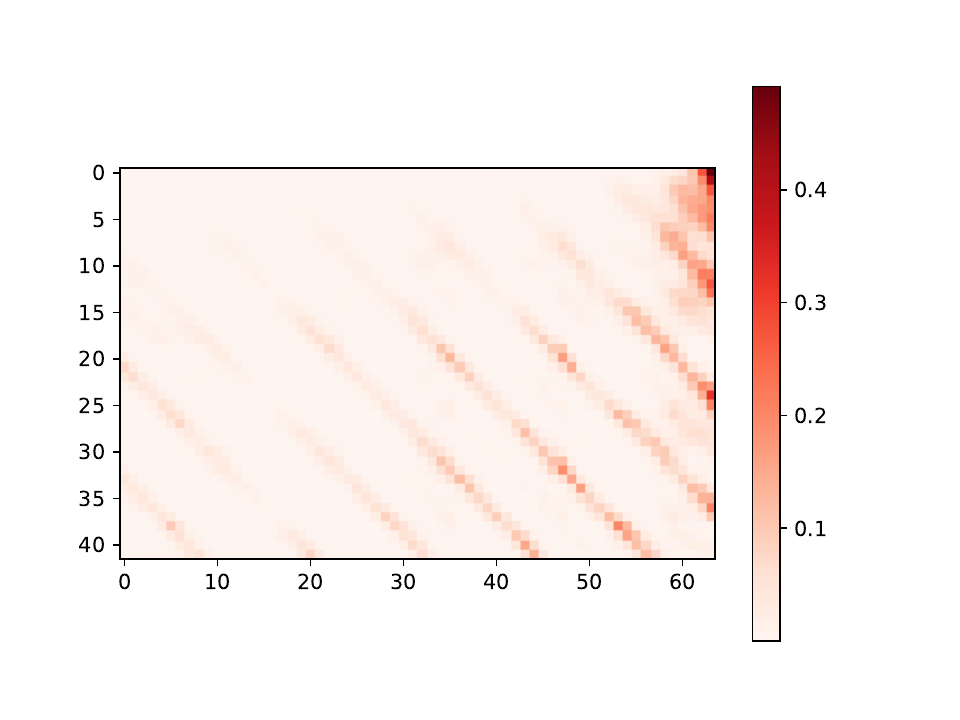}
		\end{minipage}
		\qquad		
	
		\begin{minipage}{0.195\linewidth}
			\includegraphics[width=\linewidth]{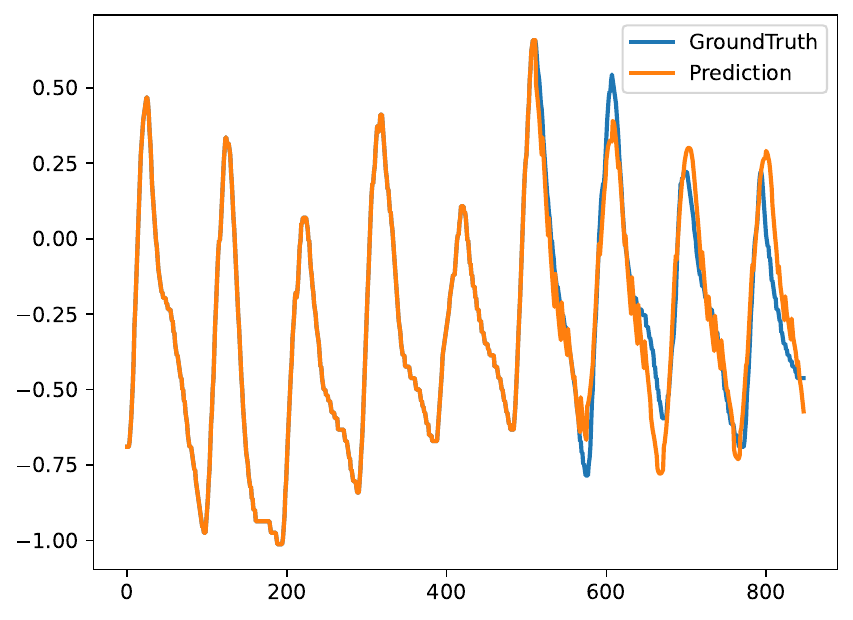}
		\end{minipage}
		\begin{minipage}{0.195\linewidth}
			\includegraphics[width=\linewidth,height=3.05cm]{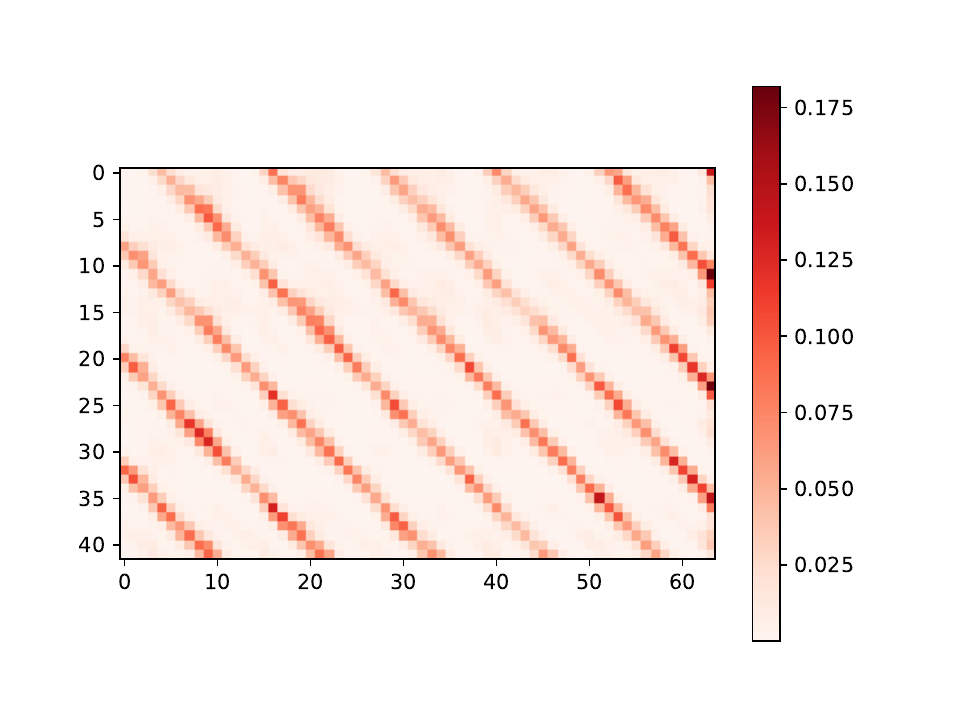}
		\end{minipage}
		\begin{minipage}{0.195\linewidth}
			\includegraphics[width=\linewidth,height=3.05cm]{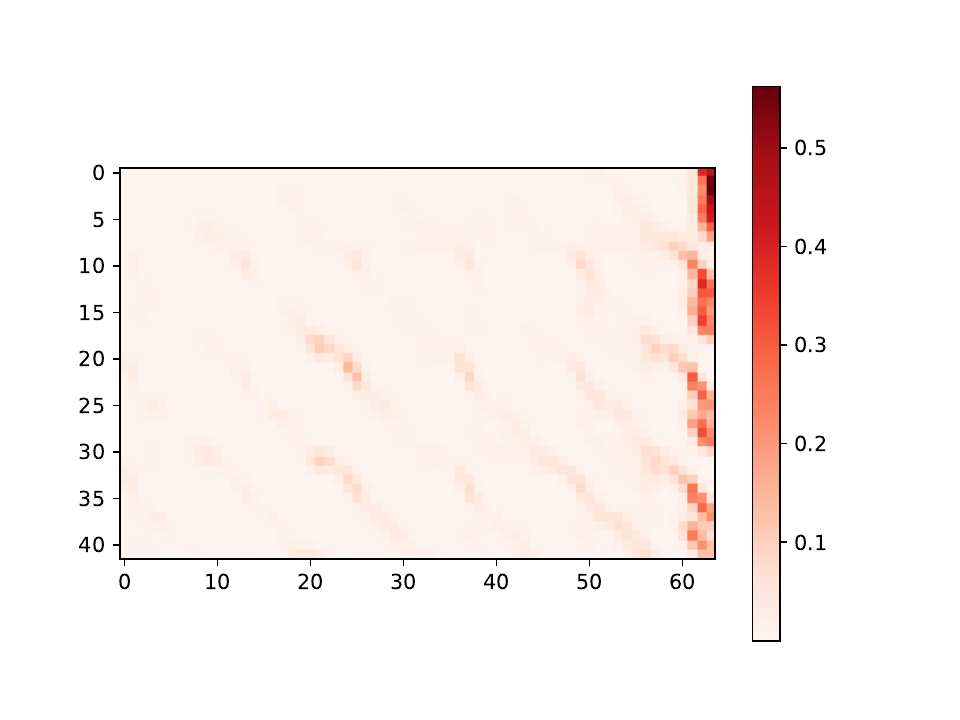}
		\end{minipage}
		\begin{minipage}{0.195\linewidth}
			\includegraphics[width=\linewidth,height=3.05cm]{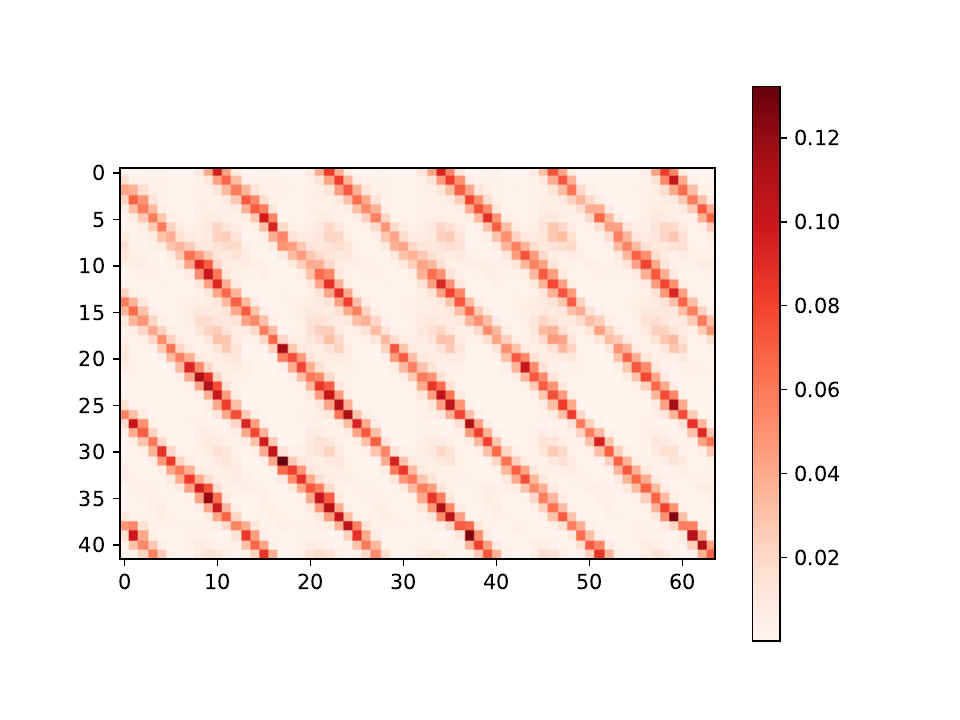}
		\end{minipage}
		\begin{minipage}{0.195\linewidth}
			\includegraphics[width=\linewidth,height=3.05cm]{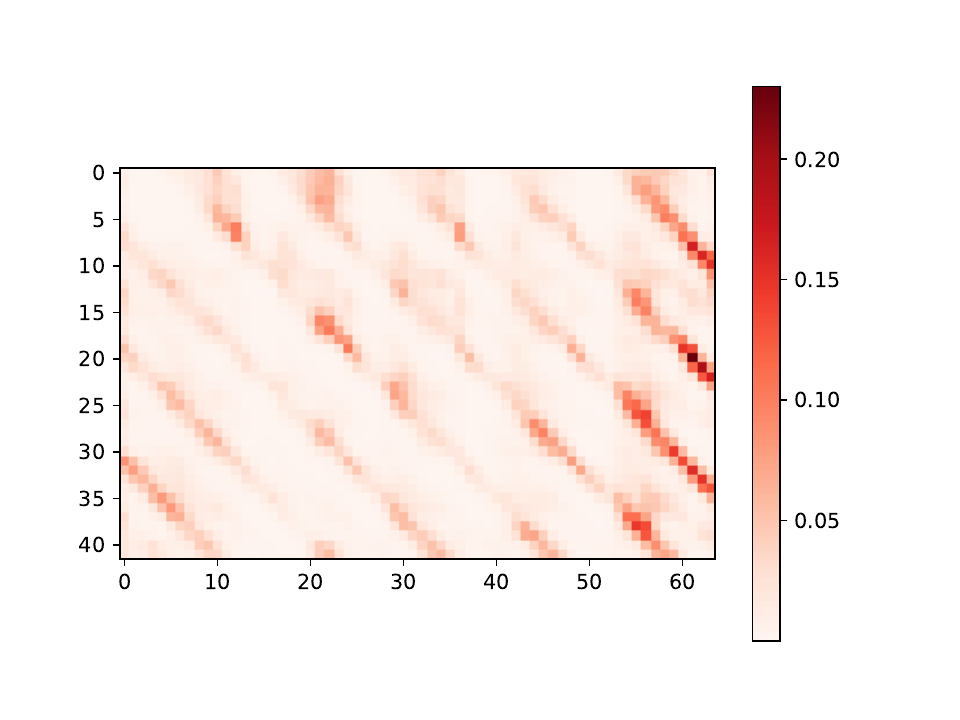}
		\end{minipage}
	\end{center}
	\caption{Illustration of the Attention Maps with an input length 512 and output length 336 on ETTh1, ETTh2, ETTm1 and ETTm2 datasets from top to bottom respectively.}
	\label{fig3}
\end{figure*}

\end{document}